\documentclass[a4paper,fleqn]{cas-sc}

\usepackage[authoryear]{natbib}
\usepackage{amssymb}
\usepackage{amsmath}
\usepackage{amsmath,amssymb,amsfonts}
\usepackage{algorithmic}
\usepackage{graphicx}
\usepackage{textcomp}
\usepackage{booktabs}      % \toprule/\midrule/\bottomrule）
\usepackage{multirow}      % \multirow）
\usepackage{subcaption}
\usepackage{multicol}
\usepackage{array}
\usepackage{textcomp}
\usepackage{makecell}   % 支持 makecell
\usepackage{graphicx}   % 支持 resizebox
\usepackage{adjustbox}
\def\tsc#1{\csdef{#1}{\textsc{\lowercase{#1}}\xspace}}
\tsc{WGM}
\tsc{QE}
\tsc{EP}
\tsc{PMS}
\tsc{BEC}
\tsc{DE}
\begin{document}
\let\WriteBookmarks\relax
\def\floatpagepagefraction{1}
\def\textpagefraction{.001}
\shorttitle{Leveraging social media news}
\shortauthors{J.N. Chi et~al.}
%\begin{frontmatter}

\title [mode = title]{HCHR: High-confidence and High-rationality Weakly Supervised Segmentation Framework for Thyroid Nodule Ultrasound Images}                      
% \tnotemark[1,2]

% \tnotetext[1]{This document is the results of the research
%    project funded by the National Science Foundation.}

% \tnotetext[2]{The second title footnote which is a longer text matter
%    to fill through the whole text width and overflow into
%    another line in the footnotes area of the first page.}

\author[1]{Jianning Chi}%[<options>]
%[orcid=0000-0002-9748-5619]
\cormark[1]
\ead{chijianning@mail.neu.edu.cn}

\author[1]{Zelan Li}
%[orcid=0000-0003-0156-6084]
\ead{2410844@stu.neu.edu.cn}

\author[2]{Xiangyu Li}
%[orcid=0000-0003-3218-1486]
\ead{lixiangyu@hit.edu.cn}

\author[1]{Geng Lin}
%[orcid=0009-0008-1700-0388]
\ead{2202048@stu.neu.edu.cn}

\author[1]{Mingyang Sun}
%[orcid=0009-0001-5972-9707]
\ead{2302161@stu.neu.edu.cn}

\author[3]{Ying Huang}
%[orcid=0000-0002-3361-7184]
\cormark[1]
\ead{huangying712@163.com}

\affiliation[1]{
    organization={Faculty of Robot Science and Engineering, Northeastern University},
    addressline={No. 195 ChuangXin Road, Hunnan District},
    city={Shenyang},
    postcode={110167},
    state={Liaoning},
    country={China}}
                
\affiliation[2]{
    organization={Harbin Institute of Technology},
    addressline={No. 92 Xidazhi St., Nangang District},
    city={Harbin},
    postcode={150001},
    state={Heilongjiang},
    country={China}}

\affiliation[3]{
	organization={Department of Ultrasound, Shengjing Hospital of China Medical University},
	addressline={No. 36 Sanhao Street, Heping District},
	city={Shenyang},
	postcode={110000},
	state={Liaoning},
	country={China}}

\cortext[1]{Corresponding author}

\begin{abstract}
Weakly supervised segmentation (WSS) methods can delineate thyroid nodules in ultrasound images using training data with clinical point annotations. Existing WSS methods directly utilize the topological geometric transformations of sparse annotations as pseudo-labels, then directly compare predictions with such pseudo-labels to optimize the segmentation model. However, the noisy information contained in the single-level pseudo-labels brings low-confident references as training targets, and the incomplete metrics introduced by the single-level comparison induce low-rational learning of discriminative information between nodules and background. To address these challenges, we propose a WSS framework that generates high-confidence labels by distilling multi-level reliable prior information from clinical point annotations, and constructs a high-rationality learning strategy of comparing multi-level attributes between predictions and the proposed pseudo-labels. Specifically, our approach integrates topological geometric transformations of clinical point annotations with outputs from the MedSAM model prompted by these annotations, generating high-confidence location, foreground, and background labels. Additionally, we introduce a high-rationality learning strategy consisting of: (1) an alignment loss that measures spatial consistency between predictions and location labels, guiding the network to learn nodule feature spatial arrangements; (2) a contrastive loss that encourages feature consistency within the same region while distinguishing between foreground and background features; and (3) a prototype correlation loss that ensures consistency between feature correlations with foreground and background prototypes, refining nodule boundary delineation. Experimental results show that our method achieves state-of-the-art performance on the TN3K and DDTI datasets, proving its potential of automatically segmenting thyroid nodules in clinical practice. The code is publicly available at \href{https://github.com/bluehenglee/MLI-MSC}{HCHR}.
% \noindent\texttt{\textbackslash begin{abstract}} \dots 
% \texttt{\textbackslash end{abstract}} and
% \verb+\begin{keyword}+ \verb+...+ \verb+\end{keyword}+ 
% which contain the abstract and keywords respectively. 
% Each keyword shall be separated by a \verb+\sep+ command.
\end{abstract}


\maketitle

\section{Introduction}
\label{sec:introduction}
Thyroid nodule segmentation in the ultrasound image is critical for accurate thyroid disease diagnosis~\cite{tessler2017Ti-rads, chen2020review},  but suffers from the blurred structures of anatomy with speckle noise, making it highly dependent on the expertise of the radiologist~\citet{khor2022ultrasound}. Employing deep learning algorithms~\cite{chen2022mtnnet, chi2023htunet, ozcan2024enhancedtransunet, xiang2025federated} for thyroid nodule segmentation can significantly enhance diagnostic efficiency for healthcare professionals. While fully supervised algorithms~\cite{ronneberger2015unet,cai2020Denseunet,SETR,Swin-unet,tao2022cenet} achieve promising performance on specific datasets where precise ground truth masks are available for training, acquiring a large number of delicate annotations remains resource-intensive and time-consuming~\cite{liu2024procns}. 

Weakly supervised segmentation (WSS) algorithms offer attractive alternatives by utilizing coarse annotations such as bounding boxes~\cite{zhang2020scrf,mahani2022uncrf,tian2021BoxInst,chen2024region}, points~\cite{zhao2020weakly,li2023wsdac,zhao2024IDMPS}, or scribbles~\cite{wang2023s2me,han2024dmsps,li2024scribformer} to achieve accurate segmentation results. WSS approaches are particularly suitable for thyroid nodule segmentation in clinical practice, since they can utilize the clinical point (aspect ratio) annotations as training supervision and obtain precise segmentation results~\cite{zhao2024IDMPS}. 
For instance, Zhao et al.~\cite{zhao2024IDMPS} generate basic geometric shapes as radical and conservative labels for asymmetric learning, Chi et al.~\cite{chi2025coarse} calibrated semantic features into a rational spatial distribution under the indirect, coarse guidance of the bounding box mask, and Li et al.~\cite{li2023wsdac} generate octagon labels from point annotations as an initial contour and iteratively deform the contour.  

However, existing weakly supervised methods still face the following challenges: 1) pseudo-labels generated based on topological geometric priors only are of low-confidence~\cite{zhao2024IDMPS, chi2025coarse, li2023wsdac, zhang2020scrf, mahani2022uncrf, lei2023uncertainty2, fan2024uncertainty}, introducing label noise and potentially misleading training according to these uncertain or ambiguous conditions; 2) rigid learning strategies constructed by directly comparing the segmentation with single-level or fixed-shape segmentation pseudo-labels~\cite{zhang2020scrf, mahani2022uncrf, du2023weakly3D, zhai2023paseg}, which severely limiting their flexibility and adaptability in handling diverse and complex nodule variations. It is worth note that the ``level'' refers to ``the segmentation attributes such as location and shape or entire results'' instead of ``the number of pseudo-labels''.

%However, existing weakly supervised methods still face the following challenges: 1) they typically generate low-confidence pseudo-labels based on topological geometric priors only~\cite{zhao2024IDMPS, chi2025coarse, li2023wsdac, zhang2020scrf, mahani2022uncrf, lei2023uncertainty2, fan2024uncertainty}, introducing label noise and potentially misleading training according to these uncertain or ambiguous conditions; 2) they primarily adopt rigid learning strategies such as comparing the segmentation with single-level or fixed-shape segmentation pseudo-labels~\cite{zhang2020scrf, mahani2022uncrf, du2023weakly3D, zhai2023paseg}, which severely limiting their flexibility and adaptability in handling diverse and complex nodule variations. It is worth note that the 'level' refers to 'the segmentation attributes such as location or shape' instead of 'the number of pseudo-labels'.

To address the aforementioned challenges, we propose a framework where noise-free, high-confidence pseudo labels are generated at the very beginning by distilling multi-level reliable prior information from clinical point annotations, and a comprehensive, high-rationality learning strategy is constructed by comparing multi-level attributes between predictions and the proposed pseudo-labels as well. Specifically, a series of reliable attributes of weak annotations is utilized to generate multi-level initial topological shapes through geometric transformations. They are then integrated with the results obtained from MedSAM model prompted by weak annotations, leading to high-confidence multi-level pseudo-labels with over 99.5\% precision in representing foreground and background regions. Furthermore, we proposed a series of loss functions to learn location-level, region-level, and boundary-level discriminative information by comparing multi-level attributes between predictions and the generated labels: 1) Alignment loss that measures the spatial projection consistency between the segmentation and the box label, guiding the network to perceive the location arrangement of nodule features; 2) Contrastive loss that reduce the intra-class variance of foreground and background features while increasing the inter-class variance between foreground and background features, guiding the network to capture the regional distribution of nodule and background features; 3) Prototype correlation loss that measures the consistency between dynamic correlation maps derived by comparing deep features with foreground and background prototypes respectively, so that the uncertain regions are gradually evolved to precise nodule edge delineation. 

In summary, the contributions of our work are as follows:
\begin{enumerate}
\itemsep=0pt
\item We claim a series of weakly supervised segmentation objectives that precise segmentation could be obtained by learning the location and shape of nodules from partial reference. Following these objectives, high-confidence pseudo-labels and a high-rationality learning strategy are essential for improving the performance of weakly supervised thyroid nodule segmentation.
% We clarify a partial-reference weakly-supervised segmentation objective theory, which claims that we can obtain precise and entire segmentation results by learning location and shape of nodule from multi-level high-confidence labels respectively. To achieve these objectives, high-confidence multi-level references and high-rationality multi-level learning strategy are essential.
\item We propose a high-confidence label generation strategy that fuses geometric transformations of point annotations and segmentation provided by the prompt-guided MedSAM model. The generation of multi-level location, foreground, and background labels as learning reference prevents the misleading information from low-confidence labels during network training.
\item We introduce a series of high-rationality losses, including alignment, contrastive, and prototype correlation loss. These losses guide the network to capture multi-level discriminative information of thyroid nodules' locations and shapes, significantly enhancing the reliability of the training process.
\item Extensive experiments on the publicly available thyroid nodule ultrasound datasets, TIN3K~\cite{gong2021tn3k} and DDTI~\cite{pedraza2015DDTI}, demonstrate that our proposed framework does not only achieve state-of-the-art segmentation performance but also exhibits remarkable generalizability, making it highly adaptable to mainstream segmentation backbones.
% Extensive experiments on the publicly available thyroid nodule ultrasound dataset TIN3K~\cite{gong2021tn3k} and DDTI~\cite{pedraza2015DDTI} show that our proposed framework achieves state-of-the-art segmentation performance and high scalability to mainstream segmentation backbones.
% Extensive experiments show that our method achieves state-of-the-art on the publicly available thyroid nodule ultrasound dataset TN3K~\cite{gong2021tn3k} and DDTI~\cite{pedraza2015DDTI}.
\end{enumerate}

\section{Related Work}
\label{sec:related_work}
\subsection{Medical Image Segmentation}
Medical image segmentation serves as a cornerstone in radiology and pathology, facilitating computer-assisted automatic analysis for diagnostic and therapeutic planning~\cite{obuchowicz2024review}. Deep learning has driven remarkable progress in this field~\cite{rayed2024deepreview, das2024nodulereview}, with fully supervised methods evolving through three key phases:
(1) CNN-based Architectures: Encoder-decoder frameworks such as U-Net~\cite{ronneberger2015unet} and its variants (e.g., UNet++~\cite{zhou2018unet++}, nnU-Net~\cite{isensee2021nnunet}, CE-Net~\cite{tao2022cenet}) extracted multi-scale features while preserving spatial resolution, establishing the standard for medical segmentation. (2) Transformer-enhanced Models~\cite{vaswani2017attentiontransformer}: Vision Transformers (ViTs)~\cite{dosovitskiy2020vit} addressed the limited long-range dependency modeling of CNNs. Networks like TransUNet~\cite{chen2024transunet} and Swin-Unet~\cite{Swin-unet} integrated self-attention mechanisms to capture global context, significantly improving accuracy for complex structures~\cite{ozcan2024enhancedtransunet, bi2023bpat}. (3) Prompt-guided Foundation Models: Emerging paradigms like the Segment Anything Model (SAM)~\cite{kirillov2023SAM} and its medical extension MedSAM~\cite{ma2024MED-SAM} leveraged prompt engineering (e.g., bounding boxes) to enable zero-shot segmentation across diverse objects, with training on million-level or even billion-level large-scale image datasets.

Despite the architecture of the segmentation network, the learning strategy has attracted great research attention. For instance, (1) contrastive Learning optimized feature spaces by attracting positive pairs and repelling negative pairs~\cite{khosla2020supcontrast, wang2021ContrastiveSeg,zhao2021contrastive-pixel,wang2022contrastmask}. Supervised extensions~\cite{khosla2020supcontrast, wang2021ContrastiveSeg} utilized segmentation masks to sample category-specific pixels, demonstrating strong boundary delineation capabilities. (2) Prototype Learning clustered intra-class feature centers to guide segmentation~\cite{snell2017prototypical, zhou2022prototypereview, zhou2024prototypeseg}. Early approaches used non-learnable prototypes (e.g., class-mean features~\cite{zhou2022prototypereview}) for nearest-prototype prediction. Advanced methods integrated uncertainty-aware mechanisms (e.g., entropy-based prototype fusion~\cite{lu2023upcol}) and medical-specific adaptations like tumor sub-center prototypes for breast lesion segmentation~\cite{zhou2024prototypebreast}.

Although providing competitive performance, current approaches faced several limitations. Firstly, fully supervised networks required extensive pixel-level labels, incurring prohibitive annotation costs~\cite{ma2024tnseg}. Secondly, the heavy reliance on prompts in prompt-guided foundation models~\cite{zhao2024sam, ma2024MED-SAM} may present limitations in fully automatic settings or situations with impractical human intervention, making them unsuitable for clinical applications.

\subsection{Weakly Supervised Segmentation}
Weakly supervised learning represents an emerging learning paradigm that requires only a small amount of coarse-grained annotation information for model training~\cite{guo2024weaklysup, lin2024weaklysup}. This approach significantly reduces the annotation workload while maintaining promising segmentation accuracy~\cite{roth2021weaklysegmentation}.

Typical methods focused on directly exploiting sparse annotations or inaccurate geometric shapes to generate pseudo-labels~\cite{liu2024procns} for pixel-to-pixel region learning. For instance, some approaches incorporated conditional probability modeling techniques, such as conditional random fields (CRF)~\cite{zhang2020scrf, mahani2022uncrf}, and uncertainty estimates~\cite{lei2023uncertainty2, fan2024uncertainty} into the training process directly using weakly supervised labels to learn predictions. Other methods generated pseudo-masks based on topological geometric transformations~\cite{wang2023s2me, zhao2024IDMPS, wang2023WSL-MIS, li2023wsdac}. Tang et al.~\cite{tang2021weakly} utilized ellipse masks as pseudo-labels and proposed a regional level set (RLS) loss for optimizing lesion boundary delineation.

Recently, BoxInst~\cite{tian2021BoxInst} employed box annotations to localize segmentation targets, combining color similarity with graph neural networks to delineate segmentation boundaries. 
% Nevertheless, for thyroid ultrasound images with low contrast and blurred boundaries, color similarity cannot fully indicate the thyroid nodule's boundary. 
Inspired by BoxInst, Du et al.~\cite{du2023weakly3D} proposed an algorithm that learned the location and geometric prior of organs mainly relying on the region of interest (ROI) feature, which was useful for organ segmentation with fixed prior shapes but not suitable for diverse and complex shapes.

Although recent advancements in weakly supervised segmentation methods have yielded promising results, they still relied on the quality of weak labels and the learning strategy.

\subsection{Thyroid Nodule Segmentation}
In recent years, algorithms for thyroid nodule segmentation based on fully supervised precise labels~\cite{chi2023htunet, chen2024mlmseg, wu2024medsegdiff, li2023novelthyroid, gong2023thyroid, ma2024tnseg} have been extensively studied. Compared to other imaging modalities like CT and MRI, ultrasound images suffer from issues such as speckle noise and low contrast, increasing the difficulty of thyroid nodule segmentation~\cite{das2024nodulereview}. The thyroid ultrasound image segmentation methods mainly concentrated on improving feature extraction through the innovation of a network architecture. For example, Chi et al.~\cite{chi2023htunet} utilized transformer attention mechanisms to capture intra-frame and inter-frame contextual features in thyroid regions, yielding competitive segmentation results. Chen et al.~\cite{chen2024mlmseg} developed a multi-view learning model to encode local view features and a cross-layer graph convolution module to capture correlations between high-level and low-level features. Wu et al.~\cite{wu2024medsegdiff} introduced dynamic conditional encoding and a feature frequency parser based on the diffusion probabilistic model, achieving excellent segmentation results for thyroid nodules.

In order to better suit the clinical scene where delicate segmentation labels are difficult to obtain, the weakly supervision of thyroid nodule segmentation has gained great progress in recent years~\cite{yu2022sse-wssn,li2023wsdac, zhao2024IDMPS, chi2025coarse}. For example, Li et al.~\cite{li2023wsdac} proposed a method that generated octagons from point annotations to serve as initial contours to iteratively refine the boundaries of thyroid nodules through active contour learning. Zhao et al.~\cite{zhao2024IDMPS} employed quadrilaterals as conservative labels and irregular ellipses as radical labels while introducing dual-branch designs to improve the consistency of pseudo-labels during training, thereby enhancing prediction accuracy. Chi et al.~\cite{chi2025coarse} proposed a weakly supervised dual-branch learning framework to optimize the backbone segmentation network by calibrating semantic features into a rational spatial distribution under the indirect and coarse guidance of the bounding box mask.

For thyroid nodule segmentation, weak supervision has been considered as a trend with various promising models have been proposed. However, due to the quality of ultrasound images and the morphological variability of nodules, applying weak supervision to thyroid nodule segmentation in ultrasound images still faced two issues: pseudo-label noise and irrational learning strategies.

% The package is available at author resources page at Elsevier
% (\url{http://www.elsevier.com/locate/latex}).
% The class may be moved or copied to a place, usually,
% \verb+$TEXMF/tex/latex/elsevier/+, %$%%%%%%%%%%%%%%%%%%%%%%%%%%%%
% or a folder which will be read                   
% by \LaTeX{} during document compilation.  The \TeX{} file
% database needs updation after moving/copying class file.  Usually,
% we use commands like \verb+mktexlsr+ or \verb+texhash+ depending
% upon the distribution and operating system.

\begin{figure*}[!htbp]
	\centering
	\includegraphics[width=.98\textwidth]{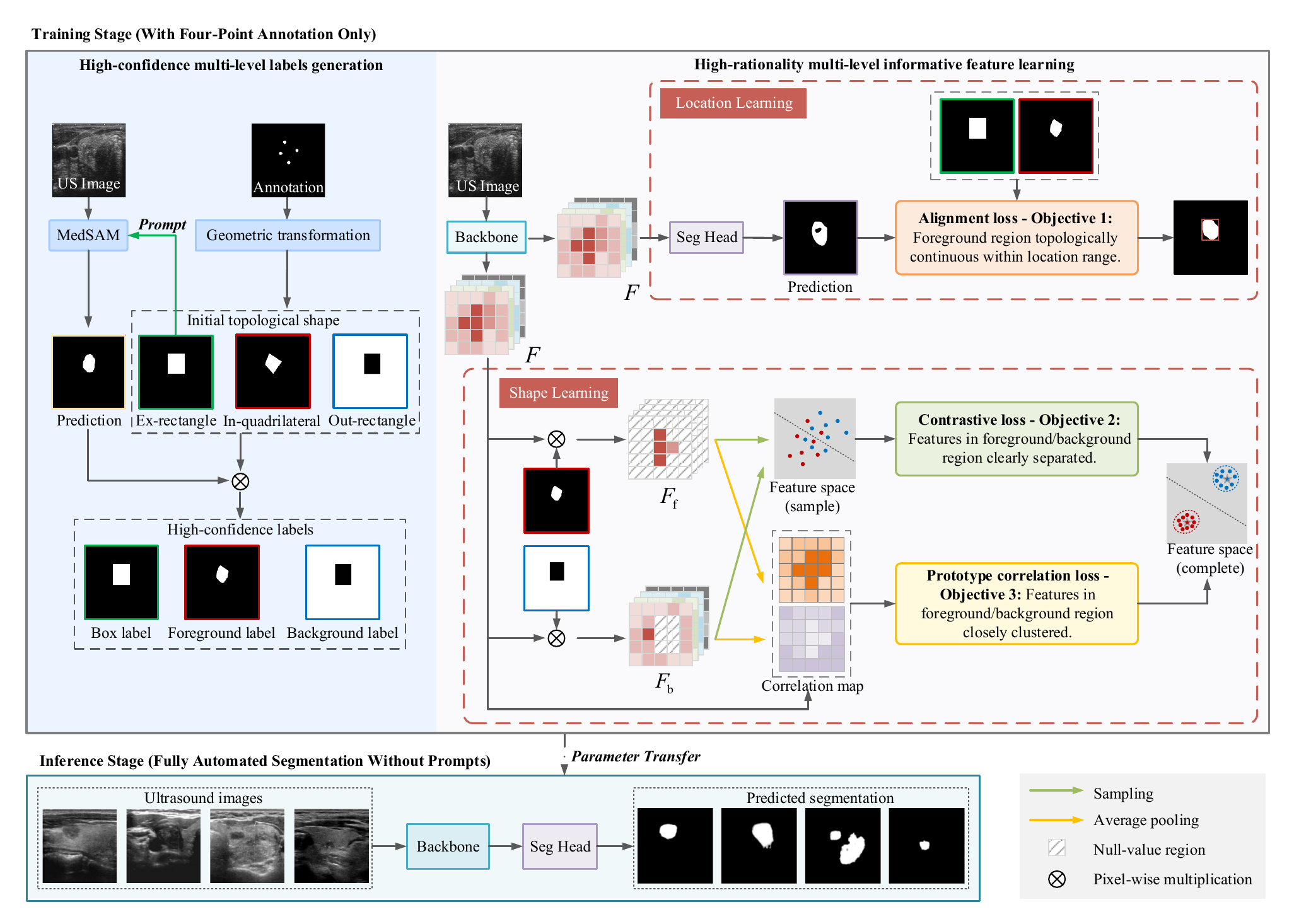}
	\caption{Overview of the proposed HCHR framework. \textbf{Training process}: 1) High-confidence multi-level labels are generated by fusing MedSAM results with geometric transformations as feature learning references. 2) High-rationality learning strategy is constructed through post-processing of deep features to optimize feature expression. The upper branch leverages the segmentation head to generate segmentation predictions and applies alignment loss for location-level learning. The lower branch refines feature representations for the nodule shape by calculating contrastive loss for region-level learning and the prototype correlation loss for boundary-level learning. \textbf{Inference process}: Images are fed into the network, followed by the segmentation head to get the results \textbf{with no need for any prompts}.}
	\label{fig:framework}
\end{figure*}
\section{Method}
\label{sec:method}
\subsection{Overall Framework}
As shown in Fig.~\ref{fig:framework}, we propose a novel weakly supervised segmentation (WSS) framework named HCHR for thyroid nodule segmentation. The framework consists of a high-confidence multi-level labels generation flow and a high-rationality multi-level learning strategy branch. In the label generation flow, we integrate prompted MedSAM results with geometric transformations of point annotations to generate high-confidence labels as training references. In the learning strategy branch, we compare multi-level attributes between predictions and generated labels by combining alignment loss, contrastive loss, and prototype correlation loss, in order to jointly learn segmentation location and delicate shape information with high rationality.

\subsection{Partial Reference Segmentation Objective}
\label{sec:objective}
Image segmentation aims to locate and delineate regions precisely. Fully supervised methods can use ground truth labels with clear location $Loc$ and shape $S$ to learn feature distributions through pixel-to-pixel comparison. However, weakly supervised approaches for thyroid ultrasound images can only refer to coarse pseudo-label cues without precise shape details, making it necessary to directly assess the rationality of feature distributions across the image domain.

Let $I$ denote the image, with $R_f$ as the foreground region and $R_b$ as the background region. Sample sets within these regions are denoted by $X_f$ and $X_b$. The network feature is represented by $\boldsymbol{\mathcal{F}}$. The subscripts $f$ and $b$ represent foreground and background. To complete the segmentation task in the image, besides the basic conditions:

1) the union of $R_f$ and $R_b$ equals the entire image ( $R_f \cup R_b = I$ ); 

2) their intersection is empty ( $R_f \cap R_b = \emptyset$ );

3) both regions of $R_f$ and $R_b$ are fully connected, \\
weakly supervised algorithms to identify regions $R_f$ and $R_b$ \textbf{should also satisfy}:

\textbf{Objective 1}: $R_f$ should lie within a predefined location range $Loc$, while $R_b$ should lie outside this range.
%\begin{align}
%%	R_f \subseteq Loc \And R_b \cap Loc = \emptyset
%	\label{condition1}
%\end{align}
\begin{equation}
\begin{aligned}
R_f \subseteq Loc \And R_b \cap Loc = \emptyset.
\label{condition1}
\end{aligned}
\end{equation}

\textbf{Objective 2}: The cluster of the sample within the foreground regions $\boldsymbol{C_{(X_f)}}$ should closely match the reference foreground cluster $\boldsymbol{C_{(R_f)}}$, while the sample within background regions cluster $\boldsymbol{C_{(X_b)}}$ should align with the reference background cluster $\boldsymbol{C_{(R_f)}}$, as shown below:
\begin{equation}
\begin{aligned}
&\mathcal{D}(\boldsymbol{C_{(X_f)}}, \boldsymbol{C_{(R_f)}}) < \epsilon_f,\\
&\mathcal{D}(\boldsymbol{C_{(X_b)}}, \boldsymbol{C_{(R_b)}}) < \epsilon_b,
\label{condition2}
\end{aligned}
\end{equation}

where $\mathcal{D}(\cdot, \cdot)$ denotes the distance between the feature clusters and $\epsilon$ are small thresholds that ensure the proximity of the segmentation prototypes to the reference prototypes.

\textbf{Objective 3}: The distribution of pixel features in the sampled foreground $\boldsymbol{\mathcal{F}_{(x, x \in X_f)}}$ should closely match the foreground prototype $\boldsymbol{P_{(R_f)}}$, and those in the predicted background $\boldsymbol{\mathcal{F}_{(x,x \in X_b)}}$ should closely match the background prototype $\boldsymbol{P_{(R_b)}}$, as shown below:
\begin{equation}
\begin{aligned}
&\mathcal{D}(\boldsymbol{\mathcal{F}_{(x,x \in X_f)}}, \boldsymbol{P_{(R_f)}}) < \delta_f,\\
&\mathcal{D}(\boldsymbol{\mathcal{F}_{(x,x \in X_b)}}, \boldsymbol{P_{(R_b)}}) < \delta_b,
\label{condition3}
\end{aligned}
\end{equation}
where $\delta$ are thresholds ensuring similarity between the sampled feature distribution and reference prototypes.

To achieve these objectives, high-confidence references used in the optimal process are essential as follows: 

\textbf{Reference 1}: A correct range must be provided to guide the segmentation location process $Loc$.

\textbf{Reference 2}: High-confidence foreground and background region labels are necessary to define partial but precise foreground and background references $X_f$ and $X_b$ for image distribution $R_f$ and $R_b$ shape learning.

\subsection{High-confidence Multi-level labels Generation}
According to \textbf{Reference 1} and \textbf{Reference 2} discussed in Sec. \ref{sec:objective}, weakly supervised segmentation requires nodule location labels for location learning and region distribution labels for shape learning. In this section, we integrate geometric transformations of point annotations and segmentation from prompted MedSAM to generate high-confidence location labels $\boldsymbol{G_l} \in (0,1)^{H \times W}$ for location $Loc$ learning in Eq. \eqref{condition1}, as well as high-confidence foreground labels $\boldsymbol{X_{f}} \in (0,1)^{H \times W}$ and background labels $\boldsymbol{X_{b}} \in (0,1)^{H \times W}$ for shape $S$ learning in Eq. \eqref{condition2} and Eq. \eqref{condition3}.

Specifically, as illustrated in Fig.~\ref{fig:framework}, in the high-confidence labels generation phase, we derive three geometric transformations representing low-level topological information from clinical annotations:
\begin{itemize}
    \item Connecting the endpoints of the annotations along each axis to form quadrilateral regions.
    \item Identifying and filling the minimum bounding box enclosing these four points per target to create box regions encompassing all foreground pixels.
    \item Negating the bounding box regions results in obtaining background-only regions.
\end{itemize}

The regions that contain high-level semantic information are generated by prompted MedSAM:
\begin{itemize}
    \item Using MedSAM with prompts computed from point annotations to obtain segmentation masks that reflect anatomical distributions from input images.
\end{itemize}

Finally, we fuse the initial topological shape ex-rectangle $\boldsymbol{g_\text{b}}$, in-quadrilateral $\boldsymbol{g_\text{i}}$, and out-rectangle $\boldsymbol{g_\text{o}}$ and results $\boldsymbol{y_\text{medsam}}$ from prompted MedSAM to generate high-confidence location $\boldsymbol{G_l}$, foreground $\boldsymbol{X_f}$, and background labels $\boldsymbol{X_b}$.
\begin{equation}
\begin{aligned}
\boldsymbol{G_l} &= \boldsymbol{g_\text{b}} \vee \boldsymbol{y_\text{medsam}},\\
\boldsymbol{X_f} &= \boldsymbol{g_\text{i}}  \wedge \boldsymbol{y_\text{medsam}},\\
\boldsymbol{X_b} &= \boldsymbol{g_\text{o}} \wedge (1 - \boldsymbol{y_\text{medsam}}),
\label{hc-labels}
\end{aligned}
\end{equation}
where $\vee$ denotes the logical OR operation applied to the positions of two masks, while $\wedge$ represents the logical AND operation.

\subsection{High-rationality Multi-level Learning Strategy}
According to the \textbf{Objective 1, 2, and 3} discussed in Sec. \ref{sec:objective}, we design a high-rationality learning strategy that guides the network to learn nodule location and shape to satisfy all conditions for weakly supervised segmentation tasks, which consisting of the following three losses: alignment loss, contrastive loss and prototype correlation loss. 
\begin{figure*}[!htbp]
	\centering
	\includegraphics[width=0.92\textwidth]{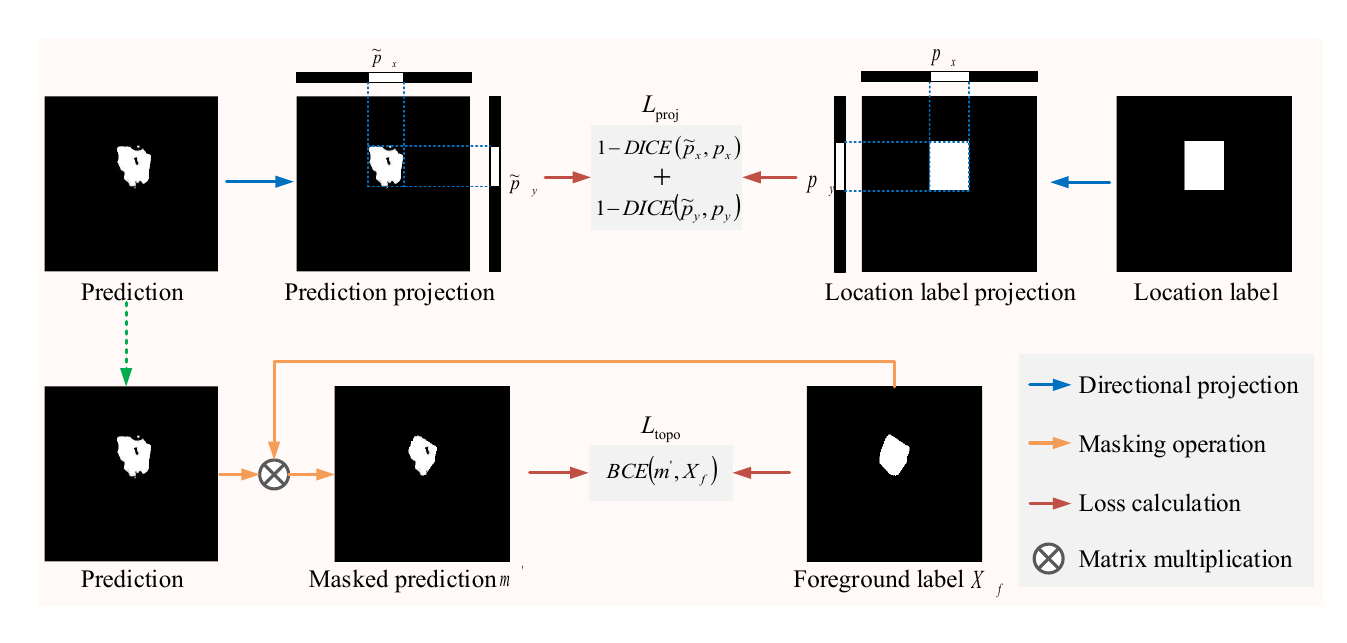}
	\caption{Diagram of alignment loss learning.}
	\label{fig:alignment_loss}
\end{figure*}
\subsubsection{Alignment Loss for Location-level Learning}
To learn the location information as described in Eq. \eqref{condition1}, the alignment loss consists of two components: 1) alignment of projected segmentation results with location labels, and 2) topological continuity loss in high-confidence foreground regions.

As showed in Fig.~\ref{fig:alignment_loss}, for a predicted result $\boldsymbol{m} \in (0,1)^{H \times W}$ and location label $\boldsymbol{G_l} \in (0,1)^{H \times W}$ derived from the points annotations, the directional projection~\cite{tian2021BoxInst} are defined as:
\begin{equation}
\begin{aligned}
\boldsymbol{p_{x}} = \max_{col}(\boldsymbol{G_l})&, \boldsymbol{p_{y}} = \max_{row}(\boldsymbol{G_l}),&\\
\boldsymbol{\tilde{p_x}} = \max_{col}(\boldsymbol{m})&, \boldsymbol{\tilde{p_y}} = \max_{row}(\boldsymbol{m}),
\label{proj}
\end{aligned}
\end{equation}
where $\boldsymbol{p_x} \in (0,1)^{W}$ and $\boldsymbol{p_y} \in (0,1)^{H}$ denote projections derived from the location label $\boldsymbol{G_l}$ onto the x-axis and y-axis, respectively, $\boldsymbol{\tilde{p_x}} \in (0,1)^{W}$ and $\boldsymbol{\tilde{p_y}} \in (0,1)^{H}$ are the projections of predicted result $\boldsymbol{m}$ onto the x and y axis. 

The first component $\mathcal{L}_{proj}$ of the alignment loss between the predicted mask and the location label is then computed as follows:
% \begin{equation}
\begin{align}
\mathcal{L}_\text{proj} = (1 - \frac{2 \times |\boldsymbol{\tilde{p_x}} \cap \boldsymbol{p_x|}}{|\boldsymbol{\tilde{p_x}}| + |\boldsymbol{p_x}|}) + (1 - \frac{2 \times |\boldsymbol{\tilde{p_y}} \cap \boldsymbol{p_y|}}{|\boldsymbol{\tilde{p_y}}| + |\boldsymbol{p_y}|}).
\end{align}
% \end{equation}

This projection loss provides region localization constraints with high feasibility and efficiency, but still faces issues such as segmentation results lying along the projection axis or containing holes and discontinuities in the topological shape. To solve these issues, we introduce topological continuity loss as the second component of the alignment loss. The topological continuity loss $\mathcal{L}_\text{topo}$ is defined as:
\begin{equation}
\begin{aligned}
\mathcal{L}_\text{topo} = -\left[ \boldsymbol{m'} \log (\boldsymbol{X_{f}}) + (1 - \boldsymbol{m'}) \log (1 - \boldsymbol{X_{f}}) \right],
\label{consistency}
\end{aligned}
\end{equation}
where $\boldsymbol{m'} = {\boldsymbol{m}} \wedge \boldsymbol{X_{f}}$ denotes predicted regions ${\boldsymbol{m}}$ covered by the high-confidence foreground $\boldsymbol{X_{f}}$.

The topological continuity loss ensures topological continuity within predicted foreground regions covered by the high-confidence foreground, where foreground pixels' precision of over $99.65\%$ as calculated in Table~\ref{tab:label_precision}. This specific loss function is exclusively utilized to identify foreground regions to prevent local optima in coordinate axis-based optimization.

By combining loss $\mathcal{L}_\text{proj}$ and loss $\mathcal{L}_\text{topo}$, we obtain the final alignment loss $\mathcal{L}_\text{alignment}$ as follows:
% \begin{eqnarray}
\begin{align}
\mathcal{L}_\text{alignment} = \mathcal{L}_\text{proj} + \mathcal{L}_\text{topo}.
% \end{eqnarray}
\end{align}

This loss allows weak supervision labels to constrain segmentation localization effectively while avoiding over-fitting to a single shape. 

\subsubsection{Contrastive Loss for Region-level Shape Learning}
\begin{figure*}[!htbp]
\centering
\includegraphics[width=0.92\textwidth]{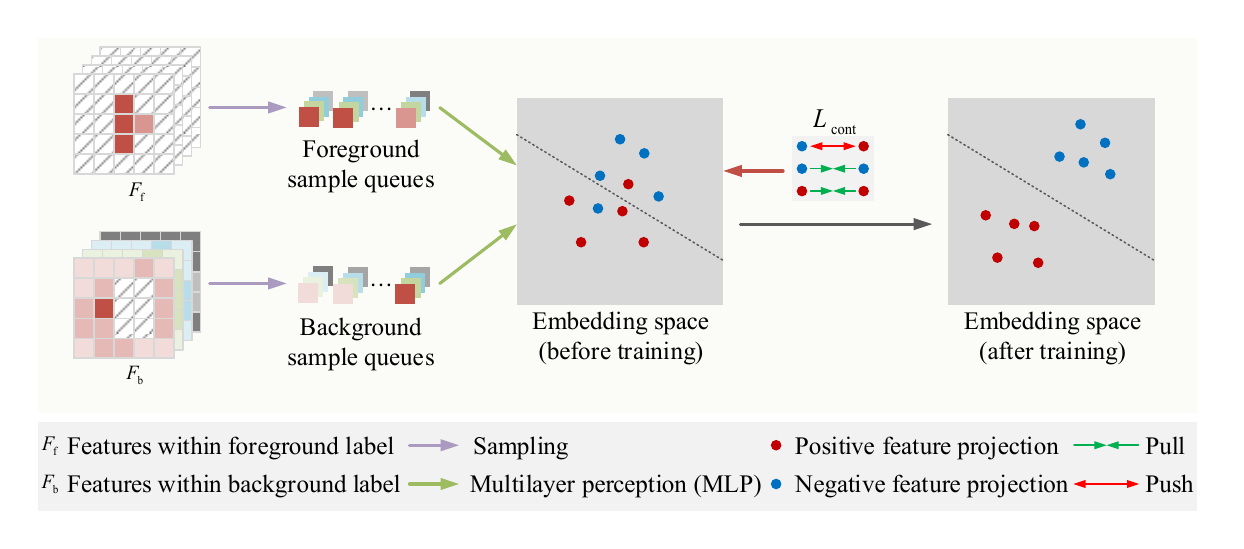}
\caption{Diagram of contrastive loss learning.}
\label{fig:contrastive}
\end{figure*}
As outlined in Sec. \ref{sec:objective}, we propose to learn the segmentation shape by refining the feature representation of the foreground and background regions to achieve Eq. \eqref{condition2}. Due to the features extracted by the network being inherently uncertain, multi-scale contrastive loss is exploited to optimize the inter-class variance and intra-class similarity of features sampled from the foreground and background label regions, ultimately refining the predicted foreground and background feature cluster extracted from the whole image domain.
% Due to the features extracted by the network being inherently uncertain, instead of directly comparing the sampled features cluster with the label features cluster, we employ multi-scale contrastive learning loss to optimize the inter-class variance and intra-class similarity of the foreground and background regions' features, ultimately refining the predicted cluster for both categories.

Given high-confidence foreground areas $\boldsymbol{X_{f}}$ and background areas $\boldsymbol{X_{b}}$. As shown in Fig.~\ref{fig:contrastive}, the generic contrastive loss with a sampling scale size $k \times k$, denoted as 
% \begin{equation}
% \begin{aligned}
% \mathcal{L}_\text{cnt}^k = \frac{1}{n} \sum_{\boldsymbol{q_k}^{+} \in \boldsymbol{Q_{p}}} -\log \left( \frac{\exp{{\frac{\boldsymbol{s^+}}{\tau}}}}{(\exp{{\frac{\boldsymbol{s^+}}{\tau}}} + \sum_{\boldsymbol{q_k^{-}} \in \boldsymbol{Q_{n}}} \exp{{\frac{\boldsymbol{s^-}}{\tau}}}} \right),
% \end{aligned}
% \end{equation}
\begin{equation}
\begin{aligned}
\mathcal{L}_\text{cnt}^k = \frac{1}{n} \sum_{\boldsymbol{q_k}^{+} \in \boldsymbol{Q_{p}}} -\log \left( \frac{\exp\left(\boldsymbol{q_k} \cdot \boldsymbol{q_k^{+}}/\tau\right)}{\exp\left(\boldsymbol{q_k} \cdot \boldsymbol{q_k^{+}}/\tau\right) + \sum_{\boldsymbol{q_k^{-}} \in \boldsymbol{Q_{n}}} \exp\left(\boldsymbol{q_k} \cdot \boldsymbol{q_k^{-}}/\tau\right)} \right),
\end{aligned}
\end{equation}
where $\boldsymbol{Q_{p}}$ and $\boldsymbol{Q_{n}}$ represent the sets of positive and negative feature embedding queues extracted from foreground and background feature map regions, respectively. $\boldsymbol{q_k^{+}}$ denotes a foreground feature sample of size $k\times k$, while $\boldsymbol{q_k^{-}}$ represents corresponding background feature samples. The anchor feature queue $\boldsymbol{q}$ is selected from high-confidence foreground regions $\boldsymbol{X_{f}}$. The temperature parameter $\tau > 0$ controls the slope of the loss function and its smoothness. Empirically, we evaluate the contrastive loss at scale sizes $k=3$ and $k=1$, with $\tau = 0.07$.

The contrastive loss is designed to minimize the global distance between intra-class features (i.e., either both foregrounds or both backgrounds) in the embedding space and maximize the distance between inter-class features (i.e., foreground and background). Through this learning mechanism, the obtained feature representations effectively sharpen the classification boundaries between foreground and background regions. 

\subsubsection{Prototype Correlation Loss for Boundary-level Shape Learning}
\begin{figure*}[!htbp]
\centering
\includegraphics[width=0.92\textwidth]{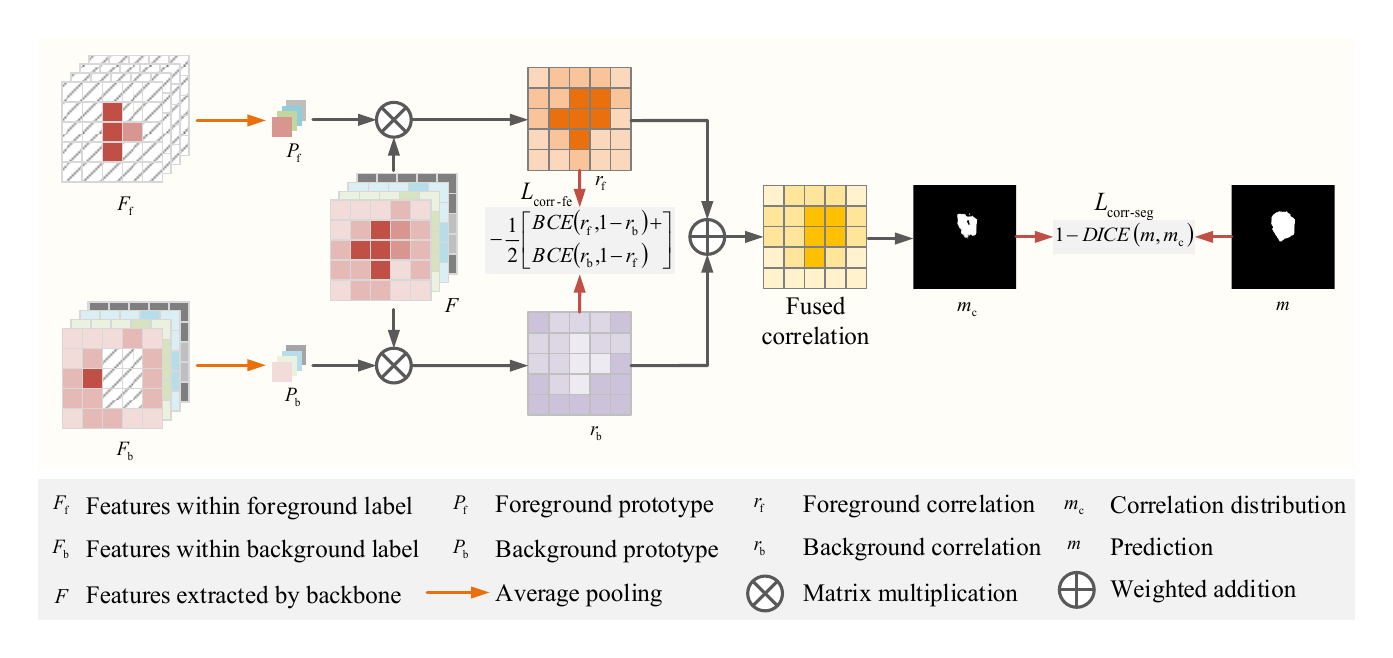}
\caption{Diagram of prototype correlation loss learning.}
\label{fig:correlation_loss}
\end{figure*}
To learn pixel-wisely precise segmentation shapes, building on the segmentation objective defined as Eq. \eqref{condition3}, we extend the shape constraint to boundary-level by introducing a prototype correlation loss, to make the uncertain pixels on the classification boundary in the feature space closer to their category prototypes. 

As illustrated in Fig.~\ref{fig:correlation_loss}, given the high-confidence foreground and background labels $\boldsymbol{X_{f}}$ and $\boldsymbol{X_{b}}$, we first extract the corresponding region features $\boldsymbol{\mathcal{F}_f}$ and $\boldsymbol{\mathcal{F}_b}$. The foreground/background prototype $\boldsymbol{P_f} \in \mathbb{R}^{C \times 1 \times 1}$ and $\boldsymbol{P_b} \in \mathbb{R}^{C \times 1 \times 1}$ is obtained by extracting deep features located in regions of the high-confidence foreground/background labels and performing average pooling, where $C$ means feature channels. 
% Then global average pooling is applied to reduce the dimensionality of these features to $C \times 1 \times 1$ to obtain the prototype $\boldsymbol{P_f}$ and $\boldsymbol{P_b}$, where $C$ means channels. 
\begin{equation}
\begin{aligned}
\boldsymbol{\mathcal{F}_f} = \boldsymbol{\mathcal{F}} \cdot \boldsymbol{X_{f}},& \boldsymbol{\mathcal{F}_b} = \boldsymbol{\mathcal{F}} \cdot \boldsymbol{X_{b}},& \\
\boldsymbol{P_f} = \frac{\boldsymbol{\mathcal{F}_f}}{\left\| \boldsymbol{\mathcal{F}_f} \right\|_2 + \epsilon},& \boldsymbol{P_b} = \frac{\boldsymbol{\mathcal{F}_b}}{\left\| \boldsymbol{\mathcal{F}_b} \right\|_2 + \epsilon}.
\label{pf_pb}
\end{aligned}
\end{equation}

This loss comprises two components: 1) complementary consistency between feature correlations that refer to high-confidence foreground prototype and background prototype, and 2) consistency between network segmentation results and fused correlation segmentation results.

Using metric learning, we evaluate the correlation response of each position in the feature map $\boldsymbol{\mathcal{F}}$ with respect to the foreground prototypes and background prototypes, obtaining the foreground correlation response $\boldsymbol{r_f}$ and the background correlation response $\boldsymbol{r_b}$ as follows: 
\begin{equation}
\begin{aligned}
\boldsymbol{r_f} = \max\left(0, \frac{\boldsymbol{\mathcal{F}}^T\cdot \boldsymbol{r_f}}{\|\boldsymbol{P_f}\|_2 + \epsilon}\right),\\
\boldsymbol{r_b} = \max\left(0, \frac{\boldsymbol{\mathcal{F}}^T\cdot \boldsymbol{r}}{\|\boldsymbol{P_b}\|_2 + \epsilon}\right).
\label{cf_cb}
\end{aligned}
\end{equation}

Since foreground and background are mutually exclusive in segmentation tasks, the foreground correlation response $\boldsymbol{\hat{r_f}}$ can be derived from the background prototype $\boldsymbol{r_b}$ as follows:
% should be consistent with the foreground correlation response \( C_f \). 
\begin{align}
\boldsymbol{\hat{r_f}} = 1 - \boldsymbol{r_b}.
\label{back_corr}
\end{align}

The correlation map represents the similarity between image features and each prototype. The first component $\mathcal{L}_\text{corr-fe}$ of the prototype correlation loss reflects the complementary consistency between the feature correlations referring to high-confidence foreground and background prototypes, which is defined as follows:
\begin{align}
\mathcal{L}_\text{corr-fe} = -\frac{1}{2}\left[ \boldsymbol{r_f} \log (\boldsymbol{\hat{r_f}})\right. 
+ \left. (1 - \boldsymbol{r_f} \log (1 - \boldsymbol{\hat{r_f}}) \right] -\frac{1}{2}\left[ \boldsymbol{\hat{r_f}} \log (\boldsymbol{r_f})\right. 
+ \left. (1 - \boldsymbol{\hat{r_f}}) \log (1 - \boldsymbol{r_f}) \right].
\label{correlation1}
\end{align}

The fused predictions $\boldsymbol{m_c}$ are obtained by balancing the foreground correlation maps $\boldsymbol{r_f}$ based on the foreground and those $\boldsymbol{\hat{r_f}}$ on the background prototypes. 
\begin{align}
\boldsymbol{m_c} = \frac{\boldsymbol{r_f} + \boldsymbol{\hat{r_f}}}{2}.
\label{fuse_correlation}
\end{align}

The second component $\mathcal{L}_\text{corr-seg}$ of the correlation loss that measures the consistency between the fused correlation map $\boldsymbol{m_c}$ and the segmentation prediction $\boldsymbol{m}$ is defined as:

\begin{align}
\mathcal{L}_\text{corr-seg} =1 - \frac{2 \times |\boldsymbol{m} \cap \boldsymbol{m_c}|}{|\boldsymbol{m}| + |\boldsymbol{m_c}|}.
\label{correlation2}
\end{align}

The total prototype correlation loss $\mathcal{L}_\text{correlation}$ is then calculated as follows:
\begin{align}
\mathcal{L}_\text{correlation} = \mathcal{L}_\text{corr-fe} + \mathcal{L}_\text{corr-seg}.
\label{correlation_loss}
\end{align}

By considering the complementary consistency of foreground and background prototype correlation, the algorithm obtained segmentation edges with low uncertainty. By directly propagating the fused segmentation boundaries to the predicted results, we achieve explicit shape learning, effectively addressing the challenge of refining the boundaries of nodule regions. 

\subsubsection{Overall Loss Function}
By combining the abovementioned losses of learning location information and shape information, the overall loss function during the training process can be expressed as:
\begin{align}
\mathcal{L}_\text{all} = \mathcal{L}_\text{alignment} + \lambda \mathcal{L}_\text{cnt} \pm \beta \mathcal{L}_\text{correlation},
\label{loss_loc_shape}
\end{align}
where $\mathcal{L}_\text{all}$ indicates the overall loss, $\mathcal{L}_\text{alignment}$ represents alignment loss. $\mathcal{L}_\text{cnt}$ represents the contrastive loss and $\mathcal{L}_\text{correlation}$ is the prototype correlation loss, $\lambda$ and $\beta$ are weighted parameter.

\section{Experiments}
\subsection{Experimental Materials}
\subsubsection{Dataset}
To evaluate the effectiveness of the proposed segmentation framework, we conduct ablation and comparative experiments on two publicly available thyroid ultrasound datasets: TN3K~\cite{gong2021tn3k} and DDTI~\cite{pedraza2015DDTI}. The TN3K dataset consists of 3,494 high-resolution thyroid nodule images, following a standardized clinical split protocol with 2,879 images designated for training and 614 for testing. For the DDTI dataset, which contains 637 thyroid ultrasound scans, the data was randomly split into training and testing sets at a 4:1 ratio. Additionally, 10-fold cross-validation was applied to address the challenge of limited data while ensuring statistical reliability.

Notably, in both datasets, the ground truth annotation of thyroid nodules in ultrasound images for metrics was performed by experienced radiologists, while the point annotation was carried out by less-experienced senior medical students to simulate a real-world weakly supervised learning scenario.

\subsubsection{Evaluation Metrics}
% We performed a quantitative comparison using four common segmentation evaluation metrics: 
% Mean Intersection over Union \textbf{(mIoU)}~\cite{ling2023dsc_iou_pr}, Dice Similarity Coefficient \textbf{(DSC)}~\cite{ling2023dsc_iou_pr, wang2022dsc_pr, zhu2024dsc_hd}, 
% Hausdorff Distance at 95th percentile \textbf{(HD95)}~\cite{zhu2024dsc_hd}, and Prediction \textbf{Precision}~\cite{ling2023dsc_iou_pr, wang2022dsc_pr}, which evaluate the segmentation's overall accuracy, precision of segmented regions, boundary matching, and the reliability of predicted positives, respectively.
We conducted a quantitative comparison using four standard segmentation evaluation metrics: Mean Intersection over Union (mIoU)~\cite{ling2023dsc_iou_pr}, Dice Similarity Coefficient (DSC)~\cite{ling2023dsc_iou_pr, wang2022dsc_pr, zhu2024dsc_hd}, Hausdorff Distance at the 95th percentile (HD95)~\cite{zhu2024dsc_hd}, and Prediction Precision~\cite{ling2023dsc_iou_pr, wang2022dsc_pr}. These metrics assess the segmentation’s overall accuracy, precision of segmented regions, boundary alignment, and the reliability of predicted positives, respectively. The definitions of each metric are as follows:
\begin{align}
&\text{mIoU} = \frac{|A \cap B|}{|A \cup B|}, \\
&\text{DSC} = \frac{2 |A \cap B|}{|A| + |B|}, \\
&\text{Precision} = \frac{|A \cap B|}{|A|},
\label{metrics}
\end{align}
where \( A \) is the predicted region and \( B \) is the ground truth region.
\begin{align}
&\text{HD95} = \min \left( \sup_{p \in A} \inf_{q \in B} \| p - q \|, \sup_{q \in B} \inf_{p \in A} \| p - q \| \right),
\label{hd95}
\end{align}
where \( A \) and \( B \) represent the predicted and ground truth boundaries, and \( \| p - q \| \) is the distance between points \( p \) and \( q \). \( \inf_{q \in B} \| p - q \| \) denotes the minimum distance from a point \( p \in A \) to the closest point \( q \in B \), while \( \sup_{p \in A} \inf_{q \in B} \| p - q \| \) represents the maximum of these minimum distances over all points in \( A \). Similarly, \( \sup_{q \in B} \inf_{p \in A} \| p - q \| \) is the maximum of the minimum distances from points in \( B \) to their closest points in \( A \). 

\subsubsection{Parameter Setting and Implementation}
In the training process, we set a learning rate of 0.0001, a batch size of 16, and trained for 100 epochs with images resized to $256 \times 256$. The network was optimized using the Adam optimizer. For comparative experiments, we adhered to the parameter configurations outlined in their respective papers. All experiments were carried out using PyTorch on an Nvidia 3090 GPU equipped with 24GB of memory.

\subsection{Ablation Study}
We conducted ablation studies to analyze the effectiveness of individual components in our framework. Table~\ref{tab:ablation_results_tn3k} and Table~\ref{tab:ablation_results_ddti} provide an overview of different design strategies (Model P and Models A to E) with different pseudo-labels (represented with subscripts 1-3), together with the quantitative assessment of each model. The segmentation results together with feature heatmaps of example images across two datasets are visualized and qualified in Fig.~\ref{fig:ablation_segmentation_result_hclabel}, Fig.~\ref{fig:ablation_heatmap_result_hclabel}, Fig.~\ref{fig:ablation_segmentation_result_hrloss} and Fig.~\ref{fig:ablation_heatmap_result_hrloss}.

\subsubsection{Effectiveness of High-Confidence Labels Generation Strategy}
\begin{table}[!htbp]
    \centering
    \caption{The precision (Mean $\pm$ STD, \%) of different labels designated as foreground/background reference prototype region. The generation strategy of Topological-based denotes generating pseudo-labels by topological geometric transformation, Prompted SAM denotes generating pseudo-labels by MedSAM using point annotation as prompts, and High-confidence f/b represents our proposed high-confidence foreground/background labels.}
    \label{tab:label_precision_transposed}
    \small
    \setlength{\tabcolsep}{5pt}
    \renewcommand{\arraystretch}{1.3}
    \begin{tabular}{cccccc}
        \toprule[1.2pt]
        \multirow{2}{*}{\textbf{Generation Strategy}} & \multirow{2}{*}{\textbf{Pseudo Label}} & \multicolumn{2}{c}{\textbf{TN3K}} & \multicolumn{2}{c}{\textbf{DDTI}} \\
        \cmidrule(r){3-4} \cmidrule(l){5-6}
        & & Foreground & Background & Foreground & Background \\
        \midrule[0.8pt]
        \multirow{2}{*}{Topological-based} & Ex-/Out-rectangle & $73.44\pm7.07$ & $99.98\pm0.51$ & $73.64\pm5.59$ & $99.98\pm0.11$ \\
        & In-Quadrilateral & $98.26\pm3.05$ & $93.59\pm8.22$ & $98.65\pm0.98$ & $93.98\pm6.51$\\
        \hline
        Prompted MedSAM & MedSAM Results& $97.11\pm4.44$ & $97.70\pm3.21$ & $95.97\pm3.66$ & $97.85\pm2.38$ \\
        \hline
        Ours & High-confidence f/b & $\textbf{99.66}\pm\textbf{1.58}$ & $\textbf{99.99}\pm\textbf{0.26}$ & $\textbf{99.79}\pm\textbf{0.88}$ & $\textbf{99.98}\pm\textbf{0.01}$ \\
        \bottomrule[1.2pt]
    \end{tabular}
\label{tab:label_precision}
\end{table}

As shown in Table~\ref{tab:label_precision}, the comparison of different pseudo-labels as supervision references in terms of precision, which is the ratio of correct pixel categories in pseudo-label pixels, showed that the proposed high-confidence labels can provide more reliable reference information. When high-confidence foreground/background labels served as supervision references to learn the foreground and background, they achieved higher precision (over 99.66\% for foreground and 99.98\% for background) than those obtained using solely topological geometric transformation (no more than 98.65\% for foreground) and MedSAM results (no more than 97.11\% for foreground and 97.85\% for background). In two datasets, our generation of high-confidence foreground and background labels achieved precision exceeding 99.66\% across domain distributions.

% \begin{table}[!htbp]
%     \centering
%     \caption{The precision (Mean $\pm$ STD, \%) of different labels' foreground or background as referrnce prototype region. High-confidence f/b represents our proposed high-confidence foreground/background labels.}
%     \label{tab:label_precision_transposed}
%     \small
%     \setlength{\tabcolsep}{5pt}
%     \renewcommand{\arraystretch}{1.3}
%     \begin{tabular}{lcccc}
%         \toprule[1.2pt]
%         \multirow{2}{*}{\textbf{Pseudo Label}} & \multicolumn{2}{c}{\textbf{TN3K}} & \multicolumn{2}{c}{\textbf{DDTI}} \\
%         \cmidrule(r){2-3} \cmidrule(l){4-5}
%         & Foreground & Background & Foreground & Background \\
%         \midrule[0.8pt]
%         Ex-/Out-rectangle & 73.44$\pm$7.07 & 99.98$\pm$0.51 & 73.64$\pm$5.59 & 99.98$\pm$0.11 \\
%         In-Quadrilateral & 98.26$\pm$3.05 & 93.59$\pm$8.22 & 98.65$\pm$0.98 & 93.98$\pm$6.51\\
%         MedSAM Results& 97.11$\pm$4.44 & 97.70$\pm$3.21 & 95.97$\pm$3.66 & 97.85$\pm$2.38 \\
%         High-confidence f/b & \textbf{99.66}$\pm$\textbf{1.58} & \textbf{99.99}$\pm$\textbf{0.26} & \textbf{99.79}$\pm$\textbf{0.88} & \textbf{99.98}$\pm$\textbf{0.01} \\
%         \bottomrule[1.2pt]
%     \end{tabular}
% \label{tab:label_precision}
% \end{table}

The region reference precision of high-confidence labels is parallel to 100\% because we combine geometric topological transformation with topology priors and MedSAM prediction with anatomical information, further reducing areas of uncertainty. Therefore, they can serve as foreground and background reference $\boldsymbol{X_f}$ and $\boldsymbol{X_b}$ without misleading network learning, as outlined in Sec. \ref{sec:objective}
% Therefore, they can provide a region reference with the parallel accuracy as the true value.
% be used as precise foreground and background reference $\boldsymbol{X_f}$ and $\boldsymbol{X_b}$, as outlined in~\ref{sec:objective}.

% During the process of learning the alignment loss with different labels as references, we demonstrated that the higher precision of high-confidence labels leads to better performance metrics.
% using prompted MedSAM results as pixel-wise supervision (DSC: 73.48\% and HD95: 21.92 pixels on TN3K, DSC: 68.87\% and HD95: 27.25 on DDTI) outperformed topological and high-confidence labels when adopting pixel-to-pixel loss learning strategies. However, 
% there was still a gap compared to the results obtained by high-confidence labels training with the proposed learning strategies (DSC: 79.10\% and HD95: 17.20 pixels on TN3K, DSC: 74.55\% and HD95: 23.82 pixels on DDTI). Moreover, using our high-confidence labels illustrated higher mIoU and DSC than MedSAM results and topological geometry labels in every ablation result.

\begin{table*}[!htbp]
\centering
\caption{Quantitative ablation results (Mean $\pm$ STD) of conducting models with different pseudo-label generation and network learning strategies on the TN3K Dataset. The backbone we used is U-Net. T denotes using quadrilateral regions as foreground label regions and out-of-rectangle regions as background label regions. M and H denote using prompt-generated results from the MedSAM as labels and proposed high-confidence labels, respectively. The $P_n$ indicates the model using corresponding pseudo-label following pixel-to-pixel learning strategy.}
\small
\setlength{\tabcolsep}{4pt}
\renewcommand{\arraystretch}{1.3}
\begin{tabular}{ccccccccc}
	\toprule[1.2pt]
	\multirow{2}{*}{Model} & \multirow{2}{*}{Labels}& \multicolumn{3}{c}{Losses} & \multicolumn{4}{c}{Metrics}\\
	\cmidrule(r){3-5} \cmidrule(l){6-9}
	& & $\mathcal{L}_{\text{alignment}}$ & $\mathcal{L}_{\text{cnt}}$ & $\mathcal{L}_{\text{correlation}}$ & mIoU (\%) $\uparrow$ & DSC (\%) $\uparrow$ & Precision (\%) $\uparrow$ & HD95 (px) $\downarrow$ \\
	\midrule[0.8pt]
	$P_1$ & T 
	& - & - & - & $54.44\pm21.25$ & $67.47\pm22.30$ & $78.20\pm29.05$  & $29.42\pm30.81$ \\
	%\cmidrule(lr){2-9}
	$A_1$ & T & \checkmark & & & $62.38\pm23.41$ & $73.47\pm23.33$ & $76.79\pm25.17$ & $22.77\pm29.49$\\
	$B_1$ & T & \checkmark & \checkmark & & $66.32\pm22.02$ & $77.04\pm20.83$ & $78.73\pm22.18$ & $20.60\pm29.65$\\
	$C_1$ & T & \checkmark & & \checkmark & $66.89\pm22.65$ & $77.28\pm21.63$ & $77.21\pm23.52$ & $20.07\pm28.13$\\
	$D_1$ & T & & \checkmark & \checkmark & $52.76\pm25.05$ & $65.11\pm24.58$ & $59.44\pm29.14$ & $30.87\pm32.74$\\
	$E_1$ & T & \checkmark & \checkmark & \checkmark & $\textbf{67.50}\pm\textbf{23.75}$ & $\textbf{77.50}\pm\textbf{22.24}$ & $\textbf{80.57}\pm\textbf{23.85}$ & $\textbf{18.91}\pm\textbf{27.22}$\\
	\midrule[0.8pt]
	%\multirow{6}{*}{\rotatebox[origin=c]{90}{MedSAM results}}
	% & MedSAM-pixel & & & & 59.43$\pm$26.23 & 5.80$\pm$2.08 & 70.41$\pm$25.63 & 77.36$\pm$27.39 \\
	$P_2$ & M & - & - & - & $62.39\pm23.83$ & $73.48\pm23.33$ & $76.79\pm25.17$ & $21.92\pm31.13$ \\
	%\cmidrule(lr){2-9}
	$A_2$ & M & \checkmark & & & $60.78\pm23.47$ & $72.16\pm22.57$ & $76.10\pm23.76$ & $22.19\pm28.27$\\
	$B_2$ & M & \checkmark & \checkmark & & $66.17\pm22.76$ & $76.72\pm21.80$ & $76.77\pm23.24$ & $20.04\pm28.05$\\
	$C_2$ & M & \checkmark & & \checkmark & $65.54\pm22.52$ & $76.31\pm21.50$ & $74.34\pm23.63$ & $19.91\pm27.65$\\
	$D_2$ & M & & \checkmark & \checkmark & $53.78\pm24.74$ & $66.09\pm24.32$ & $59.43\pm26.87$ & $28.27\pm31.41$\\
	$E_2$ & M & \checkmark & \checkmark & \checkmark & $\textbf{67.25}\pm\textbf{22.71}$ & $\textbf{77.51}\pm\textbf{21.82}$ & $\textbf{77.37}\pm\textbf{23.24}$ & $\textbf{17.94}\pm\textbf{27.13}$\\
	\midrule[0.8pt]
	%\multirow{6}{*}{\rotatebox[origin=c]{90}{High-confidence}} 
	$P_3$ & H & - & - & - & $52.92\pm21.14$ & $66.18\pm22.22$ & $80.27\pm26.50$ & $26.17\pm26.97$ \\
	%\cmidrule(lr){2-9}
	$A_3$ & H & \checkmark & & & $62.72\pm23.42$ & $73.91\pm22.49$ & $79.42\pm23.30$  & $21.72\pm29.60$\\
	$B_3$ & H & \checkmark & \checkmark & & $67.49\pm22.49$ & $77.67\pm21.58$ & $78.73\pm22.38$ & $20.05\pm29.72$ \\
	$C_3$ & H & \checkmark & & \checkmark & $68.73\pm23.68$ & $78.35\pm22.59$ & $79.24\pm23.80$ & $18.09\pm28.51$\\
	$D_3$ & H & & \checkmark & \checkmark & $61.59\pm23.50$ & $73.00\pm22.55$ & $73.32\pm27.27$ & $26.83\pm32.13$\\
	$E_3$ & H & \checkmark & \checkmark & \checkmark & $\textbf{69.30}\pm\textbf{22.38}$ & $\textbf{79.10}\pm\textbf{21.37}$ & $\textbf{82.49}\pm\textbf{22.69}$ & $\textbf{17.20}\pm\textbf{26.97}$\\
	\bottomrule[1.2pt]
\end{tabular}
\label{tab:ablation_results_tn3k}
\end{table*}

\begin{table*}[!htbp]
\centering
\caption{Quantitative ablation results (Mean $\pm$ STD) of conducting models with different pseudo-label generation and network learning strategies on the DDTI Dataset. The backbone we used is U-Net. T denotes using quadrilateral regions as foreground label regions and out-of-rectangle regions as background label regions. M and H denote using prompt-generated results from the MedSAM as labels and proposed high-confidence labels, respectively. The $P_n$ indicates the model using corresponding pseudo-label following pixel-to-pixel learning strategy.}
\small
\setlength{\tabcolsep}{4pt}
\renewcommand{\arraystretch}{1.3}
\begin{tabular}{ccccccccc}
\toprule[1.2pt]
\multirow{2}{*}{Model} & \multirow{2}{*}{Labels} & \multicolumn{3}{c}{Losses} & \multicolumn{4}{c}{Metrics}\\
\cmidrule(r){3-5} \cmidrule(l){6-9}
& & $\mathcal{L}_{\text{alignment}}$ & $\mathcal{L}_{\text{cnt}}$ & $\mathcal{L}_{\text{correlation}}$ & mIoU (\%) $\uparrow$  & DSC (\%) $\uparrow$ & Precision (\%) $\uparrow$ & HD95 (px) $\downarrow$\\
\midrule[0.8pt]
%\multirow{6}{*}{\rotatebox[origin=c]{90}{Topological-based}} 
$P_1$ & T & & & & $44.85\pm20.77$ & $58.76\pm22.36$ & $65.65\pm31.36$ & $34.49\pm20.88$ \\
%\cmidrule(lr){2-9}
$A_1$ & T & \checkmark & & & $53.34\pm23.05$ & $66.21\pm22.60$ & $64.50\pm27.54$ & $30.65\pm24.81$\\
$B_1$ & T & \checkmark & \checkmark & & $59.31\pm21.95$ & $71.68\pm20.55$ & $70.77\pm25.68$ & $26.85\pm25.05$\\
$C_1$ & T & \checkmark & & \checkmark & $59.70\pm22.63$ & $71.78\pm21.34$ & $70.81\pm25.57$ & $26.39\pm24.89$ \\
$D_1$ & T & & \checkmark & \checkmark & $52.11\pm23.25$ & $65.09\pm22.59$ & $61.48\pm28.58$ & $32.57\pm24.94$ \\
$E_1$ & T & \checkmark & \checkmark & \checkmark & $\textbf{60.20}\pm\textbf{22.20}$ & $\textbf{72.31}\pm\textbf{20.77}$ & $\textbf{72.66}\pm\textbf{25.86}$ & $\textbf{25.84}\pm\textbf{25.79}$\\
\midrule[0.8pt]
%\multirow{6}{*}{\rotatebox[origin=c]{90}{MedSAM results}}
$P_2$ & M & & & & $56.54\pm23.52$ & $68.87\pm22.62$ & $70.10\pm28.14$ & $27.25\pm25.68$\\
%\cmidrule(lr){2-9}
$A_2$ & M & \checkmark & & & $51.52\pm22.12$ & $65.58\pm21.64$ & $63.98\pm26.04$  & $31.07\pm27.62$ \\
$B_2$ & M & \checkmark & \checkmark & & $58.25\pm22.25$ & $70.68\pm20.15$ & $72.35\pm24.84$ & $26.49\pm25.33$ \\
$C_2$ & M & \checkmark & & \checkmark & $58.84\pm22.46$ & $71.05\pm20.74$ & $71.60\pm26.08$ & $25.93\pm24.63$\\
$D_2$ & M & & \checkmark & \checkmark & $49.84\pm23.16$ & $63.04\pm22.78$ & $58.67\pm27.33$ & $33.91\pm23.60$\\
$E_2$ & M & \checkmark & \checkmark & \checkmark & $\textbf{59.82}\pm\textbf{21.98}$ 
& $\textbf{72.14}\pm\textbf{20.76}$ & $\textbf{72.96}\pm\textbf{24.60}$ & $\textbf{25.40}\pm\textbf{25.57}$\\
\midrule[0.8pt]
%\multirow{6}{*}{\rotatebox[origin=c]{90}{High-confidence}} 
$P_3$ & H & & & & $47.31\pm20.41$ & $61.27\pm21.59$ & $74.34\pm29.85$ & $31.92\pm23.87$ \\
%\cmidrule(lr){2-9}
$A_3$ & H & \checkmark & & & $53.46\pm24.04$ & $66.06\pm23.37$ & $69.12\pm25.74$ & $30.70\pm26.22$ \\
$B_3$ & H & \checkmark & \checkmark & & $60.83\pm23.51$ & $72.43\pm22.30$ & $72.75\pm23.38$ & $24.69\pm25.22$ \\
$C_3$ & H &\checkmark & & \checkmark & $61.42\pm22.53$ & $73.18\pm21.22$ & $72.29\pm24.90$ & $25.44\pm26.83$ \\
$D_3$ & H & & \checkmark & \checkmark & $56.29\pm21.70$ & $69.19\pm20.83$ & $71.97\pm26.90$ & $28.13\pm24.83$ \\
$E_3$ & H & \checkmark & \checkmark & \checkmark & $\textbf{62.52}\pm\textbf{20.72}$ & $\textbf{74.55}\pm\textbf{18.91}$ & $\textbf{74.66}\pm\textbf{24.62}$ & $\textbf{23.82}\pm\textbf{24.77}$\\
\bottomrule[1.2pt]
\end{tabular}
\label{tab:ablation_results_ddti}
\end{table*}

%\begin{figure*}[!htbp]
%\centering
%\includegraphics[width=0.98\textwidth]{ablation_all.png}
%\caption{\textbf{Heatmap results of ablation experiments.} Topological-x, MedSAM-x, and HC-x denote the use of initial topological shapes, prompt-generated results from the MedSAM, and high-confidence labels for the model training, respectively. The X-pixel denotes the learning strategy of using pseudo-labels for pixel-to-pixel comparison. The x-A to x-D indicate the ablation models defined in Table \ref{tab:ablation_results_tn3k} and \ref{tab:ablation_results_ddti}.}
%\label{fig:ablation_result}
%\end{figure*}

\begin{figure*}[!htbp]
	\footnotesize
	\centering
	\tabcolsep=0.5mm
	\begin{tabular}{cccccccc}
		\vspace{1mm}
		\adjustbox{valign=m}{\includegraphics[width=0.1\textwidth]{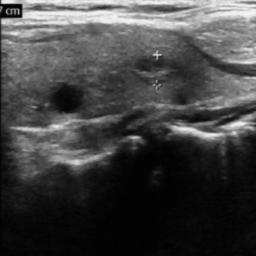}}&
		\adjustbox{valign=m}{\includegraphics[width=0.1\textwidth]{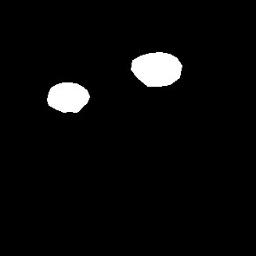}}
		&
		\adjustbox{valign=m}{\includegraphics[width=0.1\textwidth]{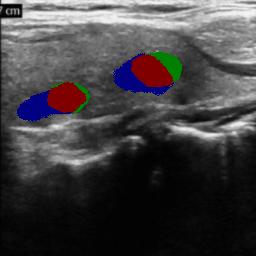}}
		&
		\adjustbox{valign=m}{\includegraphics[width=0.1\textwidth]{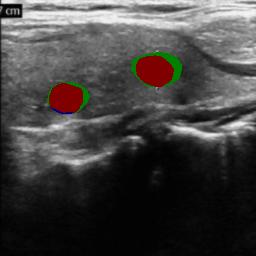}}
		&
		\adjustbox{valign=m}{\includegraphics[width=0.1\textwidth]{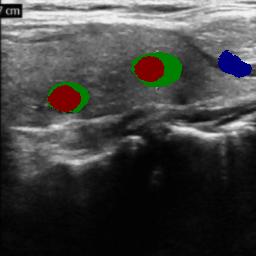}}
		&
		\adjustbox{valign=m}{\includegraphics[width=0.1\textwidth]{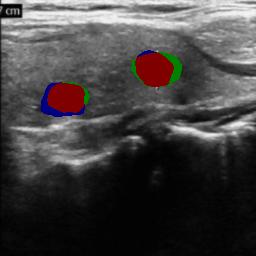}}
		&
		\adjustbox{valign=m}{\includegraphics[width=0.1\textwidth]{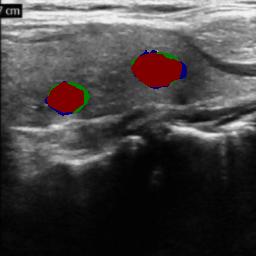}}
		&
		\adjustbox{valign=m}{\includegraphics[width=0.1\textwidth]{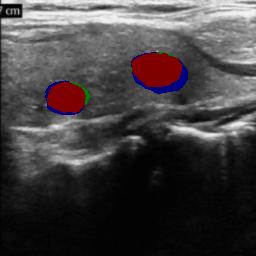}}
		\\
		\vspace{1mm}
		\adjustbox{valign=m}{\includegraphics[width=0.1\textwidth]{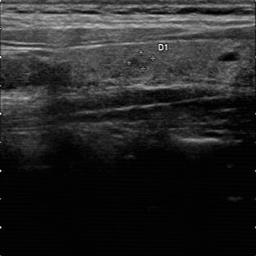}}&
		\adjustbox{valign=m}{\includegraphics[width=0.1\textwidth]{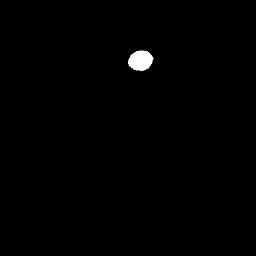}}
		&
		\adjustbox{valign=m}{\includegraphics[width=0.1\textwidth]{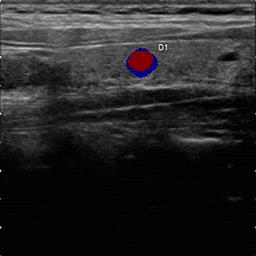}}
		&
		\adjustbox{valign=m}{\includegraphics[width=0.1\textwidth]{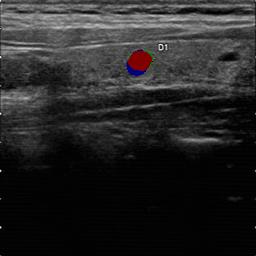}}
		&
		\adjustbox{valign=m}{\includegraphics[width=0.1\textwidth]{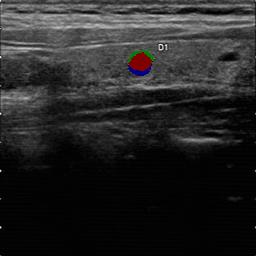}}
		&
		\adjustbox{valign=m}{\includegraphics[width=0.1\textwidth]{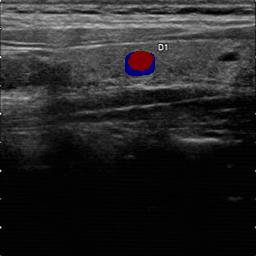}}
		&
		\adjustbox{valign=m}{\includegraphics[width=0.1\textwidth]{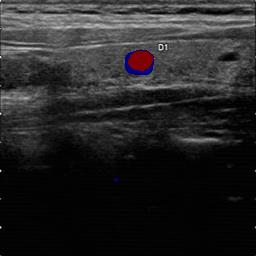}}
		&
		\adjustbox{valign=m}{\includegraphics[width=0.1\textwidth]{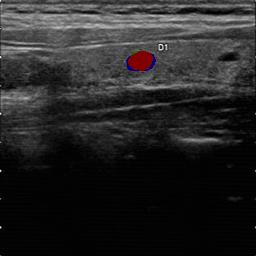}}
		\\
		\vspace{1mm}
		\adjustbox{valign=m}{\includegraphics[width=0.1\textwidth]{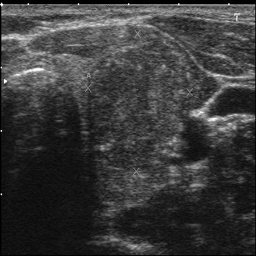}}
		&
		\adjustbox{valign=m}{\includegraphics[width=0.1\textwidth]{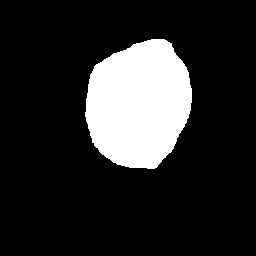}}
		&
		\adjustbox{valign=m}{\includegraphics[width=0.1\textwidth]{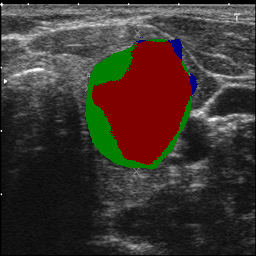}}
		&
		\adjustbox{valign=m}{\includegraphics[width=0.1\textwidth]{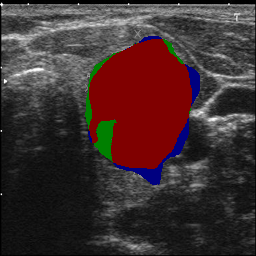}}
		&
		\adjustbox{valign=m}{\includegraphics[width=0.1\textwidth]{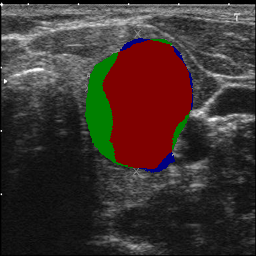}}
		&
		\adjustbox{valign=m}{\includegraphics[width=0.1\textwidth]{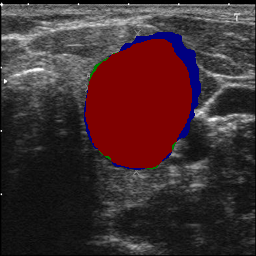}}
		&
		\adjustbox{valign=m}{\includegraphics[width=0.1\textwidth]{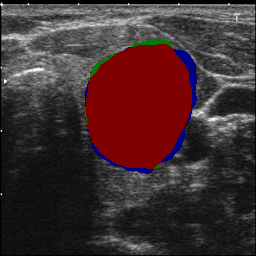}}
		&
		\adjustbox{valign=m}{\includegraphics[width=0.1\textwidth]{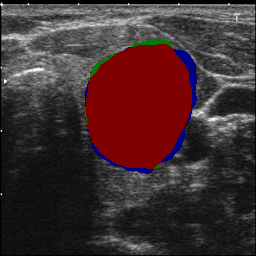}}
		\\
		\vspace{1mm}
		\adjustbox{valign=m}{\includegraphics[width=0.1\textwidth]{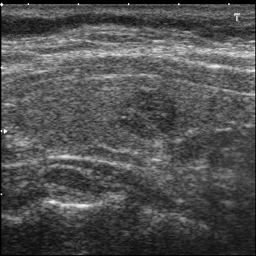}}&
		\adjustbox{valign=m}{\includegraphics[width=0.1\textwidth]{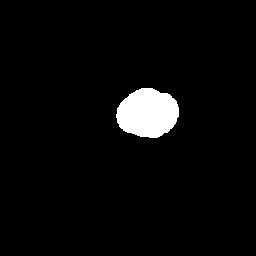}}
		&
		\adjustbox{valign=m}{\includegraphics[width=0.1\textwidth]{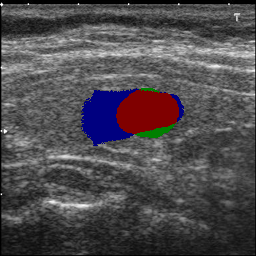}}
		&
		\adjustbox{valign=m}{\includegraphics[width=0.1\textwidth]{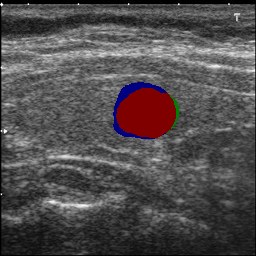}}
		&
		\adjustbox{valign=m}{\includegraphics[width=0.1\textwidth]{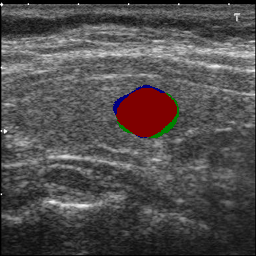}}
		&
		\adjustbox{valign=m}{\includegraphics[width=0.1\textwidth]{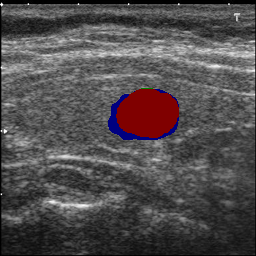}}
		&
		\adjustbox{valign=m}{\includegraphics[width=0.1\textwidth]{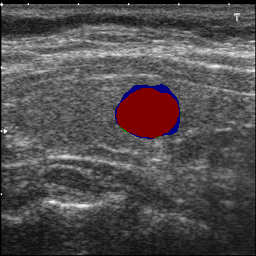}}
		&
		\adjustbox{valign=m}{\includegraphics[width=0.1\textwidth]{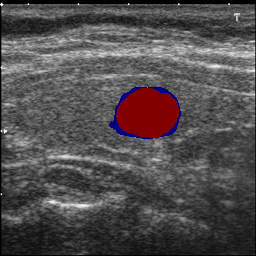}}
		\\
		Images & GT & $P_{1}$ & $P_{2}$ & $P_{3}$ & $E_{1}$ & $E_{2}$ & $E_{3}$
		\\	
	\end{tabular}
	\caption{\textbf{Segmentation results of using diiferent weak labels as training references.} $P_{1}$, $P_{2}$ and $P_{3}$ are models using the topological transformations, prompted MedSAM results and our proposed high-confidence pseudo labels discussed in Table~\ref{tab:label_precision} as training references following the pixel-to-pixel fully supervised manner, while $E_{1}$, $E_{2}$ and $E_{3}$ denote models using those labels as training references following the proposed high-rationality weakly supervised manner. Row 1 and 2 are example results from the TN3K dataset, while Row 3 and 4 are example results from the DDTI dataset. Red indicates correct thyroid nodule predictions, green represents the under-segmentations of thyroid nodules, and blue shows an over-segmentations of other tissues as thyroid nodules.}
	\label{fig:ablation_segmentation_result_hclabel}
\end{figure*}

\begin{figure*}[!htbp]
	\footnotesize
	\centering
	\tabcolsep=0.5mm
	\begin{tabular}{cccccccc}
		\vspace{1mm}
		\adjustbox{valign=m}{\includegraphics[width=0.1\textwidth]{img_0001_resize.png}}&
		\adjustbox{valign=m}{\includegraphics[width=0.1\textwidth]{gt_0001.jpg}}
		&
		\adjustbox{valign=m}{\includegraphics[width=0.1\textwidth]{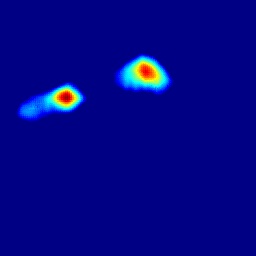}}
		&
		\adjustbox{valign=m}{\includegraphics[width=0.1\textwidth]{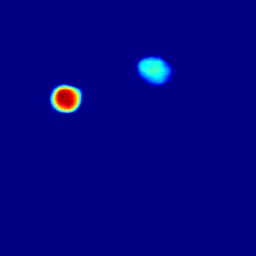}}
		&
		\adjustbox{valign=m}{\includegraphics[width=0.1\textwidth]{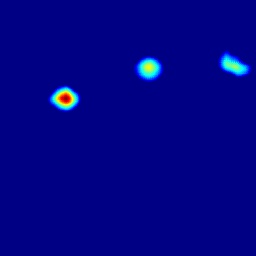}}
		&
		\adjustbox{valign=m}{\includegraphics[width=0.1\textwidth]{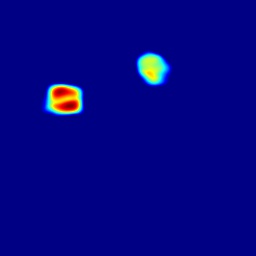}}
		&
		\adjustbox{valign=m}{\includegraphics[width=0.1\textwidth]{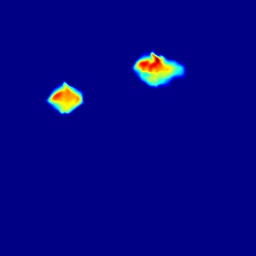}}
		&
		\adjustbox{valign=m}{\includegraphics[width=0.1\textwidth]{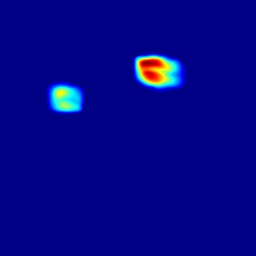}}
		\\
		\vspace{1mm}
		\adjustbox{valign=m}{\includegraphics[width=0.1\textwidth]{img_0010_resize.png}}&
		\adjustbox{valign=m}{\includegraphics[width=0.1\textwidth]{gt_0010.jpg}}
		&
		\adjustbox{valign=m}{\includegraphics[width=0.1\textwidth]{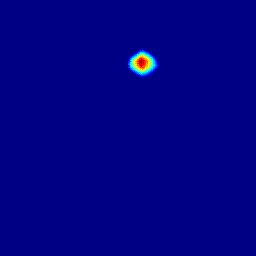}}
		&
		\adjustbox{valign=m}{\includegraphics[width=0.1\textwidth]{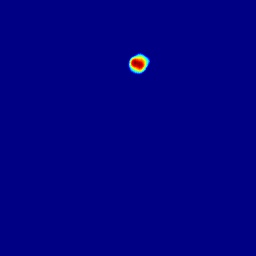}}
		&
		\adjustbox{valign=m}{\includegraphics[width=0.1\textwidth]{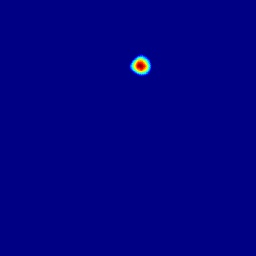}}
		&
		\adjustbox{valign=m}{\includegraphics[width=0.1\textwidth]{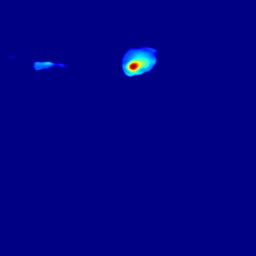}}
		&
		\adjustbox{valign=m}{\includegraphics[width=0.1\textwidth]{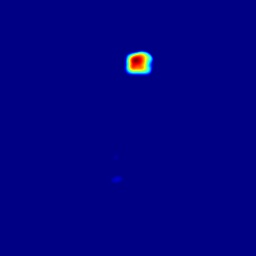}}
		&
		\adjustbox{valign=m}{\includegraphics[width=0.1\textwidth]{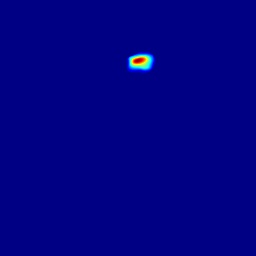}}
		\\
		\vspace{1mm}
		\adjustbox{valign=m}{\includegraphics[width=0.1\textwidth]{images_101.PNG}}
		&
		\adjustbox{valign=m}{\includegraphics[width=0.1\textwidth]{gt_101.PNG}}
		&
		\adjustbox{valign=m}{\includegraphics[width=0.1\textwidth]{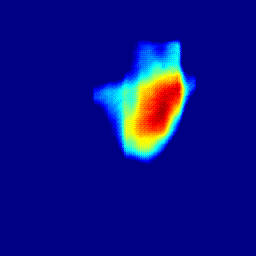}}
		&
		\adjustbox{valign=m}{\includegraphics[width=0.1\textwidth]{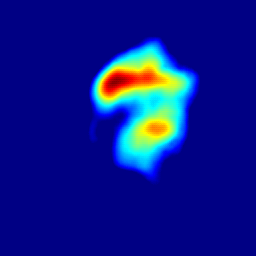}}
		&
		\adjustbox{valign=m}{\includegraphics[width=0.1\textwidth]{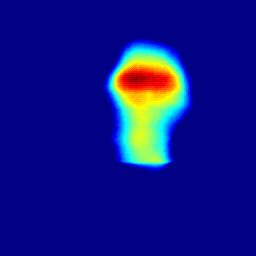}}
		&
		\adjustbox{valign=m}{\includegraphics[width=0.1\textwidth]{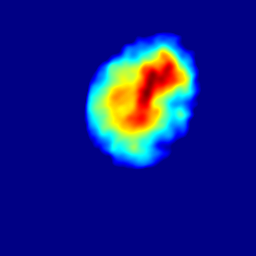}}
		&
		\adjustbox{valign=m}{\includegraphics[width=0.1\textwidth]{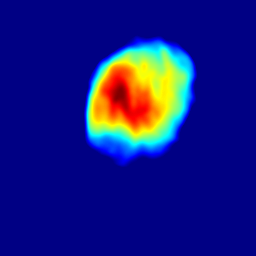}}
		&
		\adjustbox{valign=m}{\includegraphics[width=0.1\textwidth]{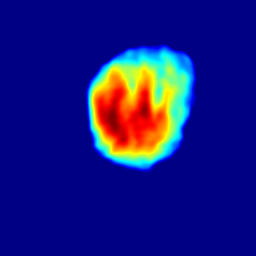}}
		\\
		\vspace{1mm}
		\adjustbox{valign=m}{\includegraphics[width=0.1\textwidth]{images_128.PNG}}&
		\adjustbox{valign=m}{\includegraphics[width=0.1\textwidth]{gt_128.PNG}}
		&
		\adjustbox{valign=m}{\includegraphics[width=0.1\textwidth]{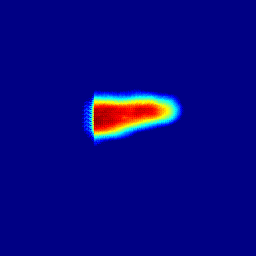}}
		&
		\adjustbox{valign=m}{\includegraphics[width=0.1\textwidth]{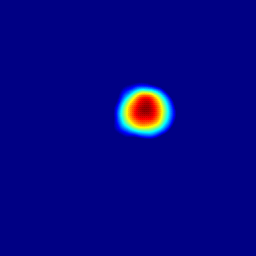}}
		&
		\adjustbox{valign=m}{\includegraphics[width=0.1\textwidth]{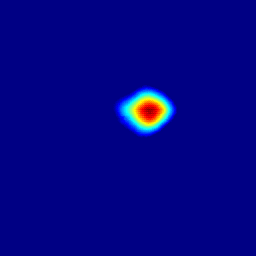}}
		&
		\adjustbox{valign=m}{\includegraphics[width=0.1\textwidth]{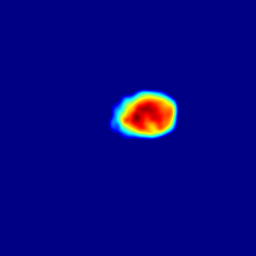}}
		&
		\adjustbox{valign=m}{\includegraphics[width=0.1\textwidth]{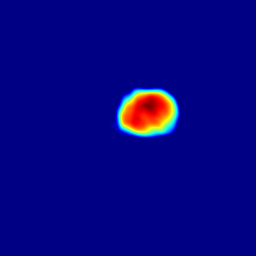}}
		&
		\adjustbox{valign=m}{\includegraphics[width=0.1\textwidth]{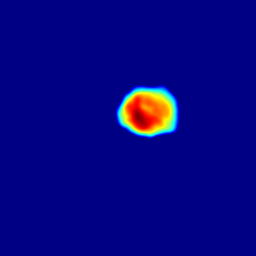}}
		\\
		Images & GT & $P_{1}$ & $P_{2}$ & $P_{3}$ & $E_{1}$ & $E_{2}$ & $E_{3}$
		\\	
	\end{tabular}
	\caption{\textbf{Feature map visualizations of segmentation results in Fig.~\ref{fig:ablation_segmentation_result_hclabel}.} $P_{1}$, $P_{2}$, $P_{3}$, $E_{1}$, $E_{2}$ and $E_{3}$ denote the same models in Fig.~\ref{fig:ablation_segmentation_result_hclabel}. Row 1 and 2 are example results from the TN3K dataset, while Row 3 and 4 are example results from the DDTI dataset.}
	\label{fig:ablation_heatmap_result_hclabel}
\end{figure*}

\subsubsection{Effectiveness of High-confidence Labels for Network Training}
%The effectiveness of high-confidence labels was proven by the outperformance of our method when using high-confidence labels as supervision references. 
The effectiveness of weak labels was firstly evaluated by using them to train the network under the pixel-to-pixel fully supervised manner. As illustrated in Fig.~\ref{fig:ablation_segmentation_result_hclabel} and Fig.~\ref{fig:ablation_heatmap_result_hclabel}, by comparing Model $P_1$, $P_2$ and $P_3$, it could be observed that Model $P_2$ with MedSAM results as labels distinguished the nodule features clearer and segmented the nodule regions more accurate. Model $P_3$ with our proposed high-confidence pseudo labels didn't show obvious advantages to the Model $P_1$ using the topological transformations of point annotations as labels. The quantitative results in Tabel~\ref{tab:ablation_results_tn3k} and Table~\ref{tab:ablation_results_ddti} confirmed our qualitative observations. It could be attributed to that the total amount of information provided by the high-confidence label was less than that of the MedSAM results. However, directly learning with a fully supervised manner led to the network only focusing on the ``deterministic'' information represented by the label region, ignoring the equally meaningful information in the uncertain regions and resulting in under-segmentation. 

In contrast, when replacing the learning strategy with our proposed weakly supervised losses, by comparing Model $E_1$, $E_2$ and $E_3$ in Fig.~\ref{fig:ablation_segmentation_result_hclabel} and Fig.~\ref{fig:ablation_heatmap_result_hclabel}, it was observed that Model $E_3$ not only provided more accurate feature extraction and nodule segmentation than Model $E_1$ and $E_2$, but also outperformed Model $P_2$ by providing much less under-segmentation of nodules and over-segmentation of other tissues. Moreover, as shown in Table~\ref{tab:ablation_results_tn3k} and Table~\ref{tab:ablation_results_ddti}, Model $A_3$ to $E_3$ provided generally higher segmentation metrics than the corresponding Model $A_1$ to $E_1$ and Model $A_2$ and $E_2$. It indicates that the high-confidence information provided by the proposed high-confidence labels can enable the network to accurately capture the localization, feature clustering, and prototype differences between foreground and background during weakly supervised learning, thereby improving segmentation performance, even better than following the fully supervised manner.

Considering the performance of model with every combination of weakly supervised losses was improved by the proposed high-confidence labels (Model $A_3$ vs. $A_1$ vs. $A_2$, Model $B_3$ vs. $B_1$ vs. $B_2$, Model $C_3$ vs. $C_1$ vs. $C_2$, Model $D_3$ vs. $D_1$ vs. $D_2$, Model $E_3$ vs. $E_1$ vs. $E_2$), we focused on discussing the performance of Model $P_3$, $A_3$, $B_3$, $C_3$, $D_3$ and $E_3$ to evaluate the effectiveness of different losses in the following sections.

\subsubsection{Effectiveness of Alignment Loss for Location Learning}
\begin{figure*}[!htbp]
\footnotesize
\centering
\tabcolsep=0.5mm
	\begin{tabular}{cccccccccc}
		\vspace{1mm}
		\adjustbox{valign=m}{\includegraphics[width=0.09\textwidth]{img_0001_resize.png}}
		&
		\adjustbox{valign=m}{\includegraphics[width=0.09\textwidth]{gt_0001.jpg}}
		&
		\adjustbox{valign=m}{\includegraphics[width=0.09\textwidth]{medsam_0001.jpg}}
		&
		\adjustbox{valign=m}{\includegraphics[width=0.09\textwidth]{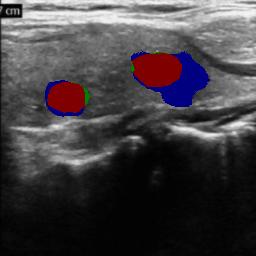}}
		&
		\adjustbox{valign=m}{\includegraphics[width=0.09\textwidth]{hc_0001.jpg}}
		&
		\adjustbox{valign=m}{\includegraphics[width=0.09\textwidth]{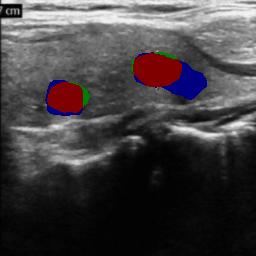}}
		&
		\adjustbox{valign=m}{\includegraphics[width=0.09\textwidth]{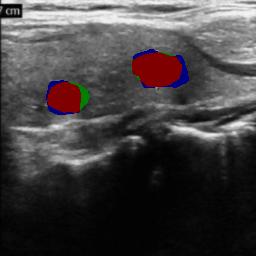}}
		&
		\adjustbox{valign=m}{\includegraphics[width=0.09\textwidth]{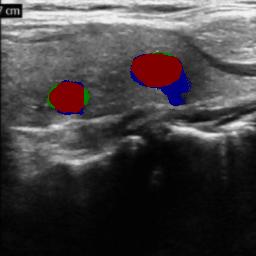}}
		&
		\adjustbox{valign=m}{\includegraphics[width=0.09\textwidth]{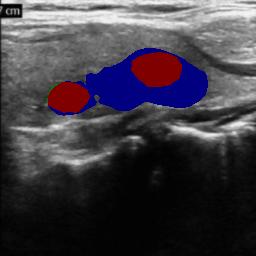}}
		&
		\adjustbox{valign=m}{\includegraphics[width=0.09\textwidth]{hc_proj_cnt_proto_0001.jpg}}	
		\\
		\vspace{1mm}
		\adjustbox{valign=m}{\includegraphics[width=0.09\textwidth]{img_0010_resize.png}}
		&
		\adjustbox{valign=m}{\includegraphics[width=0.09\textwidth]{gt_0010.jpg}}
		&
		\adjustbox{valign=m}{\includegraphics[width=0.09\textwidth]{medsam_0010.jpg}}
		&
		\adjustbox{valign=m}{\includegraphics[width=0.09\textwidth]{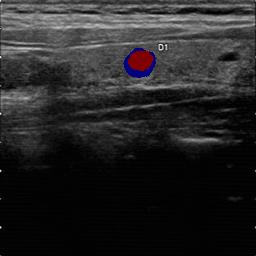}}
		&
		\adjustbox{valign=m}{\includegraphics[width=0.09\textwidth]{hc_0010.jpg}}
		&
		\adjustbox{valign=m}{\includegraphics[width=0.09\textwidth]{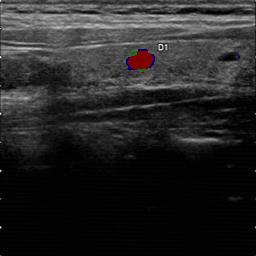}}
		&
		\adjustbox{valign=m}{\includegraphics[width=0.09\textwidth]{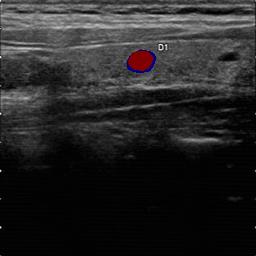}}
		&
		\adjustbox{valign=m}{\includegraphics[width=0.09\textwidth]{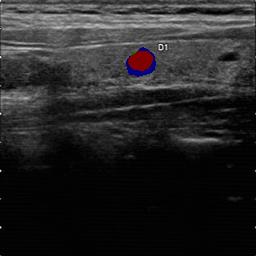}}
		&
		\adjustbox{valign=m}{\includegraphics[width=0.09\textwidth]{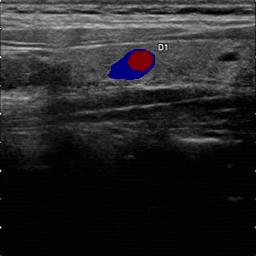}}
		&
		\adjustbox{valign=m}{\includegraphics[width=0.09\textwidth]{hc_proj_cnt_proto_0010.jpg}}	
		\\
		\vspace{1mm}
		\adjustbox{valign=m}{\includegraphics[width=0.09\textwidth]{images_101.PNG}}
		&
		\adjustbox{valign=m}{\includegraphics[width=0.09\textwidth]{gt_101.PNG}}
		&
		\adjustbox{valign=m}{\includegraphics[width=0.09\textwidth]{medsam_101.PNG}}
		&
		\adjustbox{valign=m}{\includegraphics[width=0.09\textwidth]{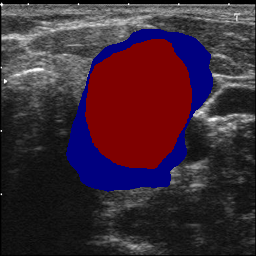}}
		&
		\adjustbox{valign=m}{\includegraphics[width=0.09\textwidth]{hc_101.PNG}}
		&
		\adjustbox{valign=m}{\includegraphics[width=0.09\textwidth]{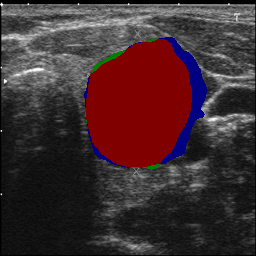}}
		&
		\adjustbox{valign=m}{\includegraphics[width=0.09\textwidth]{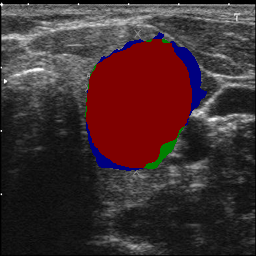}}
		&
		\adjustbox{valign=m}{\includegraphics[width=0.09\textwidth]{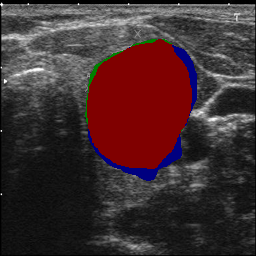}}
		&
		\adjustbox{valign=m}{\includegraphics[width=0.09\textwidth]{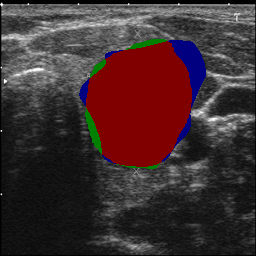}}
		&
		\adjustbox{valign=m}{\includegraphics[width=0.09\textwidth]{hc_proj_cnt_proto_101.PNG}}
		\\
		\vspace{1mm}
		\adjustbox{valign=m}{\includegraphics[width=0.09\textwidth]{images_128.PNG}}
		&
		\adjustbox{valign=m}{\includegraphics[width=0.09\textwidth]{gt_128.PNG}}
		&
		\adjustbox{valign=m}{\includegraphics[width=0.09\textwidth]{medsam_128.PNG}}
		&
		\adjustbox{valign=m}{\includegraphics[width=0.09\textwidth]{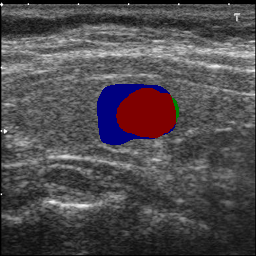}}
		&
		\adjustbox{valign=m}{\includegraphics[width=0.09\textwidth]{hc_128.PNG}}
		&
		\adjustbox{valign=m}{\includegraphics[width=0.09\textwidth]{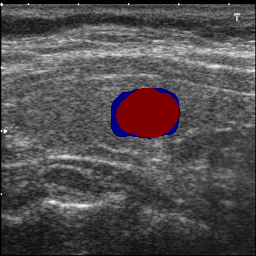}}
		&
		\adjustbox{valign=m}{\includegraphics[width=0.09\textwidth]{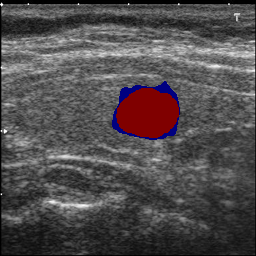}}
		&
		\adjustbox{valign=m}{\includegraphics[width=0.09\textwidth]{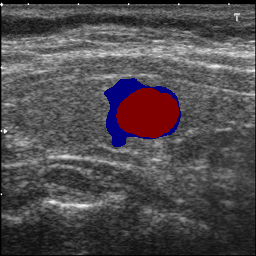}}
		&
		\adjustbox{valign=m}{\includegraphics[width=0.09\textwidth]{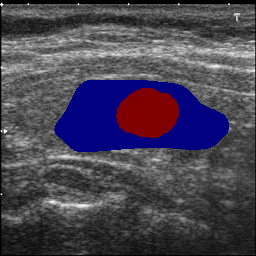}}
		&
		\adjustbox{valign=m}{\includegraphics[width=0.09\textwidth]{hc_proj_cnt_proto_128.PNG}}
		\\
		Images & GT & $P_{2}$ & $A_{2}$ & $P_{3}$ & $A_{3}$	& $B_{3}$ & $C_{3}$	& $D_{3}$ & $E_{3}$
		\\	
	\end{tabular}
\caption{\textbf{Segmentation results of using diiferent loss functions as learning strategies under the high-confidence pseudo labels.} $P_{2}$ denotes the model using the MedSAM results as training labels and following the pixel-to-pixel fully supervised manner. $A_{2}$ denotes the model using the same labels but using the alignment loss only as the weakly supervised learning. $P_{3}$ denotes the model using the high-confidence pseudo labels discussed in Table~\ref{tab:label_precision} as training references following the pixel-to-pixel fully supervised manner, while $A_{3}$, $B_{3}$, $C_{3}$, $D_{3}$ and $E_{3}$ denote models using high-confidence pseudo labels as training references following different combinations of loss functions in Table~\ref{tab:ablation_results_tn3k} and Table~\ref{tab:ablation_results_ddti}. Example images are the same as those in Fig.~\ref{fig:ablation_segmentation_result_hclabel}. Red indicates correct thyroid nodule predictions, green represents the under-segmentations of thyroid nodules, and blue shows an over-segmentations of other tissues as thyroid nodules.}
\label{fig:ablation_segmentation_result_hrloss}
\end{figure*}

\begin{figure*}[!htbp]
	\footnotesize
	\centering
	\tabcolsep=0.5mm
	\begin{tabular}{cccccccccc}
		\vspace{1mm}
		\adjustbox{valign=m}{\includegraphics[width=0.09\textwidth]{img_0001_resize.png}}
		&
		\adjustbox{valign=m}{\includegraphics[width=0.09\textwidth]{gt_0001.jpg}}
		&
		\adjustbox{valign=m}{\includegraphics[width=0.09\textwidth]{map_medsam_0001.jpg}}
		&
		\adjustbox{valign=m}{\includegraphics[width=0.09\textwidth]{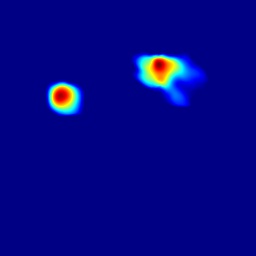}}
		&
		\adjustbox{valign=m}{\includegraphics[width=0.09\textwidth]{map_hc_0001.jpg}}
		&
		\adjustbox{valign=m}{\includegraphics[width=0.09\textwidth]{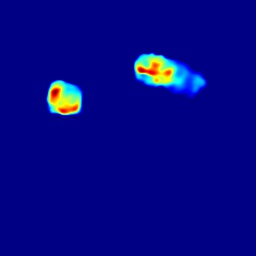}}
		&
		\adjustbox{valign=m}{\includegraphics[width=0.09\textwidth]{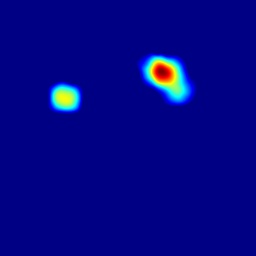}}
		&
		\adjustbox{valign=m}{\includegraphics[width=0.09\textwidth]{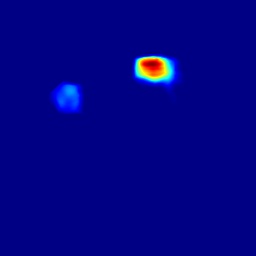}}
		&
		\adjustbox{valign=m}{\includegraphics[width=0.09\textwidth]{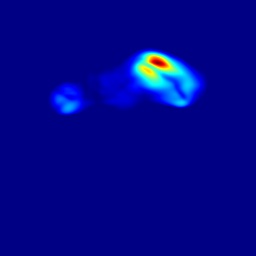}}
		&
		\adjustbox{valign=m}{\includegraphics[width=0.09\textwidth]{map_hc_proj_cnt_proto_0001.jpg}}	
		\\
		\vspace{1mm}
		\adjustbox{valign=m}{\includegraphics[width=0.09\textwidth]{img_0010_resize.png}}
		&
		\adjustbox{valign=m}{\includegraphics[width=0.09\textwidth]{gt_0010.jpg}}
		&
		\adjustbox{valign=m}{\includegraphics[width=0.09\textwidth]{map_medsam_0010.jpg}}
		&
		\adjustbox{valign=m}{\includegraphics[width=0.09\textwidth]{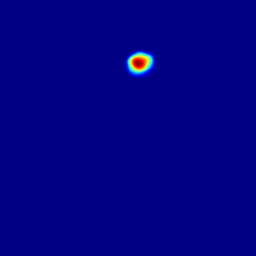}}
		&
		\adjustbox{valign=m}{\includegraphics[width=0.09\textwidth]{map_hc_0010.jpg}}
		&
		\adjustbox{valign=m}{\includegraphics[width=0.09\textwidth]{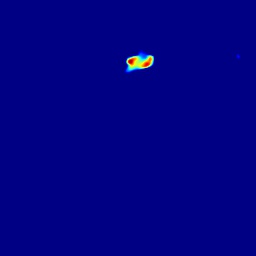}}
		&
		\adjustbox{valign=m}{\includegraphics[width=0.09\textwidth]{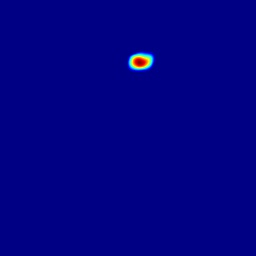}}
		&
		\adjustbox{valign=m}{\includegraphics[width=0.09\textwidth]{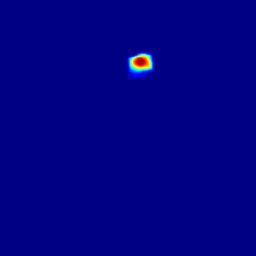}}
		&
		\adjustbox{valign=m}{\includegraphics[width=0.09\textwidth]{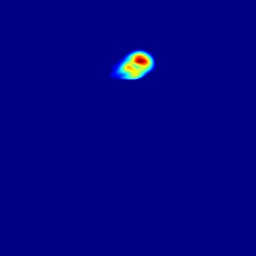}}
		&
		\adjustbox{valign=m}{\includegraphics[width=0.09\textwidth]{map_hc_proj_cnt_proto_0010.jpg}}	
		\\
		\vspace{1mm}
		\adjustbox{valign=m}{\includegraphics[width=0.09\textwidth]{images_101.PNG}}
		&
		\adjustbox{valign=m}{\includegraphics[width=0.09\textwidth]{gt_101.PNG}}
		&
		\adjustbox{valign=m}{\includegraphics[width=0.09\textwidth]{map_medsam_101.PNG}}
		&
		\adjustbox{valign=m}{\includegraphics[width=0.09\textwidth]{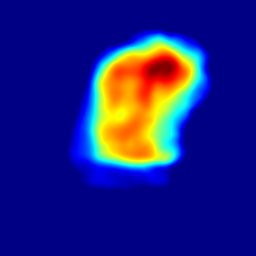}}
		&
		\adjustbox{valign=m}{\includegraphics[width=0.09\textwidth]{map_hc_101.PNG}}
		&
		\adjustbox{valign=m}{\includegraphics[width=0.09\textwidth]{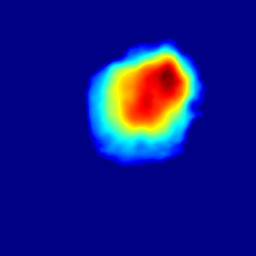}}
		&
		\adjustbox{valign=m}{\includegraphics[width=0.09\textwidth]{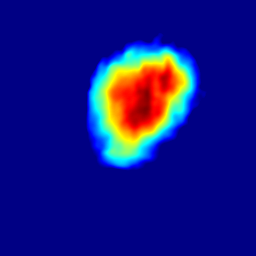}}
		&
		\adjustbox{valign=m}{\includegraphics[width=0.09\textwidth]{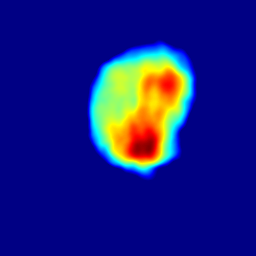}}
		&
		\adjustbox{valign=m}{\includegraphics[width=0.09\textwidth]{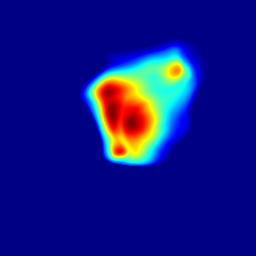}}
		&
		\adjustbox{valign=m}{\includegraphics[width=0.09\textwidth]{map_hc_proj_cnt_proto_101.PNG}}
		\\
		\vspace{1mm}
		\adjustbox{valign=m}{\includegraphics[width=0.09\textwidth]{images_128.PNG}}
		&
		\adjustbox{valign=m}{\includegraphics[width=0.09\textwidth]{gt_128.PNG}}
		&
		\adjustbox{valign=m}{\includegraphics[width=0.09\textwidth]{map_medsam_128.PNG}}
		&
		\adjustbox{valign=m}{\includegraphics[width=0.09\textwidth]{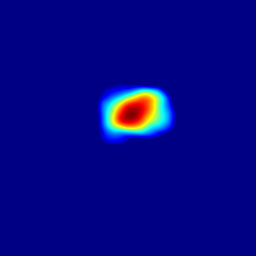}}
		&
		\adjustbox{valign=m}{\includegraphics[width=0.09\textwidth]{map_hc_128.PNG}}
		&
		\adjustbox{valign=m}{\includegraphics[width=0.09\textwidth]{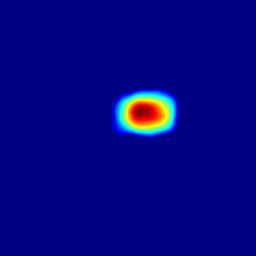}}
		&
		\adjustbox{valign=m}{\includegraphics[width=0.09\textwidth]{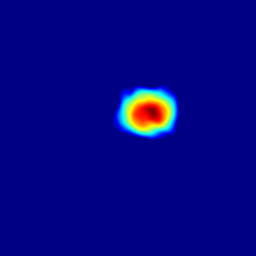}}
		&
		\adjustbox{valign=m}{\includegraphics[width=0.09\textwidth]{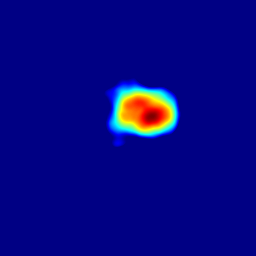}}
		&
		\adjustbox{valign=m}{\includegraphics[width=0.09\textwidth]{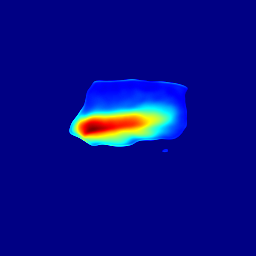}}
		&
		\adjustbox{valign=m}{\includegraphics[width=0.09\textwidth]{map_hc_proj_cnt_proto_128.PNG}}
		\\
		Images & GT & $P_{2}$ & $A_{2}$ & $P_{3}$ & $A_{3}$	& $B_{3}$ & $C_{3}$	& $D_{3}$ & $E_{3}$
		\\	
	\end{tabular}
	\caption{\textbf{Feature map visualizations of segmentation results in Fig.~\ref{fig:ablation_segmentation_result_hrloss}.} $P_{2}$, $A_{2}$, $P_{3}$, $A_{3}$, $B_{3}$, $C_{3}$, $D_{3}$ and $E_{3}$ denote the same models in Fig.~\ref{fig:ablation_segmentation_result_hrloss}. Example images are the same as those in Fig.~\ref{fig:ablation_segmentation_result_hclabel}.}
	\label{fig:ablation_heatmap_result_hrloss}
\end{figure*} 
As shown in Fig.~\ref{fig:ablation_segmentation_result_hrloss} and Fig.~\ref{fig:ablation_heatmap_result_hrloss}, Model $A_3$ reduced under-segmentation of nodules and over-segmentation of tissues compared to Model $P_3$, and obtained regions aligned to the ground truth (GT) masks. As illustrated in Table~\ref{tab:ablation_results_tn3k} and Table~\ref{tab:ablation_results_ddti}, Model $A_3$ obtained a significant improvement of over 6\% in mIoU and DSC across both datasets, and the decreased HD95 with the cost of only a small decrease in Precision. These results indicate that the alignment loss for location learning enhanced the ability to precisely capture target locations without misleading segmentation shapes.

When comparing Model $A_2$ with Model $P_2$ in Fig.~\ref{fig:ablation_segmentation_result_hrloss} and Fig.~\ref{fig:ablation_heatmap_result_hrloss}, it was disappointed to observe that the segmentation from Model $A_2$ is worse than that from Model $P_2$. The 2\% to 5\% decrese in mIoU shown in Table~\ref{tab:ablation_results_tn3k} and Table~\ref{tab:ablation_results_ddti} confirmed the observation. It could be attributed to that when using MedSAM to constrain position, the under-segmented labels introduce some erroneous information into the localization of nodule regions. At the same time, the results from prompted MedSAM introduce multiple shape information of nodules for training Model $P_2$ pixel-wisely, while the alignment loss focuses primarily on region localization and requires complementary shape losses for refinement.

In addition, by comparing Model $E_3$ with Model $D_3$, the improvement of all metrics was enormous. Fig.~\ref{fig:ablation_segmentation_result_hrloss} and Fig.~\ref{fig:ablation_heatmap_result_hrloss} further showed that without alignment loss, the segmentation of $D_3$ was inaccurate in terms of the shape and region alignment of the ground truth. These results indicate that 
%the alignment loss is the cornerstone of the algorithm's effectiveness. In the learning process, 
the alignment loss is necessary to obtain the correct localization of nodules, allowing the shape learning losses to play their maximum roles.

\begin{figure*}[!htbp]
\centering
\includegraphics[width=0.9\textwidth]{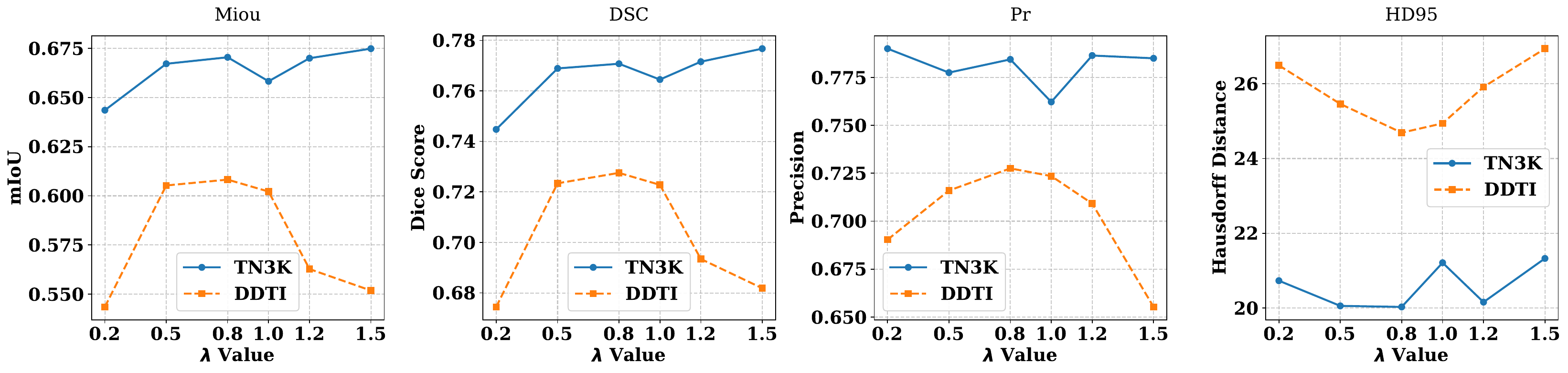}
\caption{Performance comparison of the model that combines alignment loss and contrastive loss based on HC-labels conducted on the TN3K and DDTI datasets with different weight parameters $\lambda$ of contrastive loss.}
\label{fig:weighted_comparison_lambda}
\end{figure*}

\subsubsection{Effectiveness of Contrastive Loss for Semantic Shape Learning}
The weight of the contrastive loss is controlled by the parameter $\lambda$. As shown in Fig.~\ref{fig:weighted_comparison_lambda}, experimental results supervised by proposed high-confidence labels illustrated that the Model $B_3$ achieved its best comprehensive performance when $\lambda$ was set to 0.8.

As shown in Fig.~\ref{fig:ablation_segmentation_result_hrloss} and Fig.~\ref{fig:ablation_heatmap_result_hrloss},  Model $B_3$ outperformed Model $A_3$ by providing more accurate overall shapes and feature distributions of nodules. Quantitatively, as illustrated in Table~\ref{tab:ablation_results_tn3k} and Table~\ref{tab:ablation_results_ddti}, Model $B_3$ obtained improvements in DSC up to 6.37\%, mIoU up to 7.37\%, Precision up to 3.63\%  over Model $A_3$, with HD95 reduced by at least 1.67 pixels on TN3K and 3.80 pixels on DDTI. Furthermore, Model $E_3$ showed improvements in mIoU by over 0.57\% on the TN3K dataset and nearly 0.50\% on the DDTI dataset compared to the corresponding Model $C_3$, respectively.
% These results illustrated that contrastive loss can improve segmentation accuracy and refinement.
% The feature heatmap visualization in Fig.~\ref{fig:ablation_result} shows several example images processed using different models. 
% These results illustrated that by incorporating contrastive loss, the model improved segmentation precision while addressing over-segmentation and under-segmentation, as shown by comparing the results of Model D with Model C. 
These results demonstrate that the incorporation of contrastive loss ensures the predicted segmentation regions closely align with the ground truth. By refining the predicted foreground and background feature clusters extracted from the whole image domain, the contrastive loss effectively improves segmentation precision.

\subsubsection{Effectiveness of Prototype Correlation Loss for Semantic Shape Learning}
\begin{figure*}[!htbp]
\centering
\includegraphics[width=0.9\textwidth]{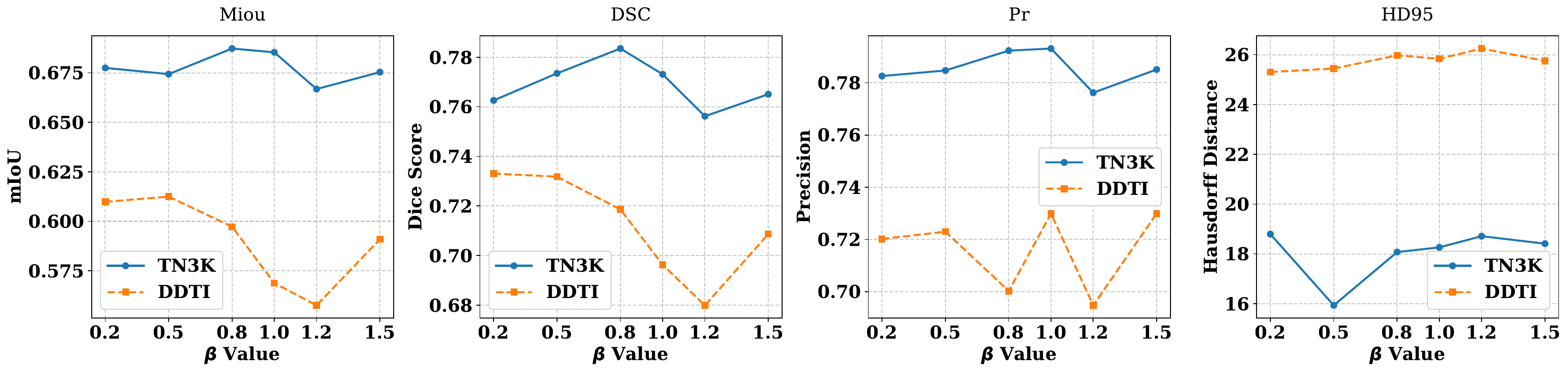}
\caption{Performance comparison of the model that combines alignment loss and prototype correlation loss based on HC-labels conducted on the TN3K and DDTI datasets with different weight parameters $\beta$ of prototype correlation loss.}
% \vspace{-0.2in}
\label{fig:weighted_comparison_beta}
\end{figure*}
The weight of the prototype correlation loss is controlled by the parameter $\beta$. Experimental results supervised by high-confidence labels illustrated that Model $C_3$ achieved its best comprehensive performance when $\beta$ was set to 0.8 on the TN3K dataset and 0.5 on the DDTI dataset, as shown in Fig.~\ref{fig:weighted_comparison_beta}. 
% To validate the benefits of adding prototype correlation loss as a shape constraint, Model C was compared with Model A. Under the condition of using high-confidence labels, the results reveal that Model C achieved a nearly 4\% improvement in mIoU on the TN3K dataset and 6\% improvement on the DDTI dataset compared to Model A, accompanied by a significant reduction in Hausdorff Distance at 95th percentile, with a decrease of over 2.28 pixels on the TN3K dataset and 4 pixels on the DDTI dataset. When comparing Model D to Model B, the integration of prototype correlation loss provided superior segmentation regions and edges. Although the impact of the combined losses was non-linear, this approach achieved consistent improvements in mIoU. These enhancements were additionally corroborated by results derived from topologically based labels and the prompted MedSAM results labels.

As shown in Fig.~\ref{fig:ablation_segmentation_result_hrloss} and Fig.~\ref{fig:ablation_heatmap_result_hrloss}, Model $C_3$ and $E_3$ provided clearer and more delicate feature distributions and boundary delineations than $A_3$ and $B_3$, respectively. Moreover, the uncertain ranges of Model $C_3$ and $E_3$ on the nodule boundaries were shrunk remarkably. The qualitative observations were confirmed by the quantitative statistics in Table~\ref{tab:ablation_results_tn3k} and \ref{tab:ablation_results_ddti}. When comparing Model $C_3$ with Model $A_3$, it achieved nearly 4\% and 6\% improvement in mIoU on the TN3K and DDTI dataset, along with a significant reduction in HD95 by over 2.28 and 4 pixels on TN3K and DDTI dataset. When comparing Model $E_3$ with Model $B_3$, the inclusion of prototype correlation loss also led to superior segmentation results. 

The improvement in qualitative and quantitative results proves that adding prototype correlation loss as a shape constraint effectively helps the network to make the uncertain pixels on the classification boundary in the feature space closer to their category prototypes, resulting in superior segmentation regions and edges.

\subsection{Comparisons with State-of-the-art Methods}
We evaluated the segmentation performance of our proposed framework and compared it with several state-of-the-art weakly supervised methods: 1) SCRF~\cite{zhang2020scrf}, which combined coarse segmentation information with conditional random fields (CRF) to produce segmenation results, 2) UNCRF~\cite{mahani2022uncrf}, which refined these pseudo-labels by incorporating uncertainty estimation to enhance segmentation accuracy, 3) BoxInst~\cite{tian2021BoxInst}, which introduced a framework use projection loss and pairwise color affinity loss to obtain segmentation results, 4) CoarseFine~\cite{chi2025coarse}, which proposed a dual-branch framework to calibrate semantic features into nodule segmentation. 5) WSDAC~\cite{li2023wsdac}, which proposed a novel weakly supervised deep active contour model for nodule segmentation, 6) S2ME~\cite{wang2023s2me}, which introduced an entropy-guided weakly supervised polyp segmentation framework, 7) IDMPS~\cite{zhao2024IDMPS}, which used an asymmetric learning framework to learn nodule segmentation from conservative labels and radical labels. 

% The quantitative results of TN3K and DDTI are shown in Table~\ref {tab:comparison_result_tn3k} and Table~\ref {tab:comparison_result_ddti}. The qualitative results of TN3K and DDTI are shown in Fig.~\ref{fig:comparison_tn3k} and Fig.~\ref {fig:comparison_ddti}, respectively.

\begin{table*}[!htbp]
\centering
\caption{Quantitative comparison results (Mean $\pm$ STD) of different methods on the TN3K~\cite{gong2021tn3k} dataset.}
%\label{tab:ablation_tn3k}
\small
\setlength{\tabcolsep}{4pt}
\renewcommand{\arraystretch}{1.3}
\begin{tabular}{lccccc}
\toprule[1.2pt]
\multirow{2}{*}{Supervision} & \multirow{2}{*}{Model} & \multicolumn{4}{c}{Metrics} \\
\cmidrule(r){3-6}
& & mIoU (\%) $\uparrow$ & DSC (\%) $\uparrow$ & Precision (\%) $\uparrow$  & HD95 (px) $\downarrow$\\
\midrule[0.8pt]
\multirow{8}{*}{Weakly} 
& SCRF~\cite{zhang2020scrf} & $56.68\pm21.85$ & $69.19\pm23.00$ & $63.47\pm22.81$ & $26.22\pm29.68$ \\
& UNCRF~\cite{mahani2022uncrf}& $56.27\pm23.89$ & $68.21\pm25.25$ & $66.29\pm24.58$ & $25.11\pm29.15$ \\
& BoxInst~\cite{tian2021BoxInst} & $64.49\pm23.53$ & $75.27\pm22.36$ & $78.54\pm23.91$ & $20.70\pm27.17$ \\
& CoarseFine~\cite{chi2025coarse} & $65.46\pm24.68$ & $76.14\pm27.06$ & $75.47\pm24.36$ & $20.07\pm28.13$ \\
& WSDAC~\cite{li2023wsdac} & $61.46\pm17.56$ & $74.25\pm17.59$ & $76.81\pm24.02$ & $19.09\pm24.69$ \\
& S2ME~\cite{wang2023s2me} & $62.76\pm23.37$ & $73.46\pm23.41$ & $75.42\pm25.73$ & $23.77\pm28.82$\\
& IDMPS~\cite{zhao2024IDMPS} & $66.52\pm26.64$ & $75.69\pm26.73$ & $80.40\pm25.72$  & $21.48\pm28.83$\\
& HCHR(ours) & $\textbf{69.30}\pm\textbf{22.38}$ & $\textbf{79.10}\pm\textbf{21.37}$ & $\textbf{82.49}\pm\textbf{22.69}$ & $\textbf{17.20}\pm\textbf{26.97}$\\
\cmidrule(r){1-6}
Fully & U-Net~\cite{ronneberger2015unet} & $68.49\pm24.34$ & $78.07\pm22.76$ & $79.08\pm24.36$ & $21.13\pm30.69$ \\
\bottomrule[1.2pt]
\end{tabular}
\vspace{2mm}
\label{tab:comparison_result_tn3k}
\end{table*}

%\begin{figure*}[!htbp]
%\centering
%\includegraphics[width=0.98\textwidth]{compare_tn3k.png}
%\caption{\textbf{Qualitative comparison results on TN3K dataset.} Red indicates correct thyroid nodule predictions, green represents the under-segmentations of thyroid nodules, and blue shows an over-segmentations of other organs as thyroid nodules.}
%\label{fig:comparison_tn3k}
%\end{figure*}
\begin{figure*}
\footnotesize
\centering
\tabcolsep=0.5mm
\begin{tabular}{ccccccccccc}
	\vspace{1mm}
	\adjustbox{valign=m}{\includegraphics[width=0.084\textwidth]{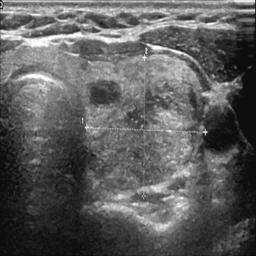}}
	&
	\adjustbox{valign=m}{\includegraphics[width=0.084\textwidth]{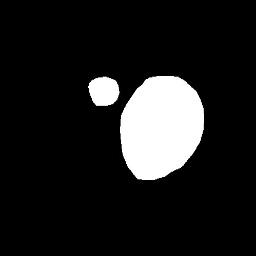}}
	&
	\adjustbox{valign=m}{\includegraphics[width=0.084\textwidth]{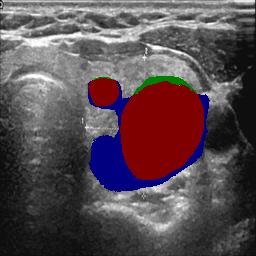}}
	&
	\adjustbox{valign=m}{\includegraphics[width=0.084\textwidth]{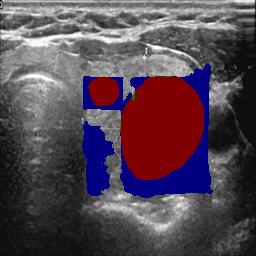}}
	&
	\adjustbox{valign=m}{\includegraphics[width=0.084\textwidth]{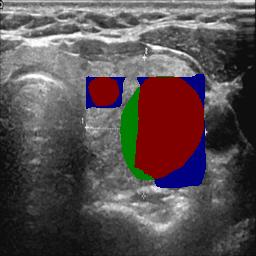}}
	&
	\adjustbox{valign=m}{\includegraphics[width=0.084\textwidth]{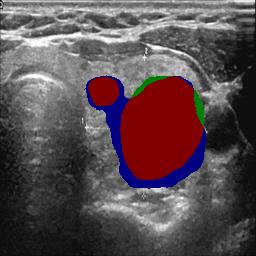}}
	&
	\adjustbox{valign=m}{\includegraphics[width=0.084\textwidth]{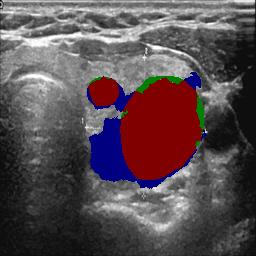}}
	&
	\adjustbox{valign=m}{\includegraphics[width=0.084\textwidth]{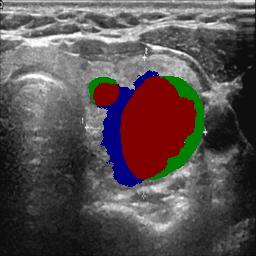}}
	&
	\adjustbox{valign=m}{\includegraphics[width=0.084\textwidth]{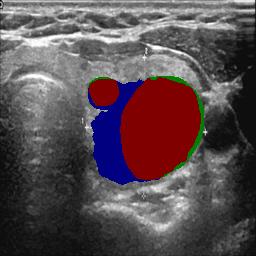}}
	&
	\adjustbox{valign=m}{\includegraphics[width=0.084\textwidth]{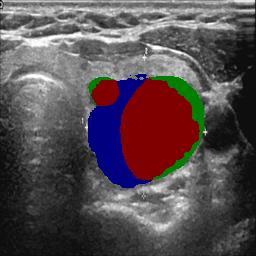}}
	&
	\adjustbox{valign=m}{\includegraphics[width=0.084\textwidth]{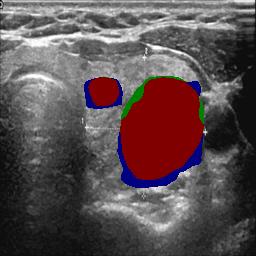}}
	\\
	\vspace{1mm}
	\adjustbox{valign=m}{\includegraphics[width=0.084\textwidth]{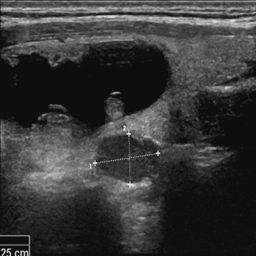}}
	&
	\adjustbox{valign=m}{\includegraphics[width=0.084\textwidth]{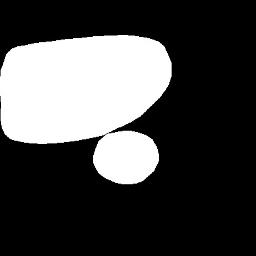}}
	&
	\adjustbox{valign=m}{\includegraphics[width=0.084\textwidth]{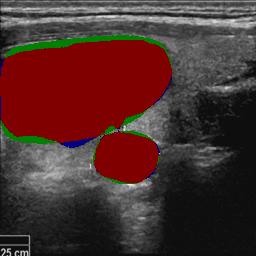}}
	&
	\adjustbox{valign=m}{\includegraphics[width=0.084\textwidth]{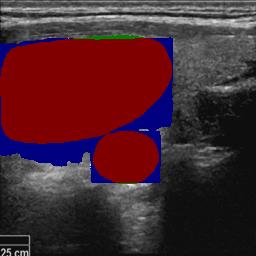}}
	&
	\adjustbox{valign=m}{\includegraphics[width=0.084\textwidth]{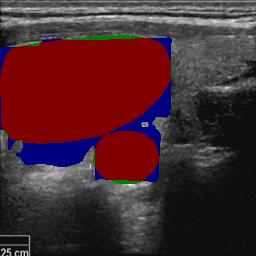}}
	&
	\adjustbox{valign=m}{\includegraphics[width=0.084\textwidth]{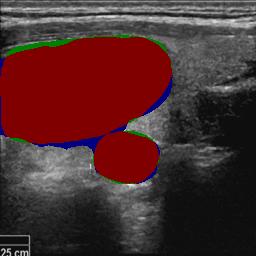}}
	&
	\adjustbox{valign=m}{\includegraphics[width=0.084\textwidth]{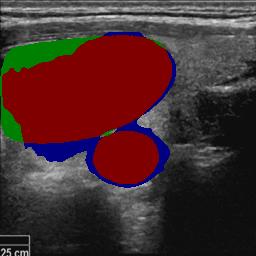}}
	&
	\adjustbox{valign=m}{\includegraphics[width=0.084\textwidth]{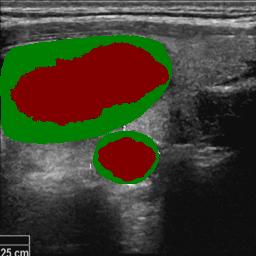}}
	&
	\adjustbox{valign=m}{\includegraphics[width=0.084\textwidth]{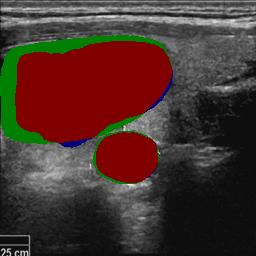}}
	&
	\adjustbox{valign=m}{\includegraphics[width=0.084\textwidth]{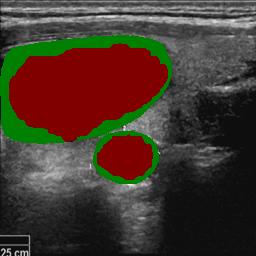}}
	&
	\adjustbox{valign=m}{\includegraphics[width=0.084\textwidth]{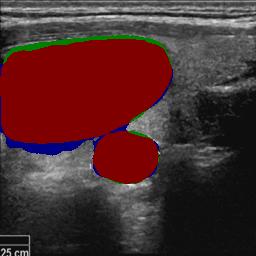}}
	\\
	\vspace{1mm}
	\adjustbox{valign=m}{\includegraphics[width=0.084\textwidth]{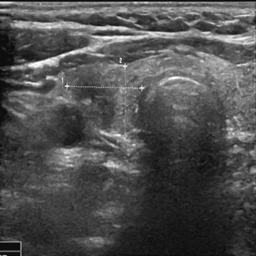}}
	&
	\adjustbox{valign=m}{\includegraphics[width=0.084\textwidth]{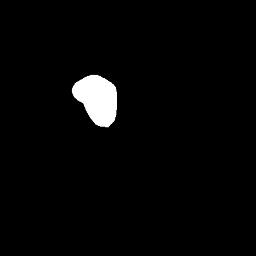}}
	&
	\adjustbox{valign=m}{\includegraphics[width=0.084\textwidth]{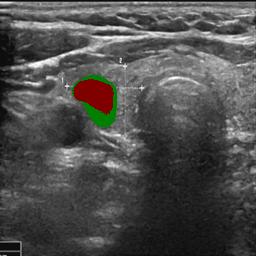}}
	&
	\adjustbox{valign=m}{\includegraphics[width=0.084\textwidth]{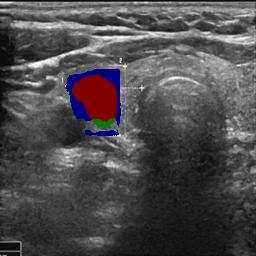}}
	&
	\adjustbox{valign=m}{\includegraphics[width=0.084\textwidth]{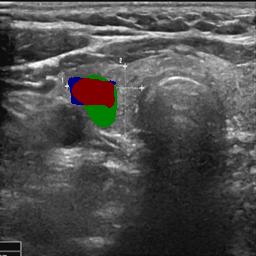}}
	&
	\adjustbox{valign=m}{\includegraphics[width=0.084\textwidth]{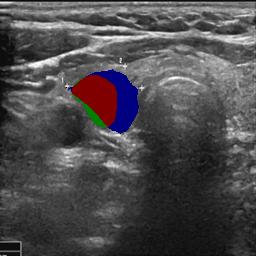}}
	&
	\adjustbox{valign=m}{\includegraphics[width=0.084\textwidth]{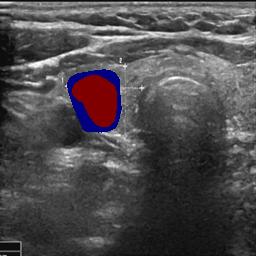}}
	&
	\adjustbox{valign=m}{\includegraphics[width=0.084\textwidth]{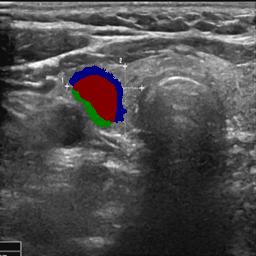}}
	&
	\adjustbox{valign=m}{\includegraphics[width=0.084\textwidth]{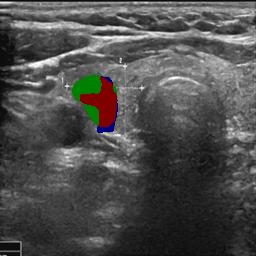}}
	&
	\adjustbox{valign=m}{\includegraphics[width=0.084\textwidth]{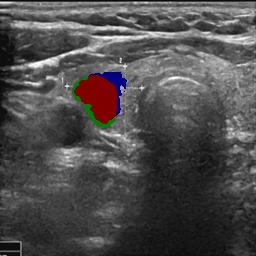}}
	&
	\adjustbox{valign=m}{\includegraphics[width=0.084\textwidth]{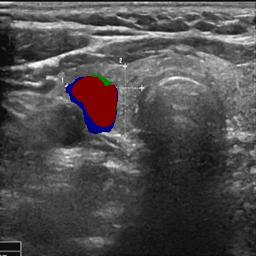}}
	\\
	Images & GT & U-Net & SCRF & UNCRF & Boxinst & CoarseFine & WSDAC & S2ME & IDMPS & proposed
	\\
\end{tabular}
\caption{\textbf{Qualitative comparison results on TN3K dataset.} Red indicates correct thyroid nodule predictions, green represents the under-segmentations of thyroid nodules, and blue shows an over-segmentations of other tissues as thyroid nodules.}
\label{fig:comparison_tn3k}
\end{figure*}
% the fully supervised networks U-Net, Dense-UNet, and CENet yield results that exhibit minor over-segmentation and under-segmentation.
\subsubsection{Comparative Results on TN3K dataset}
As shown in Fig.~\ref{fig:comparison_tn3k}, SCRF and UNCRF tended to produce box-like segmentation with significant over-segmentation. While BoxInst and IDMPS outperformed other existing methods in high-contrast scenarios (Row 3), they still struggled with over-segmentation of incomplete nodules located at image boundaries (Row 1). CoarseFine showed better overall performance than most WSS methods, but still struggled with over-segmentation (Row 1 and 3) and under-segmentation (Row 2) in all cases. WSDAC and S2ME frequently under-segmented thyroid nodules in images with multiple nodules or irregular nodule shapes, and WSDAC suffered from severe under-segmentation for images with complex backgrounds (Row 2). 
In contrast, our proposed method achieved more consistent segmentation regions and delicate segmentation edges, while exhibited even less over- and under-segmentation than fully supervised networks using the same backbone. Quantitative results in Table~\ref{tab:comparison_result_tn3k} further showed that our framework using the U-Net backbone outperformed state-of-the-art weakly supervised methods. Specifically, it achieved an average mIOU of 69.30\%, DSC of 79.10\%, Precision of 82.49\%, and HD95 of 17.20 pixels. These results were even better than the fully supervised U-Net, which achieved an average mIOU of 68.49\%, DSC of 78.07\%, Precision of 79.08\%, and HD95 of 21.13 pixels.

\subsubsection{Comparative Results on DDTI dataset}
\begin{table*}[!htbp]
\centering
\caption{Quantitative comparison results (Mean $\pm$ STD) of different methods on the DDTI~\cite{pedraza2015DDTI} dataset.}
\label{tab:ablation_tn3k}
\small
\setlength{\tabcolsep}{4pt}
\renewcommand{\arraystretch}{1.3}
\begin{tabular}{lccccc}
\toprule[1.2pt]
\multirow{2}{*}{Supervision} & \multirow{2}{*}{Model} & \multicolumn{4}{c}{Metrics} \\
\cmidrule(r){3-6}
& & mIoU (\%) $\uparrow$ & DSC (\%) $\uparrow$ & Precision (\%) $\uparrow$  & HD95 (px) $\downarrow$\\
\midrule[0.8pt]
\multirow{8}{*}{Weakly} 
& SCRF~\cite{zhang2020scrf} & $50.29\pm20.38$ & $64.22\pm20.19$ & $57.27\pm24.04$ & $36.88\pm26.26$ \\
& UNCRF~\cite{mahani2022uncrf} & $49.55\pm20.67$ & $63.42\pm20.84$ & $61.52\pm25.60$ & $33.39\pm24.46$ \\
& BoxInst~\cite{tian2021BoxInst} & $52.53\pm20.04$ & $66.27\pm20.15$ & $64.52\pm29.32$ & $33.77\pm20.04$ \\
& CoarseFine~\cite{chi2025coarse} & $60.44\pm21.38$ & $72.28\pm19.56$ & $72.79\pm24.69$ & $26.39\pm24.89$ \\
& WSDAC~\cite{li2023wsdac} & $58.60\pm18.82$ & $71.78\pm18.01$ & $\textbf{76.95}\pm\textbf{26.13}$ & $26.56\pm23.96$ \\
& S2ME~\cite{wang2023s2me} & $61.35\pm24.39$ & $72.65\pm22.81$ & $73.98\pm26.22$ & $26.09\pm28.64$ \\
& IDMPS~\cite{zhao2024IDMPS} & $61.76\pm23.11$ & $74.26\pm22.02$ & $69.44\pm26.89$ & $25.21\pm27.66$\\
& HCHR(ours) & $\textbf{62.52}\pm\textbf{20.72}$ & $\textbf{74.55}\pm\textbf{18.91}$ & $74.66\pm24.62$ & $\textbf{23.82}\pm\textbf{24.77}$\\
\cmidrule(r){1-6}
Fully & U-Net~\cite{ronneberger2015unet} & $58.97\pm23.15$ & $71.05\pm21.88$ & $70.83\pm27.23$ & $27.63\pm26.81$ \\
\bottomrule[1.2pt]
\end{tabular}
\vspace{2mm}
\label{tab:comparison_result_ddti}
\end{table*}

%\begin{figure}[!htbp]
%\centering
%\includegraphics[width=0.98\textwidth]{compare_ddti.png}
%\caption{\textbf{Qualitative comparison results on DDTI dataset.} Red indicates correct thyroid nodule predictions, green represents the under-segmentations of thyroid nodules, and blue shows an over-segmentations of other organs as thyroid nodules.}
%% \vspace{-0.2in}
%\label{fig:comparison_ddti}
%\end{figure}
\begin{figure*}
	\footnotesize
	\centering
	\tabcolsep=0.5mm
	\begin{tabular}{ccccccccccc}
		\vspace{1mm}
		\adjustbox{valign=m}{\includegraphics[width=0.084\textwidth]{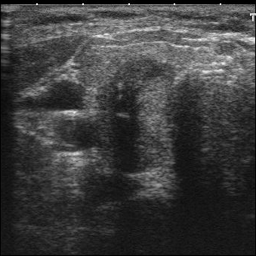}}
		&
		\adjustbox{valign=m}{\includegraphics[width=0.084\textwidth]{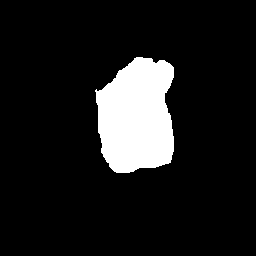}}
		&
		\adjustbox{valign=m}{\includegraphics[width=0.084\textwidth]{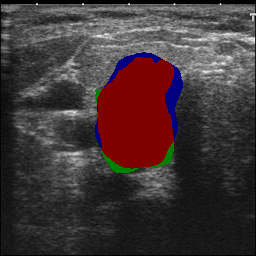}}
		&
		\adjustbox{valign=m}{\includegraphics[width=0.084\textwidth]{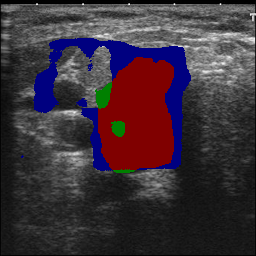}}
		&
		\adjustbox{valign=m}{\includegraphics[width=0.084\textwidth]{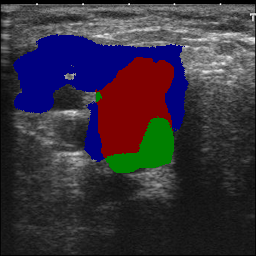}}
		&
		\adjustbox{valign=m}{\includegraphics[width=0.084\textwidth]{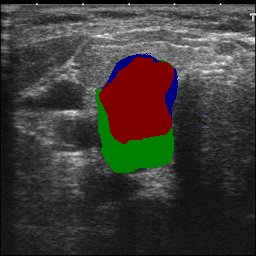}}
		&
		\adjustbox{valign=m}{\includegraphics[width=0.084\textwidth]{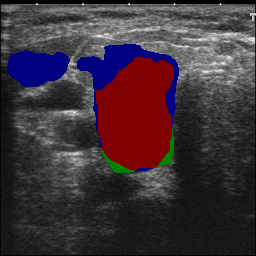}}
		&
		\adjustbox{valign=m}{\includegraphics[width=0.084\textwidth]{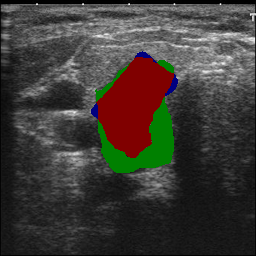}}
		&
		\adjustbox{valign=m}{\includegraphics[width=0.084\textwidth]{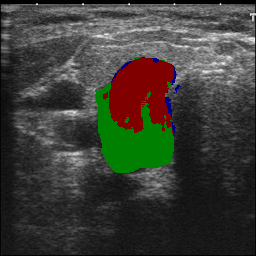}}
		&
		\adjustbox{valign=m}{\includegraphics[width=0.084\textwidth]{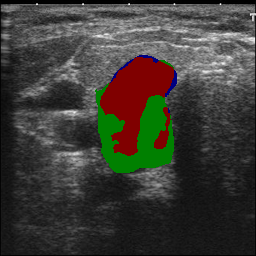}}
		&
		\adjustbox{valign=m}{\includegraphics[width=0.084\textwidth]{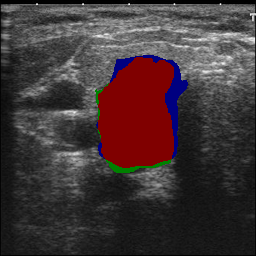}}
		\\
		\vspace{1mm}
		\adjustbox{valign=m}{\includegraphics[width=0.084\textwidth]{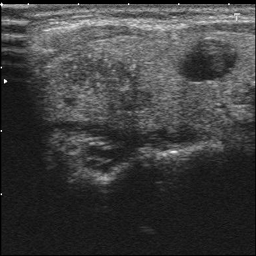}}
		&
		\adjustbox{valign=m}{\includegraphics[width=0.084\textwidth]{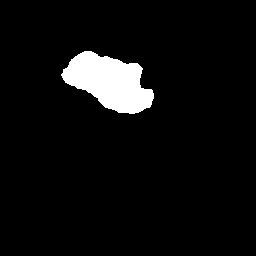}}
		&
		\adjustbox{valign=m}{\includegraphics[width=0.084\textwidth]{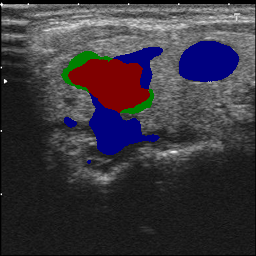}}
		&
		\adjustbox{valign=m}{\includegraphics[width=0.084\textwidth]{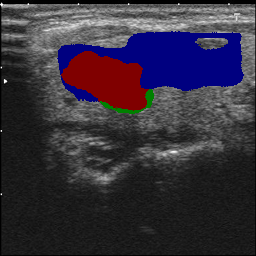}}
		&
		\adjustbox{valign=m}{\includegraphics[width=0.084\textwidth]{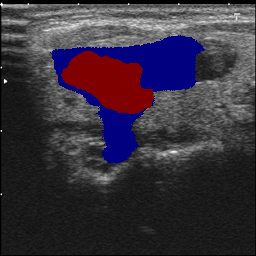}}
		&
		\adjustbox{valign=m}{\includegraphics[width=0.084\textwidth]{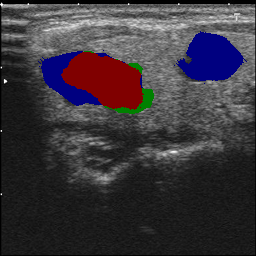}}
		&
		\adjustbox{valign=m}{\includegraphics[width=0.084\textwidth]{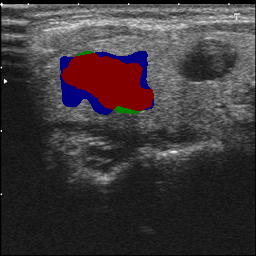}}
		&
		\adjustbox{valign=m}{\includegraphics[width=0.084\textwidth]{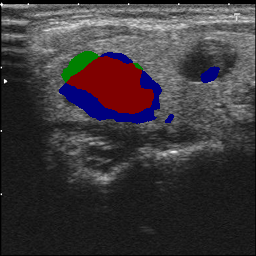}}
		&
		\adjustbox{valign=m}{\includegraphics[width=0.084\textwidth]{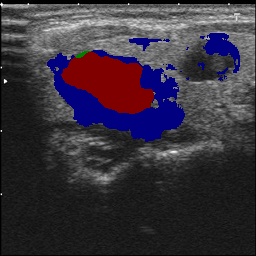}}
		&
		\adjustbox{valign=m}{\includegraphics[width=0.084\textwidth]{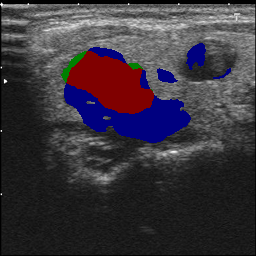}}
		&
		\adjustbox{valign=m}{\includegraphics[width=0.084\textwidth]{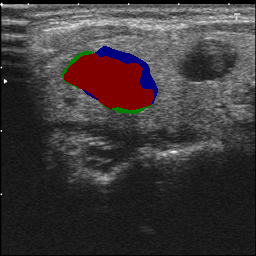}}
		\\
		\vspace{1mm}
		\adjustbox{valign=m}{\includegraphics[width=0.084\textwidth]{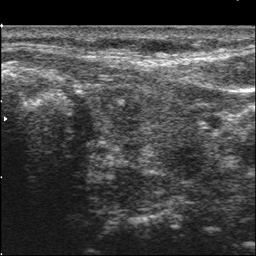}}
		&
		\adjustbox{valign=m}{\includegraphics[width=0.084\textwidth]{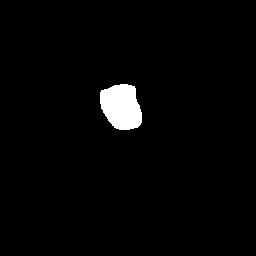}}
		&
		\adjustbox{valign=m}{\includegraphics[width=0.084\textwidth]{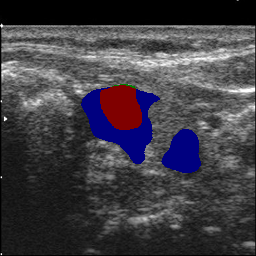}}
		&
		\adjustbox{valign=m}{\includegraphics[width=0.084\textwidth]{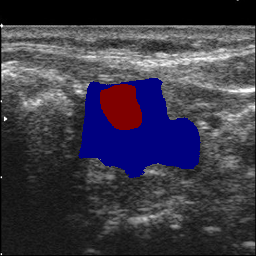}}
		&
		\adjustbox{valign=m}{\includegraphics[width=0.084\textwidth]{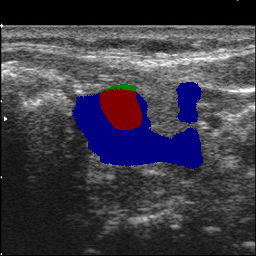}}
		&
		\adjustbox{valign=m}{\includegraphics[width=0.084\textwidth]{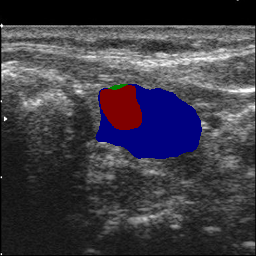}}
		&
		\adjustbox{valign=m}{\includegraphics[width=0.084\textwidth]{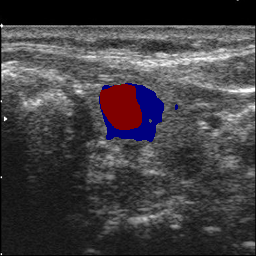}}
		&
		\adjustbox{valign=m}{\includegraphics[width=0.084\textwidth]{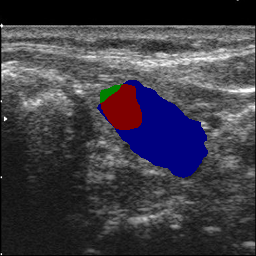}}
		&
		\adjustbox{valign=m}{\includegraphics[width=0.084\textwidth]{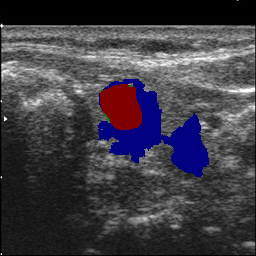}}
		&
		\adjustbox{valign=m}{\includegraphics[width=0.084\textwidth]{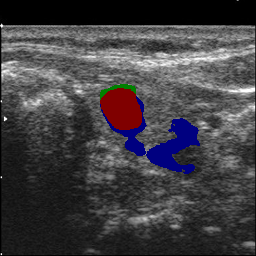}}
		&
		\adjustbox{valign=m}{\includegraphics[width=0.084\textwidth]{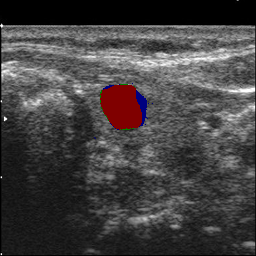}}
		\\
		Images & GT & U-Net & SCRF & UNCRF & Boxinst & CoarseFine & WSDAC & S2ME & IDMPS & proposed
		\\
	\end{tabular}
	\caption{\textbf{Qualitative comparison results on DDTI dataset.} Red indicates correct thyroid nodule predictions, green represents the under-segmentations of thyroid nodules, and blue shows an over-segmentations of other tissues as thyroid nodules.}
	\label{fig:comparison_ddti}
\end{figure*}

As illustrated in Fig.~\ref{fig:comparison_ddti}, fully supervised networks showed significant over-segmentation in images where the foreground and background tissues were similar (Row 2 and 3). Weakly supervised algorithms, such as SCRF and UNCRF, exhibited severe over-segmentation of small targets (Row 3) and targets with complex shapes (Row 1 and 2). 
% especially in images with similar foreground and background tissues 
BoxInst and CoarseFine were suitable for simple background issues (Row 2) but struggled with over- and under-segmentation when processing low-contrast images (Row 1 and 3). 
WSDAC performed the highest precision with minimal over-segmentation, but tended to exhibit more under-segmentation in all cases compared to other weakly supervised methods. 
S2ME and IDMPS covered most of the segmented areas (Row 2 and 3), but the segmentation results are discontinuous, and they faced more severe under-segmentation issues in low-contrast images (Row 1). 
% IDMPS provided superior segmentation results with reduced over-segmentation across images with complex shapes (Row 1), but they still exhibited gaps in low-contrast images with over-segmentation and under-segmentation (Row 3).
In contrast, our algorithm not only effectively reduced over-segmentation but also improved shape adaptation, surpassing fully supervised methods. 
% Furthermore, by integrating a more efficient feature extraction backbone, such as CENet, we achieved improved segmentation precision, with accurate shape delineation and fine edge fitting. 

% The quantitative comparison in Table~\ref{tab:comparison_result_ddti} further supports these observations. Our proposed method outperformed all weakly supervised methods and achieved a 3.55\% improvement in mIoU and a reduction in HD95 of 3.81 pixels compared to full-supervised methods, demonstrating its superior segmentation capabilities in terms of region alignment and shape fitting.
The quantitative comparison in Table~\ref{tab:comparison_result_ddti} further confirms he qualitative observations. Our proposed method outperformed all weakly supervised approaches, achieving a 3.55\% improvement in mIoU and a 3.81 pixels reduction in HD95 compared to fully supervised methods. This proved the superior segmentation performance of our framework in terms of region alignment and shape fitting.

\section{Discussion}
\subsection{Comparison with the State-of-the-art Weakly Supervised Segmentation Methods}
\begin{figure*}
	\footnotesize
	\centering
	\tabcolsep=0.5mm
	\begin{tabular}{ccccccccccc}
		\vspace{1mm}
		\adjustbox{valign=m}{\includegraphics[width=0.084\textwidth]{0041_resize.png}}
		&
		\adjustbox{valign=m}{\includegraphics[width=0.084\textwidth]{0041_gt.jpg}}
		&
		\adjustbox{valign=m}{\includegraphics[width=0.084\textwidth]{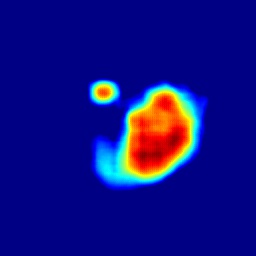}}
		&
		\adjustbox{valign=m}{\includegraphics[width=0.084\textwidth]{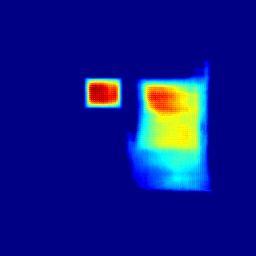}}
		&
		\adjustbox{valign=m}{\includegraphics[width=0.084\textwidth]{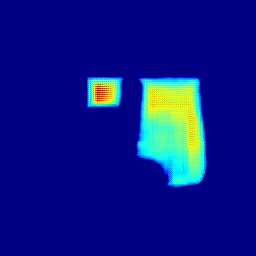}}
		&
		\adjustbox{valign=m}{\includegraphics[width=0.084\textwidth]{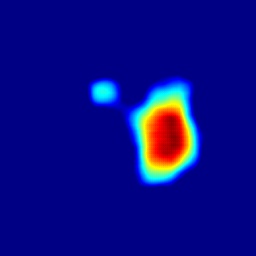}}
		&
		\adjustbox{valign=m}{\includegraphics[width=0.084\textwidth]{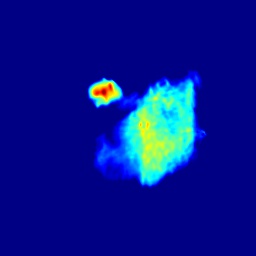}}
		&
		\adjustbox{valign=m}{\includegraphics[width=0.084\textwidth]{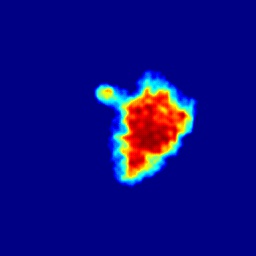}}
		&
		\adjustbox{valign=m}{\includegraphics[width=0.084\textwidth]{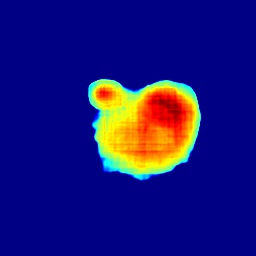}}
		&
		\adjustbox{valign=m}{\includegraphics[width=0.084\textwidth]{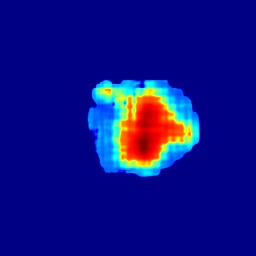}}
		&
		\adjustbox{valign=m}{\includegraphics[width=0.084\textwidth]{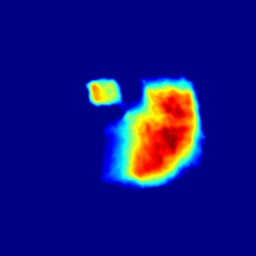}}
		\\
		\vspace{1mm}
		\adjustbox{valign=m}{\includegraphics[width=0.084\textwidth]{0080_resize.png}}
		&
		\adjustbox{valign=m}{\includegraphics[width=0.084\textwidth]{0080_gt.jpg}}
		&
		\adjustbox{valign=m}{\includegraphics[width=0.084\textwidth]{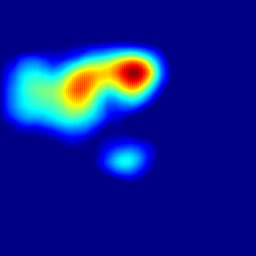}}
		&
		\adjustbox{valign=m}{\includegraphics[width=0.084\textwidth]{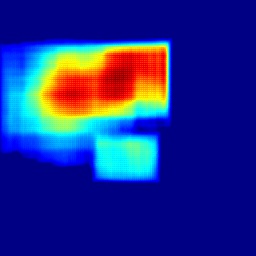}}
		&
		\adjustbox{valign=m}{\includegraphics[width=0.084\textwidth]{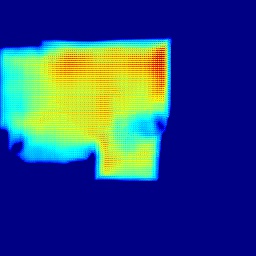}}
		&
		\adjustbox{valign=m}{\includegraphics[width=0.084\textwidth]{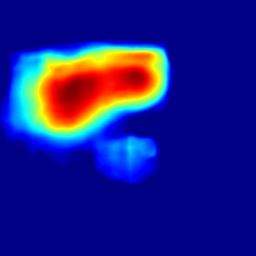}}
		&
		\adjustbox{valign=m}{\includegraphics[width=0.084\textwidth]{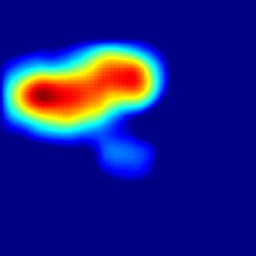}}
		&
		\adjustbox{valign=m}{\includegraphics[width=0.084\textwidth]{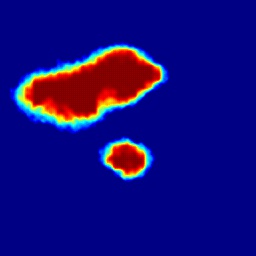}}
		&
		\adjustbox{valign=m}{\includegraphics[width=0.084\textwidth]{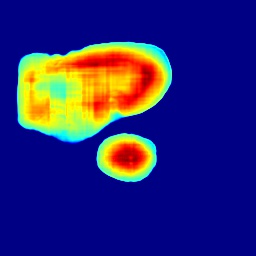}}
		&
		\adjustbox{valign=m}{\includegraphics[width=0.084\textwidth]{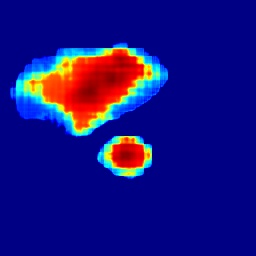}}
		&
		\adjustbox{valign=m}{\includegraphics[width=0.084\textwidth]{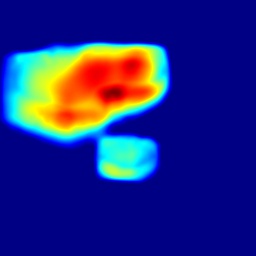}}
		\\
		\vspace{1mm}
		\adjustbox{valign=m}{\includegraphics[width=0.084\textwidth]{0081_resize.png}}
		&
		\adjustbox{valign=m}{\includegraphics[width=0.084\textwidth]{0081_gt.jpg}}
		&
		\adjustbox{valign=m}{\includegraphics[width=0.084\textwidth]{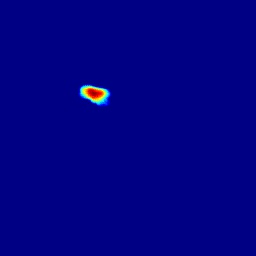}}
		&
		\adjustbox{valign=m}{\includegraphics[width=0.084\textwidth]{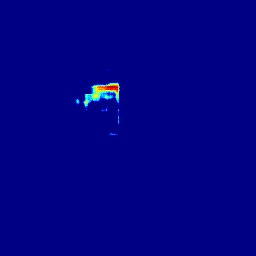}}
		&
		\adjustbox{valign=m}{\includegraphics[width=0.084\textwidth]{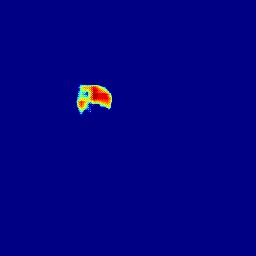}}
		&
		\adjustbox{valign=m}{\includegraphics[width=0.084\textwidth]{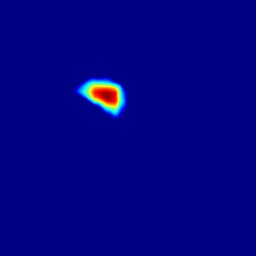}}
		&
		\adjustbox{valign=m}{\includegraphics[width=0.084\textwidth]{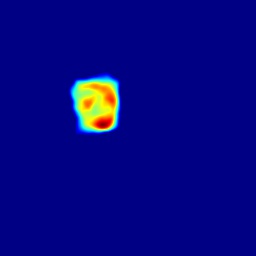}}
		&
		\adjustbox{valign=m}{\includegraphics[width=0.084\textwidth]{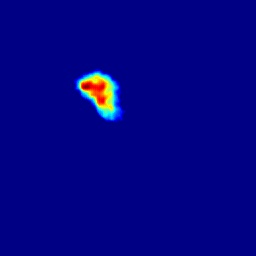}}
		&
		\adjustbox{valign=m}{\includegraphics[width=0.084\textwidth]{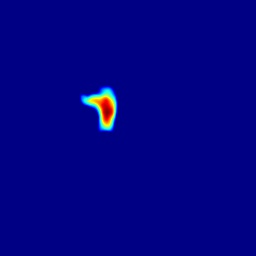}}
		&
		\adjustbox{valign=m}{\includegraphics[width=0.084\textwidth]{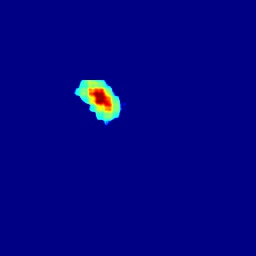}}
		&
		\adjustbox{valign=m}{\includegraphics[width=0.084\textwidth]{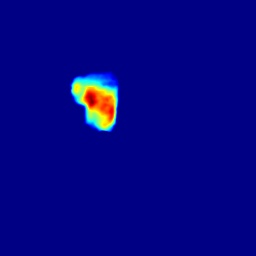}}
		\\
		\vspace{1mm}
		\adjustbox{valign=m}{\includegraphics[width=0.084\textwidth]{86.PNG}}
		&
		\adjustbox{valign=m}{\includegraphics[width=0.084\textwidth]{86_gt.PNG}}
		&
		\adjustbox{valign=m}{\includegraphics[width=0.084\textwidth]{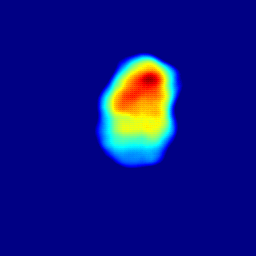}}
		&
		\adjustbox{valign=m}{\includegraphics[width=0.084\textwidth]{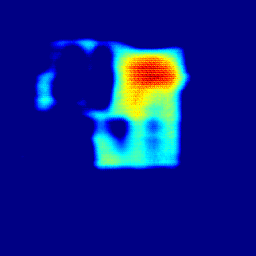}}
		&
		\adjustbox{valign=m}{\includegraphics[width=0.084\textwidth]{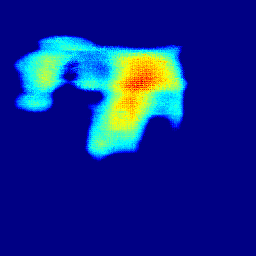}}
		&
		\adjustbox{valign=m}{\includegraphics[width=0.084\textwidth]{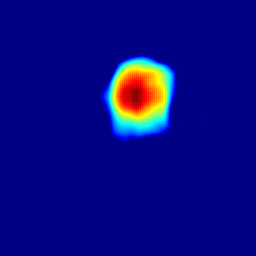}}
		&
		\adjustbox{valign=m}{\includegraphics[width=0.084\textwidth]{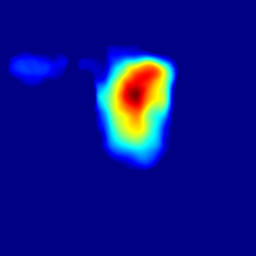}}
		&
		\adjustbox{valign=m}{\includegraphics[width=0.084\textwidth]{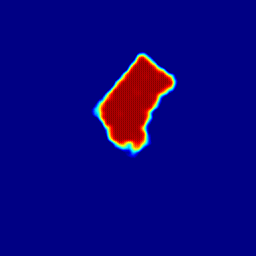}}
		&
		\adjustbox{valign=m}{\includegraphics[width=0.084\textwidth]{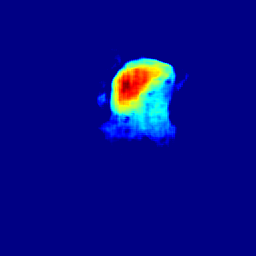}}
		&
		\adjustbox{valign=m}{\includegraphics[width=0.084\textwidth]{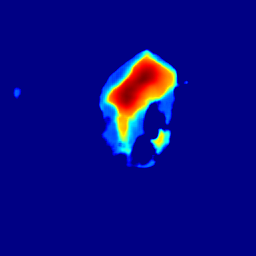}}
		&
		\adjustbox{valign=m}{\includegraphics[width=0.084\textwidth]{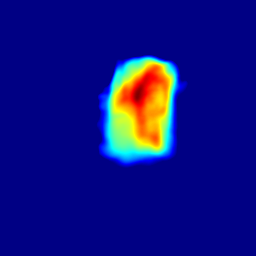}}
		\\
		\vspace{1mm}
		\adjustbox{valign=m}{\includegraphics[width=0.084\textwidth]{204.PNG}}
		&
		\adjustbox{valign=m}{\includegraphics[width=0.084\textwidth]{204_gt.PNG}}
		&
		\adjustbox{valign=m}{\includegraphics[width=0.084\textwidth]{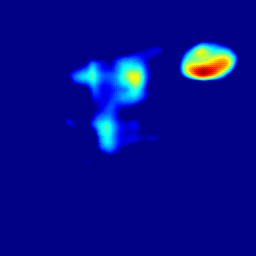}}
		&
		\adjustbox{valign=m}{\includegraphics[width=0.084\textwidth]{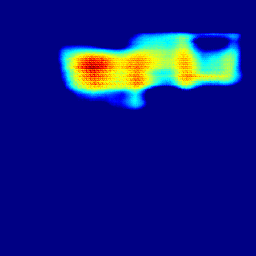}}
		&
		\adjustbox{valign=m}{\includegraphics[width=0.084\textwidth]{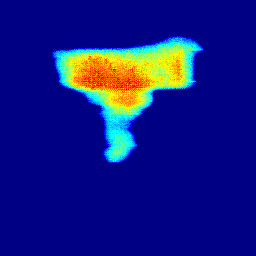}}
		&
		\adjustbox{valign=m}{\includegraphics[width=0.084\textwidth]{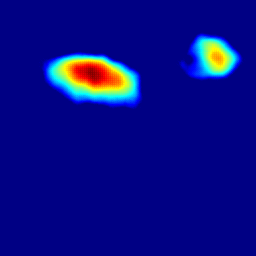}}
		&
		\adjustbox{valign=m}{\includegraphics[width=0.084\textwidth]{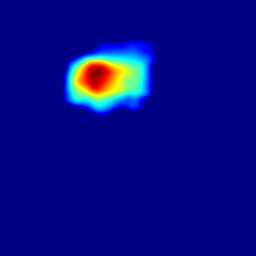}}
		&
		\adjustbox{valign=m}{\includegraphics[width=0.084\textwidth]{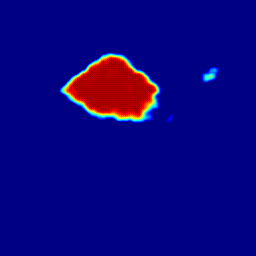}}
		&
		\adjustbox{valign=m}{\includegraphics[width=0.084\textwidth]{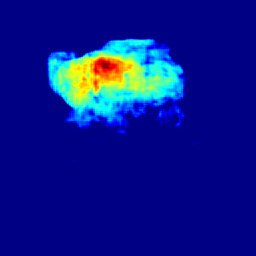}}
		&
		\adjustbox{valign=m}{\includegraphics[width=0.084\textwidth]{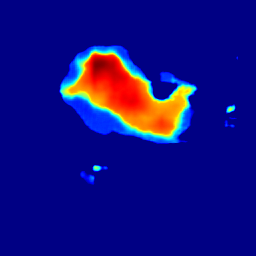}}
		&
		\adjustbox{valign=m}{\includegraphics[width=0.084\textwidth]{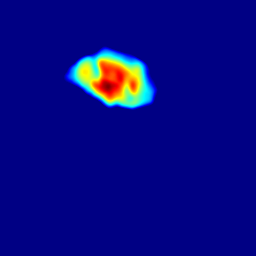}}
		\\
		\vspace{1mm}
		\adjustbox{valign=m}{\includegraphics[width=0.084\textwidth]{377.PNG}}
		&
		\adjustbox{valign=m}{\includegraphics[width=0.084\textwidth]{377_gt.PNG}}
		&
		\adjustbox{valign=m}{\includegraphics[width=0.084\textwidth]{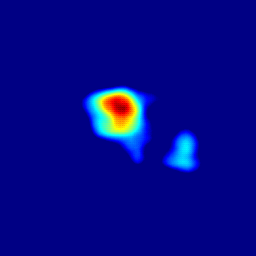}}
		&
		\adjustbox{valign=m}{\includegraphics[width=0.084\textwidth]{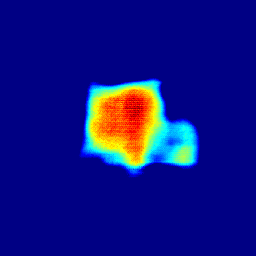}}
		&
		\adjustbox{valign=m}{\includegraphics[width=0.084\textwidth]{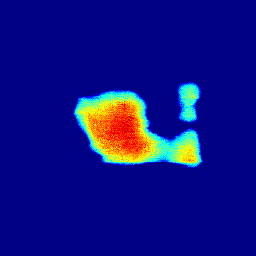}}
		&
		\adjustbox{valign=m}{\includegraphics[width=0.084\textwidth]{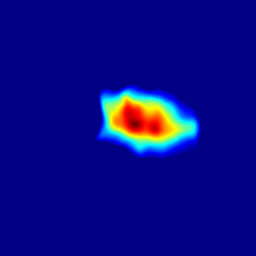}}
		&
		\adjustbox{valign=m}{\includegraphics[width=0.084\textwidth]{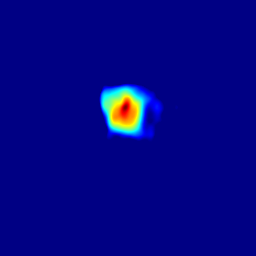}}
		&
		\adjustbox{valign=m}{\includegraphics[width=0.084\textwidth]{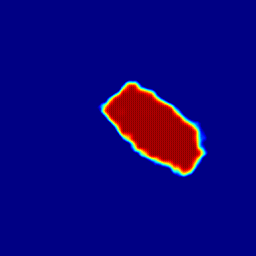}}
		&
		\adjustbox{valign=m}{\includegraphics[width=0.084\textwidth]{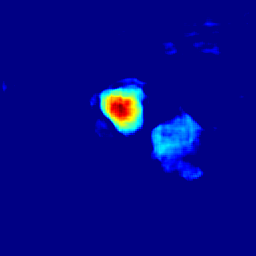}}
		&
		\adjustbox{valign=m}{\includegraphics[width=0.084\textwidth]{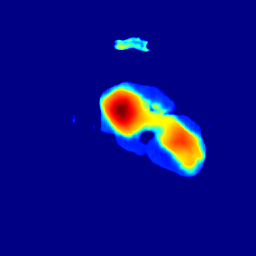}}
		&
		\adjustbox{valign=m}{\includegraphics[width=0.084\textwidth]{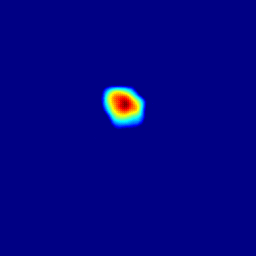}}
		\\
		Images & GT & U-Net & SCRF & UNCRF & Boxinst & CoarseFine & WSDAC & S2ME & IDMPS & proposed
		\\
	\end{tabular}
	\caption{\textbf{Feature map visualizations of segmentation results in Fig.~\ref{fig:comparison_tn3k} and Fig.~\ref{fig:comparison_ddti}.} All the methods are the same as those in Fig.~\ref{fig:comparison_tn3k} and Fig.~\ref{fig:comparison_ddti}.}
	\label{fig:discussion_featuremap}
\end{figure*}

Weakly Supervised Segmentation (WSS) methods have attracted increasing attention due to their ability to utilize sparse annotations for generating segmentation results, thereby reducing the need for fully annotated masks. However, these methods often encounter limitations due to label noise stemming from low-confidence pseudo-labels and insufficient discriminative features extracted for diverse and complex nodule variations through rigid learning strategies. In order to further explore the reasons for ultrasound image feature perception behind segmentation results, feature maps generated by different comparison methods are visualized in Fig.~\ref{fig:discussion_featuremap}.

SCRF~\cite{zhang2020scrf} and UNCRF~\cite{mahani2022uncrf} generated box-like feature distributions and predictions because they directly utilized box labels to learn segmentation results, which introduced significant inaccuracies in shape representation and misguides the training process. 

Similarly, S2ME~\cite{wang2023s2me} and IDMPS~\cite{zhao2024IDMPS} performed well on the DDTI dataset, where shape variations are minimal. However, their performance decreased on the TN3K dataset, which suffered from diverse and irregular nodule shapes. This decline in effectiveness was due to their reliance on fixed geometric pseudo-labels for shape learning, resulting in failure to capture the high-confidence discriminative features needed for accurately segmenting nodules with complex and variable shapes.

BoxInst~\cite{tian2021BoxInst} exhibited reasonable performance on the larger TN3K dataset but struggled with the smaller DDTI dataset due to its heavy dependence on color similarity for learning segmentation shapes. This approach was effective for high-contrast images but required a larger volume of training data. When the amount of training samples is limited, the uncertainty of feature distribution is difficult to remove.

CoarseFine~\cite{chi2025coarse} exhibited relatively balanced performance on two datasets, but the use of rough references to refine shapes still led to evidence of over-segmentation and under-segmentation issues in its segmentation results.

WSDAC~\cite{li2023wsdac} consistently produced under-segmentation results because it heavily relied on initial contours and image gradients, which have been proven to be less effective for images with blurred boundaries.

In contrast, our proposed method effectively addresses these challenges through two key innovations: (1) the generation of high-confidence labels to mitigate label noise and improve training stability, and (2) the introduction of a high-rationality learning strategy to capture location-level, region-level, and boundary-level features for segmentation location and shape learning. These advancements lead to more accurate localization of nodule feature arrangement, overall shape of nodule feature distribution, and delicate boundaries between nodule and background features, showing comparable or even surpassing results to those of fully supervised networks.

% demonstrating its robustness and effectiveness in addressing the unique challenges posed by thyroid ultrasound image segmentation.

\subsection{The Generalizability of the Proposed Weakly Supervised Segmentation Framework}
\label{sec:scalability}
%\begin{figure*}[!htbp]
%\centering
%\includegraphics[width=0.98\textwidth]{Generalization.png}
%\caption{\textbf{Qualitative results of different extended applications on TN3K and DDTI dataset}. The X-pixel denotes using ground truth for pixel-to-pixel comparison. The X-hchr indicates using high-confidence weakly supervised labels for high-rationality multi-level learning. Red indicates correct thyroid nodule predictions, green represents the under-segmentations of thyroid nodules, and blue shows an over-segmentations of other organs as thyroid nodules.}
%% \vspace{-0.2in}
%\label{fig:comparison_generalization}
%\end{figure*}

\begin{figure*}[!htbp]
\footnotesize
\centering
\tabcolsep=0.5mm
\begin{tabular}{cccccccccccc}
	\vspace{1mm}
	\adjustbox{valign=m}{\includegraphics[width=0.075\textwidth]{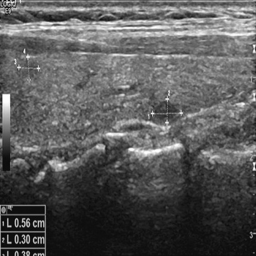}}
	&
	\adjustbox{valign=m}{\includegraphics[width=0.075\textwidth]{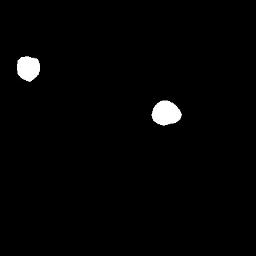}}
	&
	\adjustbox{valign=m}{\includegraphics[width=0.075\textwidth]{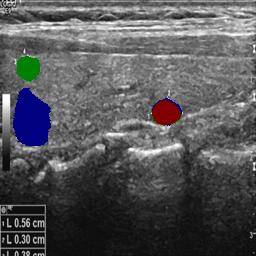}}
	&
	\adjustbox{valign=m}{\includegraphics[width=0.075\textwidth]{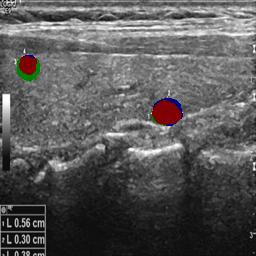}}
	&
	\adjustbox{valign=m}{\includegraphics[width=0.075\textwidth]{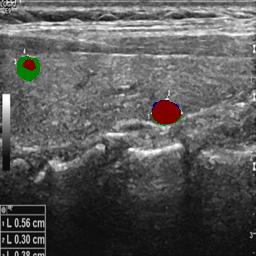}}
	&
	\adjustbox{valign=m}{\includegraphics[width=0.075\textwidth]{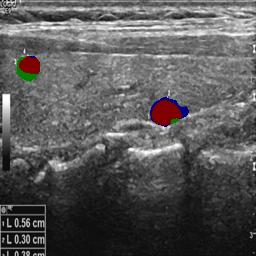}}
	&
	\adjustbox{valign=m}{\includegraphics[width=0.075\textwidth]{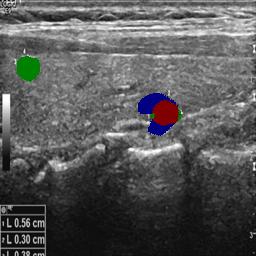}}
	&
	\adjustbox{valign=m}{\includegraphics[width=0.075\textwidth]{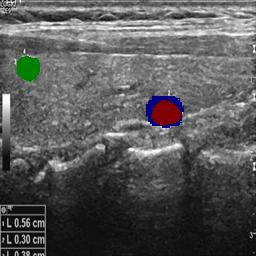}}
	&
	\adjustbox{valign=m}{\includegraphics[width=0.075\textwidth]{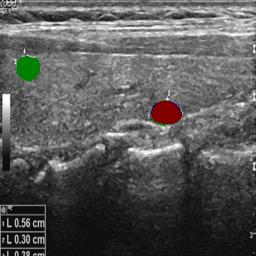}}
	&
	\adjustbox{valign=m}{\includegraphics[width=0.075\textwidth]{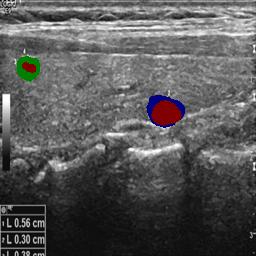}}
	&
	\adjustbox{valign=m}{\includegraphics[width=0.075\textwidth]{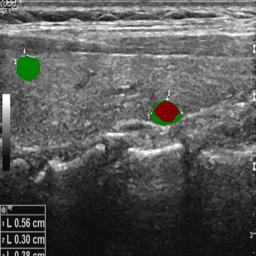}}
	&
	\adjustbox{valign=m}{\includegraphics[width=0.075\textwidth]{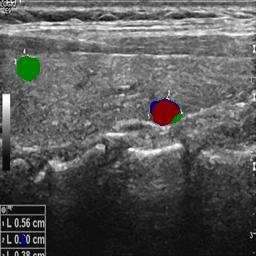}}
	\\
	\vspace{1mm}
	\adjustbox{valign=m}{\includegraphics[width=0.075\textwidth]{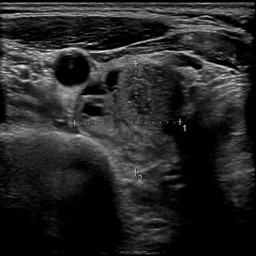}}
	&
	\adjustbox{valign=m}{\includegraphics[width=0.075\textwidth]{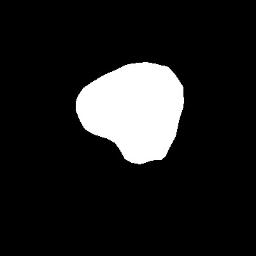}}
	&
	\adjustbox{valign=m}{\includegraphics[width=0.075\textwidth]{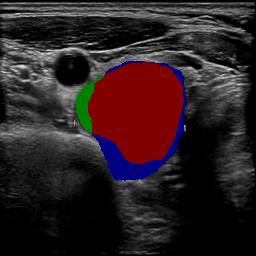}}
	&
	\adjustbox{valign=m}{\includegraphics[width=0.075\textwidth]{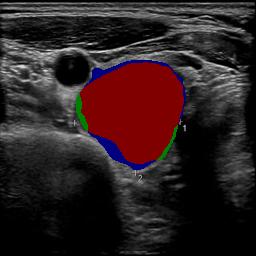}}
	&
	\adjustbox{valign=m}{\includegraphics[width=0.075\textwidth]{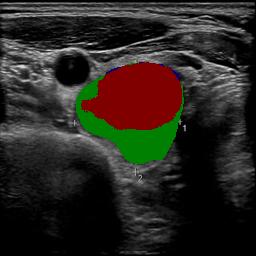}}
	&
	\adjustbox{valign=m}{\includegraphics[width=0.075\textwidth]{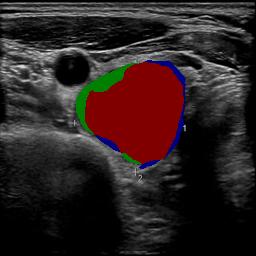}}
	&
	\adjustbox{valign=m}{\includegraphics[width=0.075\textwidth]{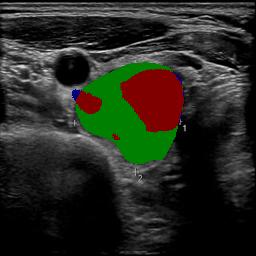}}
	&
	\adjustbox{valign=m}{\includegraphics[width=0.075\textwidth]{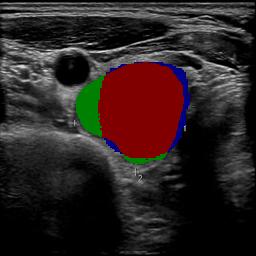}}
	&
	\adjustbox{valign=m}{\includegraphics[width=0.075\textwidth]{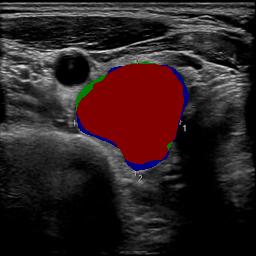}}
	&
	\adjustbox{valign=m}{\includegraphics[width=0.075\textwidth]{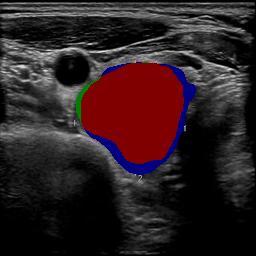}}
	&
	\adjustbox{valign=m}{\includegraphics[width=0.075\textwidth]{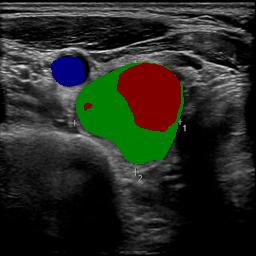}}
	&
	\adjustbox{valign=m}{\includegraphics[width=0.075\textwidth]{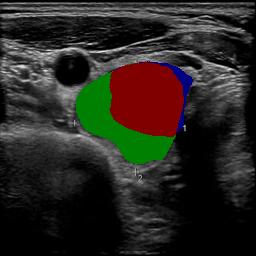}}
	\\
	\vspace{1mm}
	\adjustbox{valign=m}{\includegraphics[width=0.075\textwidth]{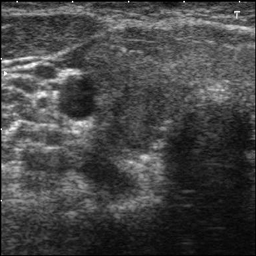}}
	&
	\adjustbox{valign=m}{\includegraphics[width=0.075\textwidth]{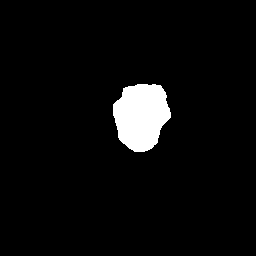}}
	&
	\adjustbox{valign=m}{\includegraphics[width=0.075\textwidth]{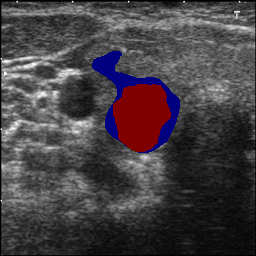}}
	&
	\adjustbox{valign=m}{\includegraphics[width=0.075\textwidth]{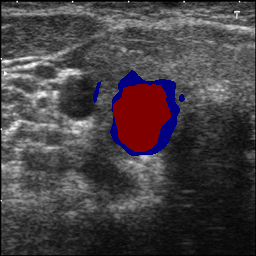}}
	&
	\adjustbox{valign=m}{\includegraphics[width=0.075\textwidth]{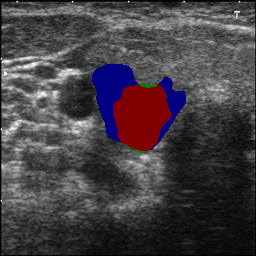}}
	&
	\adjustbox{valign=m}{\includegraphics[width=0.075\textwidth]{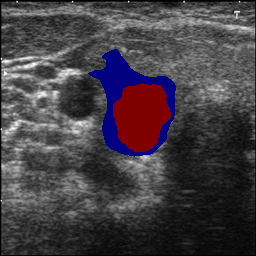}}
	&
	\adjustbox{valign=m}{\includegraphics[width=0.075\textwidth]{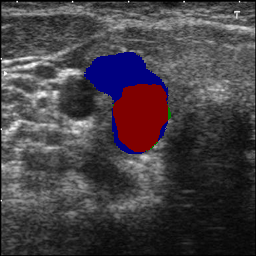}}
	&
	\adjustbox{valign=m}{\includegraphics[width=0.075\textwidth]{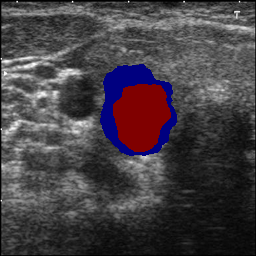}}
	&
	\adjustbox{valign=m}{\includegraphics[width=0.075\textwidth]{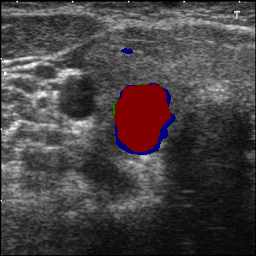}}
	&
	\adjustbox{valign=m}{\includegraphics[width=0.075\textwidth]{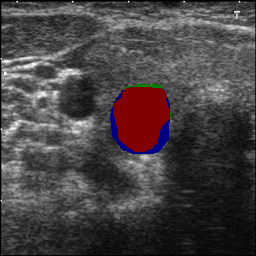}}
	&
	\adjustbox{valign=m}{\includegraphics[width=0.075\textwidth]{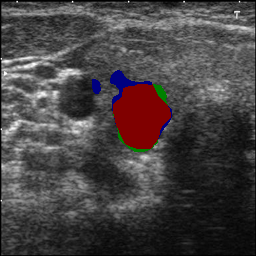}}
	&
	\adjustbox{valign=m}{\includegraphics[width=0.075\textwidth]{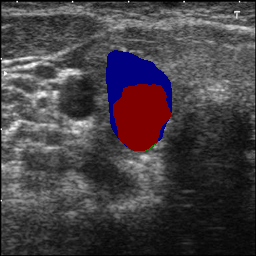}}
	\\
	\vspace{1mm}
	\adjustbox{valign=m}{\includegraphics[width=0.075\textwidth]{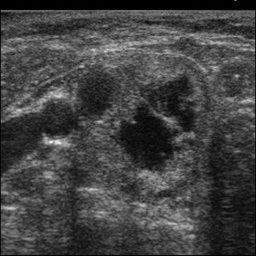}}
	&
	\adjustbox{valign=m}{\includegraphics[width=0.075\textwidth]{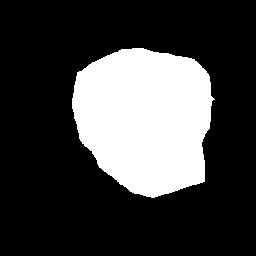}}
	&
	\adjustbox{valign=m}{\includegraphics[width=0.075\textwidth]{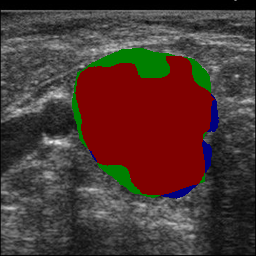}}
	&
	\adjustbox{valign=m}{\includegraphics[width=0.075\textwidth]{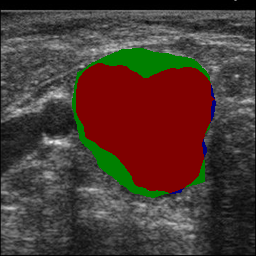}}
	&
	\adjustbox{valign=m}{\includegraphics[width=0.075\textwidth]{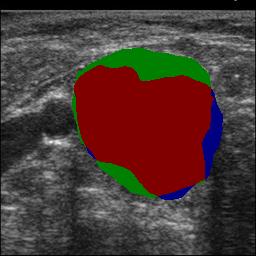}}
	&
	\adjustbox{valign=m}{\includegraphics[width=0.075\textwidth]{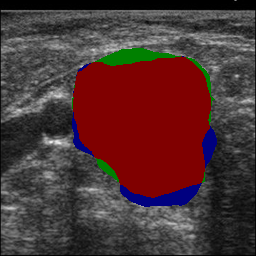}}
	&
	\adjustbox{valign=m}{\includegraphics[width=0.075\textwidth]{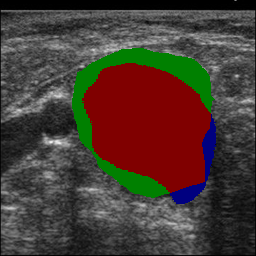}}
	&
	\adjustbox{valign=m}{\includegraphics[width=0.075\textwidth]{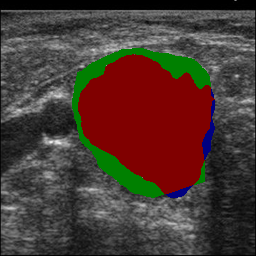}}
	&
	\adjustbox{valign=m}{\includegraphics[width=0.075\textwidth]{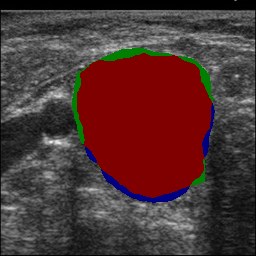}}
	&
	\adjustbox{valign=m}{\includegraphics[width=0.075\textwidth]{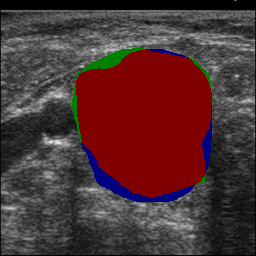}}
	&
	\adjustbox{valign=m}{\includegraphics[width=0.075\textwidth]{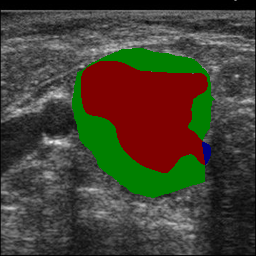}}
	&
	\adjustbox{valign=m}{\includegraphics[width=0.075\textwidth]{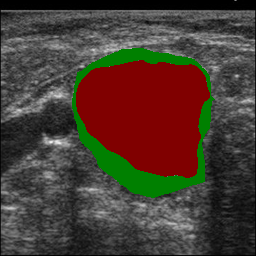}}
	\\
	Images & GT & U-Net & U-Net$^{*}$ & AU-Net & AU-Net$^{*}$ & DUNet & DUNet$^{*}$ & CENet & CENet$^{*}$ & TUNet & TUNet$^{*}$
	\\
\end{tabular}
\caption{\textbf{Qualitative results of different extended applications on the TN3K dataset}. U-Net, AU-Net, DUNet, CENet and TUNet denote the U-Net, Dense-UNet, Attention U-Net, CENet and TransUNet models in Table~\ref{tab:calability_results_tn3k} tarined by the given ground truth masks. U-Net$^{*}$, AU-Net$^{*}$, DUNet$^{*}$, CENet$^{*}$ and TUNet$^{*}$ denote those models trained by the proposed weakly supervised learning framework. Row 1 and 2 are examples from the TN3K dataset, while Row 3 and 4 are examples from the DDTI dataset. Red indicates correct thyroid nodule predictions, green represents the under-segmentations of thyroid nodules, and blue shows an over-segmentations of other tissues as thyroid nodules.}
	% \vspace{-0.2in}
\label{fig:comparison_generalization}
\end{figure*}

\begin{table*}[!htbp]
\centering
\caption{Quantitative results (Mean $\pm$ STD) of applying the proposed weakly supervised segmentation framework to different segmentation backbone models on the TN3K Datasett~\cite{gong2021tn3k}. Symbol $^*$ indicates training the model with our proposed weakly supervised segmentation framework, while symbol represents the original fully supervised learning.}
%\label{tab:ablation_tn3k}
\small
\setlength{\tabcolsep}{4pt}
\renewcommand{\arraystretch}{1.3}
\begin{tabular}{lccccc}
	\toprule[1.2pt]
	\multirow{2}{*}{Supervision} & \multirow{2}{*}{Model(Backbone)} & \multicolumn{4}{c}{Metrics} \\
	\cmidrule(r){3-6}
	& & mIoU (\%) $\uparrow$ & DSC (\%) $\uparrow$ & Precision (\%) $\uparrow$ & HD95 (px) $\downarrow$ \\
	\midrule[0.8pt]
	Fully & U-Net~\cite{ronneberger2015unet} & $68.49\pm24.34$ & $78.07\pm22.76$ & $79.08\pm24.36$ & $21.13\pm30.69$\\
	Weakly & U-Net$^{*}$~\cite{ronneberger2015unet} & $\textbf{69.30}\pm\textbf{22.38}$ & $\textbf{79.10}\pm\textbf{21.37}$ & $\textbf{82.49}\pm\textbf{22.69}$ & $\textbf{17.20}\pm\textbf{26.97}$ \\
	\cmidrule(lr){1-6}
	Fully & Dense-UNet~\cite{cai2020Denseunet} & $66.61\pm25.74$ & $76.27\pm25.19$ & $77.95\pm26.47$ & $20.48\pm28.94$ \\
	Weakly & Dense-UNet$^*$~\cite{cai2020Denseunet} & $\textbf{68.77}\pm\textbf{23.68}$ & $\textbf{78.32}\pm\textbf{22.77}$ & $\textbf{79.04}\pm\textbf{23.23}$ & $\textbf{18.19}\pm\textbf{28.84}$ \\
	\cmidrule(lr){1-6}
	Fully & Attention U-Net~\cite{oktay2018attention} & $69.21\pm26.38$ & $77.85\pm25.78$ & $\textbf{82.11}\pm\textbf{26.28}$ & $\textbf{16.71}\pm\textbf{27.54}$\\
	Weakly & Attention U-Net$^{*}$~\cite{oktay2018attention} & $\textbf{69.49}\pm\textbf{23.83}$ & $\textbf{78.85}\pm\textbf{22.84}$ & $81.09\pm24.46$ & $17.60\pm27.86$ \\
	\cmidrule(lr){1-6}
	Fully & CENet~\cite{tao2022cenet} & $\textbf{74.00}\pm\textbf{20.74}$ & $\textbf{82.92}\pm\textbf{18.32}$ & $84.68\pm19.28$ & $\textbf{14.31}\pm\textbf{23.93}$ \\
	Weakly & CENet$^{*}$~\cite{tao2022cenet} & $73.32\pm20.12$ & $82.56\pm17.97$ & $\textbf{85.31}\pm\textbf{18.80}$ & $14.83\pm25.37$ \\
	\cmidrule(lr){1-6}
	% & MedT\cite{valanarasu2021medT} & $46.01\pm26.91$ & $57.95\pm27.95$ & $62.98\pm30.90$ & $37.24\pm32.88$ \\
	Fully & TransUNet~\cite{chen2024transunet} & $67.13\pm25.15$ & $76.92\pm23.14$ & $79.36\pm25.76$ & $22.14\pm29.56$ \\
	Weakly & TransUNet$^{*}$~\cite{chen2024transunet} & $\textbf{68.92}\pm\textbf{23.28}$ & $\textbf{78.73}\pm\textbf{21.24}$ & $\textbf{81.06}\pm\textbf{22.80}$ & $\textbf{19.38}\pm\textbf{29.94}$ \\
%	\cmidrule(lr){1-6}
%	\multirow{5}{*}{Weakly}
	% & MedT\cite{valanarasu2021medT}$^{*}$ &  &  &  &  \\
	\bottomrule[1.2pt]
\end{tabular}
\vspace{2mm}
\label{tab:calability_results_tn3k}
\end{table*}
\begin{table*}[!htbp]
\centering
\caption{Quantitative results (Mean $\pm$ STD) of applying the proposed weakly supervised segmentation framework to different segmentation backbone models on the DDTI Dataset~\cite{pedraza2015DDTI}. Symbol $^*$ indicates training the model with our proposed weakly supervised segmentation framework, while no symbol represents the original fully supervised learning.}
%\label{tab:ablation_tn3k}
\small
\setlength{\tabcolsep}{4pt}
\renewcommand{\arraystretch}{1.3}
\begin{tabular}{lccccc}
\toprule[1.2pt]
\multirow{2}{*}{Supervision} &\multirow{2}{*}{Model(Backbone)} & \multicolumn{4}{c}{Metrics} \\
\cmidrule(r){3-6}
& & mIoU (\%) $\uparrow$ & DSC (\%) $\uparrow$ & Precision (\%) $\uparrow$ & HD95 (px) $\downarrow$\\
\midrule[0.8pt]
%\multirow{5}{*}{Fully}
Fully & U-Net~\cite{ronneberger2015unet} & $58.97\pm23.15$ & $71.05\pm21.88$ & $70.83\pm27.23$ & $27.63\pm26.81$ \\
Weakly & U-Net$^{*}$~\cite{ronneberger2015unet} & $\textbf{62.52}\pm\textbf{20.72}$ & $\textbf{74.55}\pm\textbf{18.91}$ & $\textbf{74.66}\pm\textbf{24.62}$ & $\textbf{23.82}\pm\textbf{24.77}$ \\
\cmidrule(lr){1-6}
Fully & Dense-UNet~\cite{cai2020Denseunet} & $55.01\pm23.53$ & $67.51\pm23.18$ & $68.41\pm27.62$ & $28.36\pm23.34$ \\
Weakly & Dense-UNet$^{*}$~\cite{cai2020Denseunet} & $\textbf{59.00}\pm\textbf{21.07}$& $\textbf{71.64}\pm\textbf{19.77}$ & $\textbf{72.50}\pm\textbf{25.21}$ & $\textbf{26.50}\pm\textbf{25.33}$ \\
\cmidrule(lr){1-6}
Fully & Attention U-Net~\cite{oktay2018attention} & $56.07\pm23.80$ & $68.56\pm21.88$ & $68.78\pm26.83$ & $28.87\pm25.90$ \\
Weakly & Attention U-Net$^{*}$~\cite{oktay2018attention} & $\textbf{58.42}\pm\textbf{23.65}$ & $\textbf{70.41}\pm\textbf{23.31}$ & $\textbf{69.03}\pm\textbf{26.14}$ & $\textbf{26.13}\pm\textbf{24.86}$ \\
\cmidrule(lr){1-6}
Fully & CENet~\cite{tao2022cenet} & $68.75\pm22.40$ & $78.84\pm20.15$ & $79.50\pm23.53$ & $19.88\pm25.19$ \\
Weakly & CENet$^{*}$~\cite{tao2022cenet} & $\textbf{70.35}\pm\textbf{18.85}$ & $\textbf{80.78}\pm\textbf{16.63}$ & $\textbf{79.65}\pm\textbf{20.72}$ & $\textbf{15.08}\pm\textbf{19.59}$ \\
\cmidrule(lr){1-6}
% & MedT~\cite{valanarasu2021medT} & $46.92\pm21.63$ & $60.65\pm22.12$ & $60.73\pm27.25$ & $31.10\pm19.95$ \\
Fully & TransUNet~\cite{chen2024transunet} & $48.83\pm24.38$ & $61.65\pm24.60$ & $62.11\pm29.67$ & $36.13\pm28.04$ \\
Weakly & TransUNet$^{*}$~\cite{chen2024transunet} & $\textbf{51.39}\pm\textbf{22.81}$ & $\textbf{64.59}\pm\textbf{22.18}$ & $\textbf{62.62}\pm\textbf{28.93}$ & $\textbf{32.66}\pm\textbf{24.61}$ \\
%\multirow{5}{*}{Weakly}
% & MedT~\cite{valanarasu2021medT}$^{*}$ &  &  &  &  \\
\bottomrule[1.2pt]
\end{tabular}
\vspace{2mm}
\label{tab:calability_results_ddti}
\end{table*}

% The framework proposed in this paper focuses on the generation of label information and the implementation of learning strategies. The proposed framework has high scalability, allowing easy replacement of feature extraction networks. 
% For example, adding attention mechanisms (Attention U-Net~\cite{oktay2018attention}) or enhancing context understanding (CENet~\cite{tao2022cenet}) in the feature extraction part to improve the performance of weakly supervised segmentation.
Instead of network architecture, our proposed network focuses on the label information refinement and learning strategy optimization of weakly supervised segmentation, leading to high generalizability to different segmentation backbones. 
% The proposed framework has high scalability, allowing easy replacement of feature extraction networks.

% As shown in Table~\ref{tab:calability_results_tn3k} and Table~\ref{tab:calability_results_ddti}, under the same feature extraction network, adopting the framework proposed in this paper, weakly-supervised segmentation results achieve comparable performance and even outperform the pixel-to-pixel fully supervised learning algorithm. 
As shown in Fig.~\ref{fig:comparison_generalization}, by applying the proposed high-confidence and high-rationality weakly supervised framework, different segmentation backbone models achieved comparable or even improved performance to those trained following the pixel-to-pixel fully supervised strategy. As illustrated in Table~\ref{tab:calability_results_tn3k}, 
% In addition, the proposed high-rationality multi-level learning strategy can be used for fully supervised tasks. Using the proposed high-rationality learning strategy with the same feature extraction network, 
on the TN3k dataset, only CENet with fully supervised training slightly outperformed our method, with a marginal increase of 0.68\% in mIoU. Other networks adopting the HCHR framework exhibited higher performance with an increase in mIoU ranging from 0.28\% (Attention U-Net) to 2.16\% (Dense-UNet) compared to the corresponding fully supervised framework. This is despite the fully supervised network's use of labor-intensive, more precise ground truth masks, whereas our approach relied on point annotations for weakly supervised learning. As shown in Table~\ref{tab:calability_results_ddti}, on the DDTI dataset with a smaller number of samples, leveraging our weakly supervised algorithm outperformed all fully supervised methods across all metrics, under the same backbone architecture. Notably, it achieved a significant improvement of over 1.5\% in mIoU and a reduction of more than 1.86 pixels in HD95. These results demonstrate the exceptional generalizability of our framework, which allows for the seamless replacement of feature extraction networks.

Therefore, for weakly supervised segmentation tasks that can provide multi-level labels for localization, partial foreground and background references, our framework provides a simple but effective solution for segmenting lesions with weakly supervised annotations.

\subsection{Limitations and Future work}
Our method illustrates strong segmentation performance on the thyroid nodule datasets, showcasing its potential for weakly supervised learning. However, there are still avenues for further enhancement.

Firstly, while our algorithm achieved a moderate inference speed of 0.0065 seconds per image, it integrated prompted MedSAM results to generate high-confidence labels for training, which required a preprocessing step for annotations. In future work, adopting more specialized and lightweight medical foundation models to generate labels could streamline this process and further reduce training complexity.

Secondly, the model introduced three loss functions to effectively constrain the learning of position and shape with two hyperparameters $\lambda$ and $\beta$ that need to be fine-tuned through ablation experiments. These hyperparameters offer potential for further optimization. Designing a solution to find the optimal parameters can further improve the performance of the algorithm in the future.
% By transforming them into learnable parameters that can be dynamically adjusted based on segmentation feature learning, we expect this adjustment to significantly improve the model’s performance.

As outlined in Sec. \ref{sec:scalability}, our framework's feature extraction backbone was designed with adaptability in mind, allowing for seamless integration with different task requirements. Future research can explore how to tailor our framework for a range of segmentation tasks by developing task-specific feature extraction backbones, unlocking even broader applications for this method.

\section{Conclusion}
In this paper, we present a novel weakly supervised segmentation framework for thyroid nodule segmentation based on clinical point annotations. We clarify the segmentation objective that integrates location and shape learning to indicate the training process. Our method integrates geometric transformations with topology priors and the prompted MedSAM results with anatomical information to generate high-confidence labels. Furthermore, we propose a high-rationality learning strategy through multi-level losses. The alignment loss is for precise location learning, while contrastive and prototype correlation losses are for robust shape understanding. Experimental results demonstrate superior performance compared to state-of-the-art weakly supervised methods on the TN3K and DDTI dataset. The framework is highly versatile and can be seamlessly integrated into various feature extraction architectures, offering flexibility for diverse clinical application scenarios.

\section{Acknowledgement}
This work was supported in part by the National Natural Science Foundation of China under Grants 61901098, and the Provincial Fundamental Research for Liaoning under Grant 2019JH1/10100005.

\bibliographystyle{cas-model2-names}

\bibliography{cas-refs}


%Author biography without author photo.
%Author biography. Author biography. Author biography.
%Author biography. Author biography. Author biography.
%Author biography. Author biography. Author biography.
%Author biography. Author biography. Author biography.
%Author biography. Author biography. Author biography.
%Author biography. Author biography. Author biography.
%Author biography. Author biography. Author biography.
%Author biography. Author biography. Author biography.
%Author biography. Author biography. Author biography.
%\bio{figs/cas-pic1}
%Author biography with author photo.
%Author biography. Author biography. Author biography.
%Author biography. Author biography. Author biography.
%Author biography. Author biography. Author biography.
%Author biography. Author biography. Author biography.
%Author biography. Author biography. Author biography.
%Author biography. Author biography. Author biography.
%Author biography. Author biography. Author biography.
%Author biography. Author biography. Author biography.
%Author biography. Author biography. Author biography.
%\vskip3pc
%
%\bio{figs/cas-pic1}
%Author biography with author photo.
%Author biography. Author biography. Author biography.
%Author biography. Author biography. Author biography.
%Author biography. Author biography. Author biography.
%Author biography. Author biography. Author biography.

\end{document}